\documentclass{article}

\usepackage[nonatbib,final]{neurips_data_2021}

\usepackage[utf8]{inputenc} 
\usepackage[T1]{fontenc}    
\usepackage{hyperref}       
\usepackage{url}            
\usepackage{booktabs}       
\usepackage{amsfonts}       
\usepackage{nicefrac}       
\usepackage{microtype}      
\usepackage[table]{xcolor}         
\usepackage{subfiles}

\usepackage[nonamelimits]{amsmath}
\usepackage{amssymb,stmaryrd}
\usepackage{bm}
\usepackage{mathtools}
\usepackage{geometry}
\usepackage{graphicx}
\usepackage{color}
\usepackage{amssymb}
\usepackage{hyperref}

\usepackage{gensymb}

\usepackage{hyperref}
\usepackage{url}
\usepackage{empheq}
\usepackage[utf8]{inputenc} 
\usepackage[T1]{fontenc}    
\usepackage{hyperref}       
\usepackage{url}            
\usepackage{booktabs}       
\usepackage{amsfonts}       
\usepackage{nicefrac}       
\usepackage{microtype}      
\usepackage{multirow}
\usepackage{multicol}
\usepackage{subfiles}
\usepackage{todonotes}
\usepackage{enumitem}
\usepackage{subcaption}
\usepackage[capitalise]{cleveref}

\usepackage{placeins}

\usepackage{appendix}

\usepackage[normalem]{ulem}
\useunder{\uline}{\ul}{}
\usepackage{siunitx}    

\newcommand\blfootnote[1]{%
  \begingroup
  \renewcommand\thefootnote{}\footnote{#1}%
  \addtocounter{footnote}{-1}%
  \endgroup
}

\title{Shifts: A Dataset of Real Distributional Shift Across Multiple Large-Scale Tasks}

\author{Andrey Malinin\textsuperscript{1,2} \and Neil Band\textsuperscript{5}  \and Yarin Gal\textsuperscript{5,6} \and Mark J. F. Gales\textsuperscript{4} \and Alexander Ganshin \textsuperscript{1} \and German Chesnokov\textsuperscript{1} \and Alexey Noskov\textsuperscript{1} \and  Andrey Ploskonosov\textsuperscript{1} \and  Liudmila Prokhorenkova\textsuperscript{1,2,3} \and  Ivan Provilkov\textsuperscript{1,3} \and Vatsal Raina\textsuperscript{4} \and Vyas Raina\textsuperscript{4}  \and Denis Roginskiy \textsuperscript{1}  \and  Mariya Shmatova\textsuperscript{1}  \and Panos Tigas\textsuperscript{5} \and Boris Yangel\textsuperscript{1} \\
\hspace{-100pt}\texttt{am969@yandex-team.ru}
\\
}

\begin{document}

\maketitle

\begin{abstract}
 There has been significant research done on developing methods for improving robustness to distributional shift and uncertainty estimation. In contrast, only limited work has examined developing standard datasets and benchmarks for assessing these approaches. Additionally, most work on uncertainty estimation and robustness has developed new techniques based on small-scale regression or image classification tasks. However, many tasks of practical interest have different modalities, such as tabular data, audio, text, or sensor data,  which offer significant challenges involving regression and discrete or continuous structured prediction. Thus, given the current state of the field, a standardized large-scale dataset of tasks across a range of modalities affected by distributional shifts is necessary. This will enable researchers to meaningfully evaluate the plethora of recently developed uncertainty quantification methods, as well as assessment criteria and state-of-the-art baselines. In this work, we propose the \emph{Shifts Dataset} for evaluation of uncertainty estimates and robustness to distributional shift. The dataset, which has been collected from industrial sources and services, is composed of three tasks, with each corresponding to a particular data modality: tabular weather prediction, machine translation, and self-driving car (SDC) vehicle motion prediction. All of these data modalities and tasks are affected by real, ``in-the-wild'' distributional shifts and pose interesting challenges with respect to uncertainty estimation. In this work we provide a description of the dataset and baseline results for all tasks.
\end{abstract}

\blfootnote{$^{1}$Yandex, $^{2}$HSE University, $^{3}$Moscow Institute of Physics and Technology, $^{4}$ALTA Institute,  University of Cambridge, $^{5}$University of Oxford, $^{6}$Alan Turing Institute}

\section{Introduction}\label{sec:intro}

Machine learning models are being applied to numerous areas~\cite{vgg,embedding1,mikolov-rnn,bahdanau2015nmt,attentionisallyouneed,dnnspeech} and are widely deployed in production. An assumption which pervades all of machine learning is that the training, validation, and deployment data are independent and identically distributed (i.i.d.). Thus, good performance and generalization on validation data imply that the model will perform well in deployment. Unfortunately, this assumption seldom holds in real, ``in the wild'', applications. In practice, data are subject to a wide range of possible \emph{distributional shifts}~--- mismatches between the training data, and test or deployment data \cite{Datasetshift,koh2020wilds,nado2021uncertainty}. In general, the greater the degree of shift, the poorer is the model's performance. The problem of distributional shift is of relevance not only to academic researchers, but to the machine learning community at-large. Indeed, \emph{all} ML practitioners have faced the issue of mismatch between the training and test sets. This is especially important in high-risk applications of machine learning, such as finance, medicine, and autonomous vehicles. In such applications a mistake on part of an ML system may incur financial or reputational loss, or possible loss of life. It is therefore increasingly important to assess both a model's \emph{robustness} to distribution shift and its estimates of \emph{predictive uncertainty}, which enable it to detect distributional shifts~\cite{aisafety,galthesis,malinin-thesis}.

The area of uncertainty estimation and robustness has developed rapidly in recent years. Model averaging ~\cite{Gal2016Dropout,deepensemble2017,ashukha2020pitfalls, trust-uncertainty} has emerged as the de-facto standard approach to uncertainty estimation.  Ensemble- and sampling-based uncertainty estimates have been successfully applied in detecting misclassifications, out-of-distribution inputs, adversarial attacks \cite{carlini-detected, gal-adversarial}, and for active learning~\cite{kirsch2019batchbald}. Recently, such approaches have been extended to structured prediction tasks such as machine translation and speech recognition~\cite{xiao2019wat,notinprincipled,fomicheva2020unsupervised,yarin-mt-uncertainty,malinin2021structured}. However, these approaches require large computational and memory budgets. Works using temperature scaling \cite{liang2020enhancing,hsu2020generalized} and other recent approaches in deterministic uncertainty estimation \cite{van2020uncertainty, havasi2020training, liu2020simple, van2021improving, mukhoti2021deterministic} aim to tackle this issue, but have only recently become comparable to ensemble methods \cite{van2021improving, mukhoti2021deterministic}. Prior Networks~\cite{malinin-pn-2018,malinin-rkl-2019,malinin2020regression}~--- models which \emph{emulate} the mechanics of an ensemble~--- have been proposed as a deterministic single model approach to uncertainty estimation which are competitive with ensembles. However, they require distributionally shifted training data, which may not be feasible in many applications. Prior Networks have also been used for \emph{Ensemble Distribution Distillation}~\cite{malinin-endd-2019,malinin2020regression, ryabinin2021scaling}~--- a distillation approach through which the predictive performance and uncertainty estimates of an ensemble are captured within a single Prior Network, reducing the inference cost to that of a single model.

While much work has been done on developing \emph{methods}, limited work has focused on new datasets and benchmarks. In \cite{filos2019systematic, band2021benchmarking}, the authors introduced benchmarks for uncertainty quantification in Bayesian deep learning but only considered the image-based task of classifying diabetic retinopathy. Recently, a range of works by Hendrycks et al.~\cite{imagenet-r,imagenet-c,imagenet-ao} proposed a set of datasets based on ImageNet~\cite{imagenet} for evaluating model robustness to various types of distributional shifts. These datasets~--- ImageNet C, A, R, and O~--- include synthetically added noise, natural adversarial attacks, renderings, and previously unseen classes of objects.\footnote{ImageNet has only ``natural'' images; thus, renderings represent a shift in texture, but not content.} The release of WILDS, a collection of datasets containing real-world distributional shifts \cite{koh2020wilds}, similarly represents a significant step forward, but again mostly focuses on images. Finally, the MTNT dataset~\cite{michel2018mtnt}, which contains many examples of highly atypical usage of language, such as acronyms, profanity, emojis, slang, and code-switching, has been used at the Workshop on Machine Translation (WMT) robustness track. However, it has not been considered by the uncertainty community in the context of \emph{detecting} distributional shift. 
 
Unfortunately, with few exceptions, most work on uncertainty estimation and robustness has focused on developing new methods on small-scale tabular regression or image classification tasks, such as UCI, MNIST \cite{mnist}, Omniglot~\cite{omniglot}, SVHN \cite{svhn}, and CIFAR10/100 \cite{cifar}. Few works have been evaluated on the ImageNet variations A, R, C, and O, or WILDS. However, even evaluation on these datasets is limited, as they mainly focus on image classification, and sometimes text. In contrast, many tasks of practical interest have different modalities, such as tabular data (in medicine and finance), audio, text, or sensor data. Furthermore, these tasks are not always classification; they often involve regression and discrete or continuous structured prediction. Given the current state of the field, we aim to draw the attention of the community to the evaluation of uncertainty estimation and robustness to distributional shift on a realistic set of large-scale tasks across a range of modalities. This is necessary to meaningfully evaluate the plethora of methods for uncertainty quantification and improved robustness, and to accelerate the development of this area and safe ML in general.

In this work, we propose the \textbf{Shifts Dataset}\footnote{Data and example code are available at \url{https://github.com/yandex-research/shifts}} for evaluation of uncertainty estimates and robustness to distributional shift. This dataset consists of data taken directly from large-scale industrial sources and services where distributional shift is ubiquitous --- settings as close to ``in the wild'' as possible. The dataset is composed of three parts, with each corresponding to a particular data modality: \emph{tabular weather prediction} data provided by the Yandex Weather service; \emph{machine translation} data taken from the WMT robustness track and mined from Reddit, and annotated in-house by Yandex Translate; and, self-driving car (SDC) data provided by Yandex SDG, for the task of \emph{vehicle motion prediction}. All of these data modalities and tasks are affected by distributional shift and pose interesting challenges with respect to uncertainty estimation. This paper provides a detailed analysis of the data as well as baseline uncertainty estimation and robustness results using ensemble methods.
\section{Evaluation Paradigm, Metrics, and Baselines}\label{sec:metrics}

\paragraph{Paradigm} In most prior work, uncertainty estimation and robustness have been assessed separately. Robustness to distributional shift is usually assessed via metrics of predictive performance on a particular task --- given two (or more) evaluation sets, where one is considered matched to the training data and the other(s) shifted, models which have a smaller degradation in performance on the shifted data are considered more robust. 
Uncertainty quality is often assessed via the ability to classify whether an example came from the ``in-domain'' dataset or a shifted dataset using uncertainty estimates. Here, performance is assessed via Area under a Receiver-Operator Curve (ROC-AUC \%) or Precision-Recall curve (AUPR \%). While these evaluation paradigms are meaningful, we believe that they are two halves of a common whole. Instead, we consider the following paradigm:

\emph{As the degree of distributional shift increases, so does the likelihood that a model makes an error and the degree of this error. Models should yield uncertainty estimates which correlate with the degree of distributional shift, and therefore are indicative of the likelihood and the degree of the error.}

This paradigm is more general, as a model may be robust to certain examples of distributional shift and yield accurate, low uncertainty predictions. A model may also perform poorly and yield high estimates of uncertainty on underrepresented data matched to the training set. Thus, splitting a dataset into ``in-domain'' and ``out-of-distribution'' may not yield partitions on which a model strictly performs well or poorly, respectively. Instead, it is necessary to \emph{jointly} assess robustness and uncertainty estimation, in order to see whether uncertainty estimates at the level of a single prediction correlate well with the likelihood or degree of error. Thus, we view the problems of robustness and uncertainty estimation as having \emph{equal} importance --- models should be robust, but where they are not, they should yield high estimates of uncertainty, which enables risk-mitigating actions to be taken (e.g., transferring control of a self-driving vehicle to a human operator). 

We assume that at training or test time \emph{we do not know a priori} about alternative domains and whether or how our data is shifted. This setup aims to emulate real-world deployments in which the variation of conditions is vast and one can never collect enough data to cover all situations. It is for this reason we view robustness and uncertainty as equally important --- we assume that one can never be fully robust in all situations, and it is in these situations that high-quality uncertainty estimation is crucial. This is a strictly more challenging setting than one in which auxiliary information about the degree or nature of shift is available at training or test time (e.g., in WILDS \cite{koh2020wilds}).

We have constructed the Shifts Dataset within the context of this paradigm. Specifically, the dataset is constructed with the following attributes. First, the annotations of distributional shift are meant to be used for analysis rather than model construction.  Second, we have ``canonically'' partitioned the datasets such that the shifts are realistic but significant and to which it is challenging to be fully robust~--- this allows us to assess the quality of uncertainty estimates. However, the weather and motion prediction datasets \emph{can} be repartitioned in alternative ways which are different from our canonical partitioning, such that alternative robustness paradigms can be evaluated.\footnote{Tools for partitioning and repartitioning are provided in our GitHub repository.} 

\paragraph{Assessment Metrics} We jointly assess robustness and uncertainty via \emph{error-retention curves}~\cite{malinin-thesis,deepensemble2017} and \emph{F1-retention curves}. Given an error metric, such as MSE, error-retention curves trace the error over a dataset as a model's predictions are replaced by ground-truth labels in order of decreasing uncertainty. F1-retention curves depict the F1 for predicting whether a model's predictions are sufficiently good based on uncertainties (here we vary retention fraction, i.e., the fraction of data with the smallest uncertainty values that we classify as acceptable). Both assess the performance of a hybrid human-AI system, where a model can consult an oracle (human) for assistance in difficult situations. The area under this curve can be decreased (error retention) or increased (F1 retention) either by improving the predictive performance of the model, such that it has lower overall error, or by providing better estimates of uncertainty, such that more errorful predictions are rejected earlier. Thus, the area under the error (R-AUC) and F1 (F1-AUC) retention curves are metrics which jointly assess robustness to distributional shift and the quality of uncertainty estimates. We also quote F1 at 95\% retention rate. These metrics, detailed in \Cref{apn:metrics}, are used for all tasks in this paper.

\paragraph{Choice of Baselines} In this work we consider ensemble-based baselines. This was done for several reasons. First, ensemble-based approaches are a standard way to obtain \emph{both} improved robustness versus single models \emph{and} interpretable uncertainty estimates. Ensembles improve robustness because each model represents a functionally different explanation of the data. Thus, even if each individual model in an ensemble is subject to spurious correlations, the models will have different spurious correlations. When the models are combined, the effects of spurious correlations are cancelled out to a certain degree, improving generalization performance. Second, ensemble methods are easy to apply to any task of choice and require little adaptation. 
Uncertainty estimates can be obtained from measures of ensemble diversity --- if the predictions are diverse, then the ensemble members cannot agree on what the prediction should be and therefore are uncertain. Other than ensemble methods, there are few alternative approaches which are known to yield improved robustness \emph{and} interpretable uncertainty estimates, can be easily applied to a broad range of large-scale tasks without significant adaptation, and do not require information about the nature of distributional shift at training or test time. We leave the exploration of these alternatives and the development of new ones to future work. We do not examine robust learning methods, such as IRM~\cite{arjovsky2019invariant,koh2020wilds}, as they require domain annotations at training time and do  not yield uncertainty estimates.

\section{Tabular Weather Prediction}

Uncertainty estimation and robustness are essential in applications like medical diagnostics and financial forecasting. In such applications, data is often represented in a heterogeneous tabular form. While it is challenging to obtain either a large medical or financial dataset, the Yandex Weather service has provided a large tabular Weather Prediction dataset that features a natural tendency for the data distribution to drift over time (concept drift~\cite{concept-drift1,concept-drift2}). Furthermore, the locations are non-uniformly distributed around the globe based on population density, land coverage, and observation network development, which means that certain climate zones, like the Polar regions or the Sahara, are under-represented. We argue that this tabular Weather Prediction data represents similar challenges to the ones faced on financial and medical data, which is often combined from different hospitals/labs, consists of population-groups that are non-uniformly represented, and has a tendency to drift over time. Thus, the data we consider in this paper can be used as an appropriate benchmark for developing more robust models and uncertainty estimation methods for tabular data. 

\paragraph{Dataset} The Shifts Weather Prediction dataset contains a scalar regression and a multi-class classification tasks: at a particular latitude, longitude, and timestamp, one must predict either the air temperature at two meters above the ground or the precipitation class, given targets and features derived from weather station measurements and weather forecast models. The data consists of 10 million 129-column entries: 123 meteorological features, 4 meta-data attributes (time, latitude, longitude and climate type) and 2 targets~--- temperature (target for regression task) and precipitation class (target for classification task). The full feature list is provided in Section~\ref{apn:tab-features}. It is important to note that the features are highly heterogeneous, i.e., they are of different types and scales. The full data is distributed uniformly between September $1^{\text{st}}$, 2018, and September $1^{\text{st}}$, 2019, with samples across all climate types. This data is used by Yandex for real-time weather forecasts and represents a real industrial application.

To provide a standard benchmark that contains both in-domain and shifted data, we use a particular ``canonical partitioning''\footnote{Alternative partitionings can be made from the full data, but we use the canonical partitioning throughout this work, and also for the Shifts Challenge: \url{http://research.yandex.com/shifts}} of the full dataset into training, development (\texttt{dev}), and evaluation (\texttt{eval}) datasets. The training, in-domain \texttt{dev} (\texttt{dev\_in}) and in-domain \texttt{eval} (\texttt{eval\_in}) data consist of measurements made from September 2018 till April $8^{\text{th}}$, 2019 for climate types \textit{Tropical}, \textit{Dry}, and \textit{Mild Temperate}.  The shifted \texttt{dev} (\texttt{dev\_out}) data consists of measurements made from $8^{\text{th}}$ July till $1^{\text{st}}$ September 2019 for the climate type \textit{Snow}. 50K data points are sub-sampled for the climate type \textit{Snow} within this time range to construct \texttt{dev\_out}. The shifted \texttt{eval} data is further shifted than the out-of-domain development data; measurements are taken from $14^{\text{th}}$ May till $8^{\text{th}}$ July 2019, which is more distant in terms of the time of the year from the in-domain data compared to the out-of-domain development data. The climate types are restricted to \textit{Snow} and \textit{Polar}. Further details are provided in Appendix~\ref{apn:weather-description}. Details on use and support plan are in Appendix~\ref{apn:datasheet}.

\paragraph{Baselines} 

To build baseline models for the temperature prediction and precipitation classification tasks, we use the open-source CatBoost gradient boosting library that is known to achieve state-of-the-art results on tabular datasets~\cite{catboost}. We use an ensemble-based approach to uncertainty estimation for GBDT models~\cite{malinin2021gdbt}. For each task, an ensemble of ten models is trained on the training data with different random seeds. For regression, the models predict the mean and variance of the normal distribution by optimizing the negative log-likelihood. For classification, the models predict a probability distribution over precipitation classes. Training details are provided in Appendix~\ref{apn:weather-training}. Additional ensemble-based baselines and results are provided in Appendix~\ref{apn:tab-results}. 

We first compare the predictive performance of ensembles and single models; the results are shown in Table~\ref{tab:tabular_res_a}. Firstly, we observe that all models perform worse on shifted data than on in-domain data. For regression, we observe that the RMSE of the ensemble (on the \texttt{eval} set) is about two degrees Celsius. Note that ensembling allows us to reduce RMSE by about $0.16\degree$ compared to a single model. Similarly, ensembling reduces the MAE by approximately $0.12\degree$. For classification, ensembling boosts the accuracy by about 2\% and macro-averaged F1 by about 1\%. Note that the classification task is unbalanced (see Appendix~\ref{apn:weather-description} for details), so for better interpretability, we also report the accuracy and Macro-F1 of the classifier always predicting the majority class.

\begin{table}[htbp!]
\caption{Predictive performance for Weather Prediction. Mean $\pm\ \sigma$ is quoted for the single models.}
\centering
\resizebox{1\textwidth}{!}{
    \begin{tabular}{c|cccc|cccccc}
    \toprule
    & \multicolumn{4}{c|}{Regression} & \multicolumn{6}{c}{Classification} \\ \midrule
  \multirow{2}*{Data} & \multicolumn{2}{c}{RMSE $\downarrow$} & \multicolumn{2}{c|}{MAE $\downarrow$} & \multicolumn{3}{c}{Accuracy (\%) $\uparrow$} & \multicolumn{3}{c}{Macro F1 (\%) $\uparrow$}  \\ 
  & Single & Ens & Single & Ens & Maj. & Single & Ens & Maj. & Single & Ens \\
  \midrule
  \texttt{dev-in} & $1.59_{\pm 0.00}$ & 1.51 & $1.18_{\pm 0.00}$ & 1.11 & 37.9 & $67.0_{\pm 0.075}$ & 68.5 & 17.2 & $42.2_{\pm 0.01}$ & 42.3 \\
  \texttt{dev-out} & $2.30_{\pm 0.01}$ & 2.12 & $1.75_{\pm 0.01}$ & 1.61 & 35.7 & $47.5_{\pm 0.249}$ & 50.3 & 19.4 & $20.2_{\pm 0.01}$ & 21.3 \\
\texttt{dev} & $1.98_{\pm 0.01}$ & 1.84 & $1.47_{\pm 0.01}$ & 1.36 & 36.8 & $57.2_{\pm 0.117}$ & 59.4 &  17.2 & $36.8_{\pm 0.01}$ & 37.2 \\
\midrule
 \texttt{eval-in} & $1.60_{\pm 0.00}$ & 1.52 & $1.19_{\pm 0.00}$ & 1.11 & 37.9 & $66.7_{\pm 0.060}$ & 68.2 & 17.2 & $42.9_{\pm 0.00}$ & 44.1   \\
 \texttt{eval-out} & $2.60_{\pm 0.03}$ & 2.37 & $1.91_{\pm 0.01}$ & 1.75 & 30.0 & $44.5_{\pm 0.184}$ & 46.7 & 17.4 & $21.5_{\pm 0.00}$ & 22.2 \\
\texttt{eval} & $2.16_{\pm 0.01}$ & 2.00 & $1.56_{\pm 0.01}$ & 1.44 & 33.9 & $55.5_{\pm 0.090}$ & 57.3 &  17.4 &  $34.4_{\pm 0.01}$ & 35.5 \\ 
\bottomrule
    \end{tabular}
}
    \label{tab:tabular_res_a}
\end{table}

We jointly evaluate the robustness and uncertainty estimates for ensembles and single models. For the regression task, we use the predicted variance as the uncertainty measure of a single model. For ensembles, we use the total variance (tvar) that is the sum of the variance of the predicted mean and the mean of the predicted variance~\cite{galthesis,malinin-thesis,malinin2021gdbt}. For the classification task, we use the entropy of the prediction as the uncertainty measure of a single model. For ensembles, we use the (negated) confidence. We measure the area under the error-retention and F1-retention curves as described in Appendices~\ref{apn:metrics} and~\ref{subsection:performance_metrics}. These two performance metrics are denoted as R-AUC and F1-AUC, respectively. A good uncertainty measure is expected to achieve low R-AUC and high F1-AUC. Additionally, we report the F1 score at a retention rate of 95\% of the most certain samples (F1@95\%). All these measures jointly assess the predictive performance and uncertainty quality.

\begin{table}[htbp!]
\caption{Retention performance for Weather Prediction. Mean $\pm\ \sigma$ is quoted for the single models.}
\centering
\resizebox{1\textwidth}{!}{
    \begin{tabular}{c|c|cccccc}
    \toprule
  & \multirow{2}*{Data} &\multicolumn{2}{c}{\text{R-AUC} $\downarrow$} & \multicolumn{2}{c}{\text{F1-AUC (\%)} $\uparrow$} & \multicolumn{2}{c}{\text{F1@}$95$\% $\uparrow$}  \\
  &  & Single & Ens & Single & Ens & Single & Ens \\
   \midrule
\multirow{2}*{Regression} & \texttt{dev} & $1.894_{\pm 0.017}$ & 1.227 & $44.35_{\pm 0.2}$ & 52.20 & $62.72_{\pm 0.1}$ & 65.83  \\ 
& \texttt{eval}  & $2.320_{\pm 0.063}$ & 1.335 & $43.41_{\pm 0.1}$ & 52.36 & $61.89_{\pm 0.1}$ & 64.72  \\ \midrule
\multirow{2}*{Classification} & \texttt{dev} & $0.1666_{\pm 0.001}$ & 0.1522 & $57.72_{\pm 0.1}$ & 59.07 & $73.04_{\pm 0.1}$ &  74.86 \\ 
& \texttt{eval}  & $0.1799_{\pm 0.001}$ & 0.1640 & $56.25_{\pm 0.1}$ & 58.22 & $71.56_{\pm 0.1}$ & 73.17  \\
\bottomrule
    \end{tabular}
}
    \label{tab:tabular_res_b}
\end{table}

\begin{figure}[t]
     \centering
     \begin{subfigure}[b]{0.245\textwidth}
         \centering
         \includegraphics[width=\textwidth]{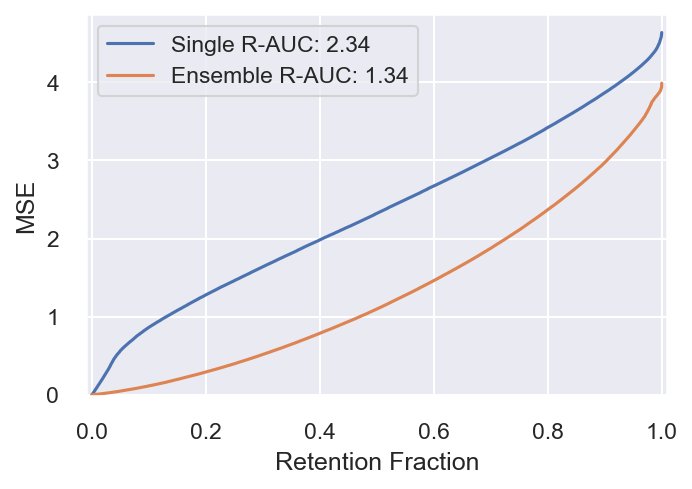}
         \caption{Regression, MSE.}
         \label{fig:eval_mse}
     \end{subfigure}
     \begin{subfigure}[b]{0.245\textwidth}
         \centering
         \includegraphics[width=\textwidth]{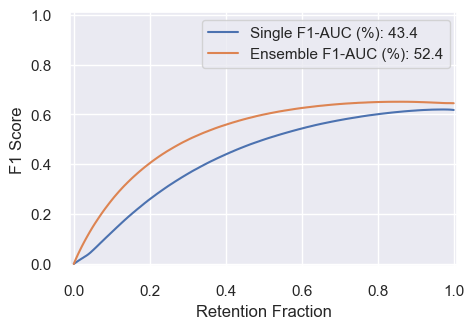}
         \caption{Regression, F1.}
         \label{fig:eval_f1}
     \end{subfigure} 
    \begin{subfigure}[b]{0.245\textwidth}
         \centering
         \includegraphics[width=\textwidth]{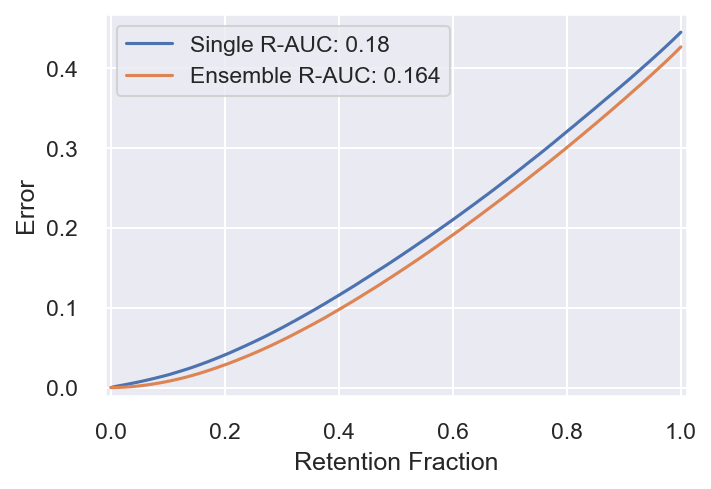}
         \caption{Classification, error.}
         \label{fig:eval_mse_class}
     \end{subfigure}
     \begin{subfigure}[b]{0.245\textwidth}
         \centering
         \includegraphics[width=\textwidth]{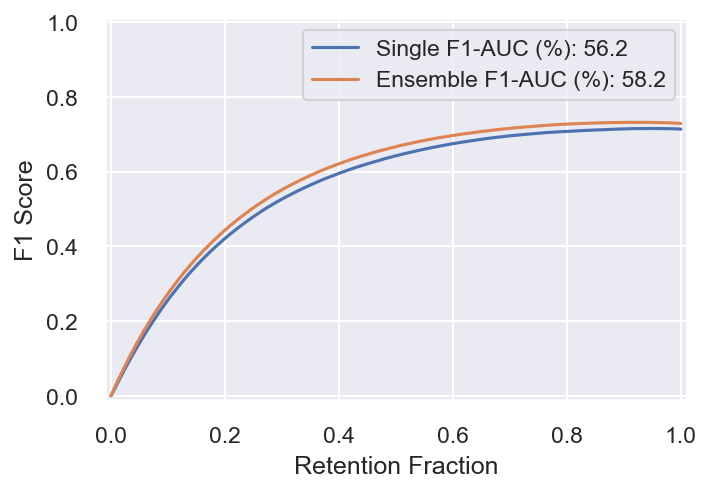}
         \caption{Classification, F1.}
         \label{fig:eval_f1_class}
     \end{subfigure} 
        \caption{Retention curves with CatBoost on \texttt{eval} for the Weather Prediction dataset.}
        \label{fig:weather_retention}
\end{figure}

The results are shown in Table~\ref{tab:tabular_res_b}. Here, as expected, ensembles significantly outperform single models. This observation is consistent over all considered evaluation measures. The associated retention curves are provided in Figure~\ref{fig:weather_retention} for \texttt{eval} and Figure~\ref{fig-apn:weather_retention} in \Cref{apn:tab} for \texttt{dev}. 


Finally, we conduct a comparison of different uncertainty measures. For this, we measure F1-AUC discussed above and ROC-AUC that evaluates uncertainty-based out-of-distribution (OOD) data detection. The results are shown in Table~\ref{tab:tabular_res_c}. In this experiment, we do not evaluate single models. For regression, we consider the following uncertainty measures: total variance (tvar) discussed above that is a measure of \emph{total uncertainty}, variance of the mean predictions across the ensemble models (varm) and the expected pairwise KL-divergence (EPKL) that are measures of \emph{knowledge uncertainty}. The results show that uncertainty measures that capture knowledge uncertainty perform best at OOD detection, as suggested by the high ROC-AUC values, while the measure of total uncertainty performs best for detecting errors (F1-AUC). Thus, as expected, the choice of a metric to use depends heavily on the task. Among measures of knowledge uncertainty, EPKL has better performance. For classification, the measures of total uncertainty are the negative confidence (Conf) and the entropy of the average prediction (Entropy). The measures of knowledge uncertainty are mutual information (MI), EPKL, and reverse mutual information (RMI). Similar to regression, uncertainty measures that capture knowledge uncertainty are better in terms of ROC-AUC. Among them, reverse mutual information performs best. The measures of total uncertainty are better for F1-AUC, and the best results are achieved with negative confidence.

\begin{table}[t]
\caption{Comparing uncertainty measures of CatBoost ensembles for Weather Prediction.}
\centering
\resizebox{1\textwidth}{!}{
    \begin{tabular}{c|l|ccc|ccccc}
    \toprule
  \multirow{3}*{Data} & & \multicolumn{3}{c|}{Regression} & \multicolumn{5}{c}{Classification} \\
    &  & \multicolumn{1}{c}{Total Unc.} & \multicolumn{2}{c|}{Knowledge Unc.} & \multicolumn{2}{c}{Total Unc.} & \multicolumn{3}{c}{Knowledge Unc.} \\ 
   & & \multicolumn{1}{c}{tvar} & \multicolumn{1}{c}{varm} & \multicolumn{1}{c|}{EPKL} &  \multicolumn{1}{c}{Conf} & \multicolumn{1}{c}{Entropy} & \multicolumn{1}{c}{MI} & \multicolumn{1}{c}{EPKL} & \multicolumn{1}{c}{RMI}\\
   \midrule
\multirow{2}*{\texttt{dev}}  
   & F1-AUC (\%) $\uparrow$ & \textbf{52.20} & 50.12 & 50.51 & \textbf{59.07} & 58.86 &  57.72 & 57.69& 57.66 \\
   & ROC-AUC (\%) $\uparrow$ & 62.96 & 82.31 & \textbf{85.29}  & 63.98 & 65.00 &  83.75 & 83.96 & \textbf{84.12} \\
   \midrule
\multirow{2}*{\texttt{eval}}  
   & F1-AUC (\%) $\uparrow$ & \textbf{52.36} & 49.81 & 50.40  & \textbf{58.22} &  57.89 & 56.99   & 56.96& 56.93 \\
   & ROC-AUC (\%) $\uparrow$ & 65.99 & 78.32 & \textbf{79.90} & 66.20 & 66.76 & 83.44  &83.59 &  \textbf{83.68}\\
   \bottomrule
    \end{tabular}
}
    \label{tab:tabular_res_c}
\end{table}

\section{Machine Translation}\label{sec:nmt}

As part of the Shifts Dataset we examine the task of machine translation for the text modality. Translation services, such as Google Translate or Yandex Translate, often encounter atypical and unusual use of language in their translation queries. This typically includes slang, profanities, poor grammar, orthography and punctuation, as well as emojis. This poses a challenge to modern translation systems, which are typically trained on corpora with a more ``standard'' use of language. Therefore, it is important for models to both be robust to atypical language use to provide high-quality translations, as well as to indicate when they are unable to provide a quality translation.

Translation is inherently a \emph{structured prediction} task, as there are dependencies between the tokens in the output sequence. Often we must make assumptions about the form of these dependencies; for example, most modern translation systems are left-to-right autoregressive. However, we could consider conditionally independent predictions or other factorization orders. The nature of these assumptions makes it challenging to obtain a theoretically sound measure of uncertainty. Only recently has work been done on developing principled uncertainty measures for structured prediction~\cite{notinprincipled, fomicheva2020unsupervised, yarin-mt-uncertainty, malinin2021structured, back-uncertainty}. Nevertheless, this remains an unsolved task and a fruitful area for research.

\paragraph{Dataset} The dataset contains training, development (\texttt{dev}) and evaluation (\texttt{eval}) data, where each set consists of pairs of source and target sentences in English and Russian, respectively. As most production Neural Machine Translation (NMT) systems are built using a variety of general purpose corpora, we use the freely available WMT`20 En-Ru corpus as training data. This dataset primarily focuses on parliamentary and news data that is, for the most part, grammatically and orthographically correct with formal language use. The \texttt{dev} and \texttt{eval} datasets consist of an ``in-domain'' partition matched to the training data, and an ``out-of-distribution'' or shifted partition, which contains examples of atypical language usage. The in-domain \texttt{dev} and \texttt{eval} sets are Newstest`19 En-Ru and a newly collected news corpus from GlobalVoices \cite{globalvoices}, respectively. For the shifted development data we use the Reddit corpus prepared for the WMT`19 robustness challenge~\cite{michel2018mtnt}. This data contains examples of slang, acronyms, lack of punctuation, poor orthography, concatenations, profanity, and poor grammar, among other forms of atypical language usage. This data is representative of the types of inputs that machine translation services find challenging. As Russian target annotations are not available, we pass the data through a two-stage process, where orthographic, grammatical, and punctuation mistakes are corrected, and the source-side English sentences are translated into Russian by expert in-house Yandex translators. The development set is constructed from the same 1400-sentence test-set used for the WMT`19 robustness challenge. For the evaluation set we use the open-source MTNT crawler which connects to the Reddit API to collect a further set of 3,000 English sentences from Reddit, which is similarly corrected and translated. The shifted \texttt{dev} and \texttt{eval} data are also annotated with 7 non-exclusive anomaly flags. Details on pre-processing, annotations and licenses are available in Appendix~\ref{apn:nmt-data}. Details on use and support plan are in Appendix~\ref{apn:datasheet}.

\paragraph{Metrics} To evaluate the performance of our models we will consider corpus-level BLEU~\cite{sacrebleu} and sentence-level GLEU~\cite{napoles2015gleu,napoles2016gleu,wu2016google}. As machine translation is a multi-modal task and translation systems often yield multiple translation hypothesis we will consider two GLEU-based metrics for evaluating translation quality. First is the \emph{expected GLEU} or \textbf{eGLEU} across all translation hypotheses, where each hypothesis is weighted by a \emph{confidence score}, and confidences across all hypotheses sum to one. Second is the maximum GLEU \textbf{maxGLEU} across all hypotheses in the beam. Details of these metrics can be found in Appendix~\ref{apn:nmt-metrics}. These metrics are then used to compute the error- and F1-retention curves which jointly assess uncertainty and robustness, as discussed in Appendix~\ref{apn:metrics}.

\paragraph{Baselines} In this work we considered an ensemble baseline based on~\cite{malinin2021structured}. Here, we use an ensemble of 3 Transformer-Big~\cite{attentionisallyouneed} models trained on the WMT`20 En-Ru corpus. Models were trained using a fork of FairSeq~\cite{fairseq} with a large-batch training set. Beam-Search decoding with a beam-width of 5 is used to obtain translation hypotheses. Hypotheses confidence weights are obtained by exponentiating the negative log-likelihood of each hypothesis and then normalizing across all hypotheses in the beam. Individual models in the ensemble are used as a single-model baseline.

Table~\ref{tab:translation_pred} presents the predictive performance on the \texttt{dev} and \texttt{eval} sets as well as on their in-domain and shifted subsets. There is a performance difference of nearly 10 BLEU and GLEU points between the in-domain news and shifted Reddit data, which shows the degradation in quality due to atypical language usage. The ensemble is able to outperform the individual models, which is expected. These results also show that BLEU correlates quite well with eGLEU. maxGLEU shows that significantly better performance is obtainable if we were better at ranking the hypotheses in the beam. 
\begin{table}[htbp!]
\caption{Predictive performance for Machine Translation. Mean $\pm\ \sigma$ is quoted for the single models.}
\centering
\small{
    \begin{tabular}{l|cccccc}
    \toprule
  \multirow{2}*{Data} & \multicolumn{2}{c}{BLEU $\uparrow$} & \multicolumn{2}{c}{eGLEU $\uparrow$} & \multicolumn{2}{c}{maxGLEU $\uparrow$} \\ 
   & Single & Ens & Single & Ens & Single & Ens \\
   \midrule
\texttt{\texttt{dev}-in} & 32.04$_{\pm 0.23}$&  32.73 &  34.45$_{\pm 0.10}$ &  35.09 & 41.08$_{\pm 0.09}$ & 42.00\\ 
\texttt{\texttt{dev}-out} & 20.65$_{\pm 0.16}$& 21.06 & 22.66$_{\pm 0.07}$  &  23.00  & 28.28$_{\pm 0.19}$ & 28.63\\ 
\texttt{\texttt{dev}} & 28.89$_{\pm 0.20}$  & 29.52 &  29.67$_{\pm 0.09}$ &  30.19 & 35.89$_{\pm 0.12}$ &  36.58 \\ 
\midrule
\texttt{\texttt{eval}-in} & 29.52$_{\pm 0.21 }$  &  30.08 & 30.39$_{\pm 0.10}$ & 30.88   & 36.19 $_{\pm 0.19}$ &  36.82 \\ 
\texttt{\texttt{eval}-out} & 21.00 $_{\pm 0.12}$&  21.54  & 23.19$_{\pm 0.07}$ &   23.60 &  29.35$_{\pm 0.11}$ & 29.88  \\ 
\texttt{\texttt{eval}} &  26.39$_{\pm 0.17}$ & 26.92  & 26.76$_{\pm 0.06}$ &  27.20  &   32.74$_{\pm 0.14 }$ &  33.31 \\ 
\bottomrule
    \end{tabular}
}
    \label{tab:translation_pred}
\end{table}

Having evaluated the baselines' predictive performance, we now jointly assess their uncertainty and robustness using the area under the error-retention curve (R-AUC), area under the F1-retention curve (F1-AUC) and F1 at 95\% retention, as detailed in Appendices~\ref{apn:metrics} and~\ref{apn:nmt-metrics}. Additionally, we evaluate in terms of \% ROC-AUC whether it is possible to discriminate between the in-domain data and the shifted data based on uncertainty estimates by the models. As the measure of uncertainty we use the negative log-likelihood, averaged across all 5 hypotheses. In the case of individual models, this is a measure of \emph{data} or \emph{aleatoric} uncertainty, and in the case of the ensemble, it is a measure of \emph{total uncertainty}~\cite{malinin2021structured}. Here, the ensemble consistently outperforms the single-model baseline. 

\begin{table}[htbp!]
\caption{Uncertainty and robustness for Machine Translation. Mean $\pm\ \sigma$ is quoted for single models.}
\centering
\resizebox{1\textwidth}{!}{
    \begin{tabular}{l|cccccccc}
    \toprule
  \multirow{2}*{Data} &\multicolumn{2}{c}{\text{R-AUC} $\downarrow$} & \multicolumn{2}{c}{\text{F1-AUC} $\uparrow$} & \multicolumn{2}{c}{\text{F1@}$95$\% $\uparrow$} & \multicolumn{2}{c}{ROC-AUC (\%) $\uparrow$} \\
   & Single & Ens & Single & Ens & Single & Ens & Single & Ens \\
   \midrule
\texttt{dev} & 33.22$_{\pm 0.48}$  &  32.87 &  0.43$_{\pm 0.00}$ & 0.44 & 0.42$_{\pm 0.01}$ &  0.43 &  68.90$_{\pm 0.28}$ & 69.30\\ 
\texttt{eval} &  34.80$_{\pm 0.06}$    & 34.57   &  0.37 $_{\pm 0.07}$ & 0.38 & 0.34 $_{\pm 0.03}$ &  0.36  & 79.18 $_{\pm 0.63}$   & 80.10 \\ 
\bottomrule
    \end{tabular}
}
    \label{tab:translation_unc}
\end{table}

\begin{figure}[htbp!]
     \centering
     \begin{subfigure}[b]{0.49\textwidth}
         \centering
         \includegraphics[width=\textwidth]{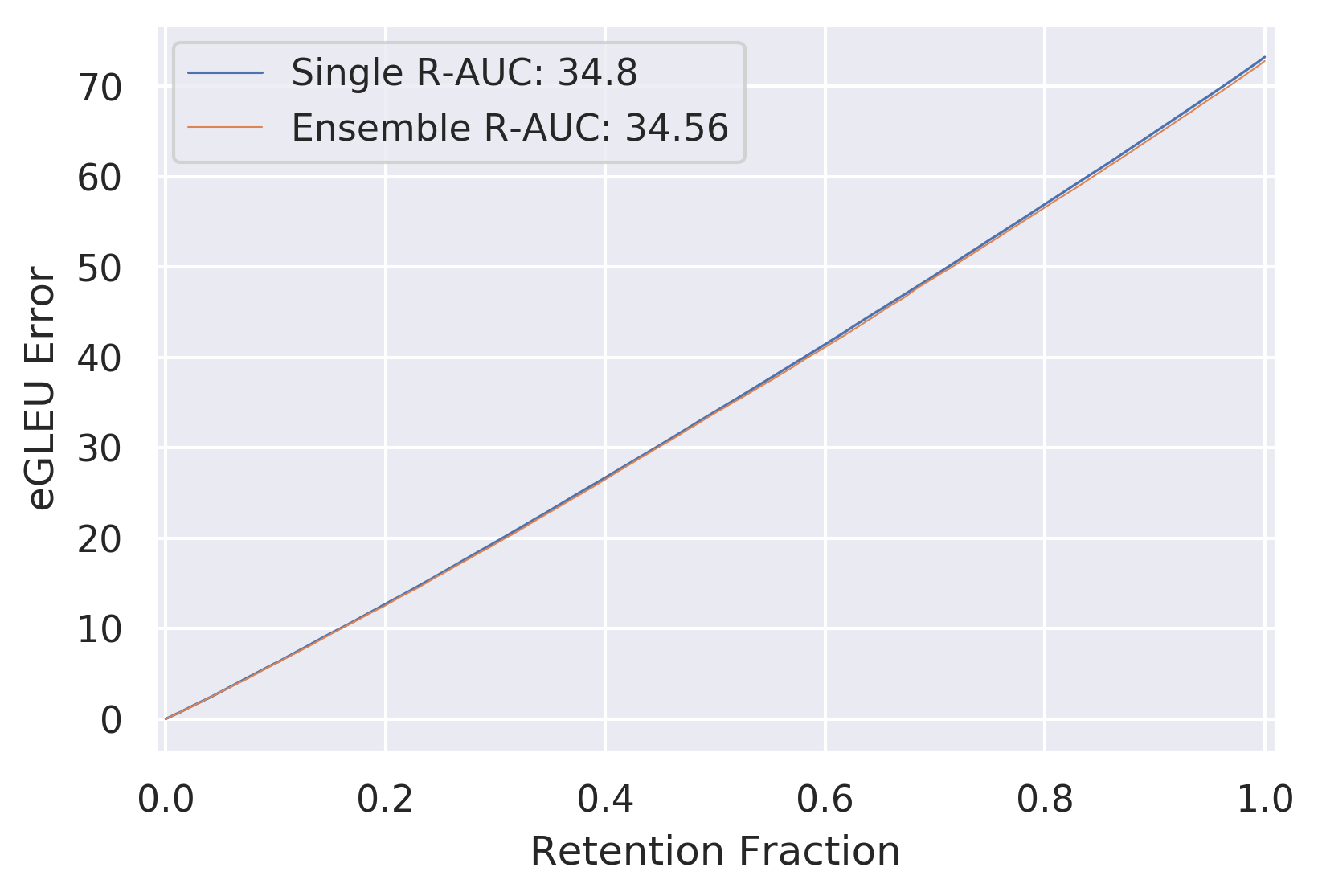}
         \caption{Error}
     \end{subfigure}
     \begin{subfigure}[b]{0.49\textwidth}
         \centering
         \includegraphics[width=\textwidth]{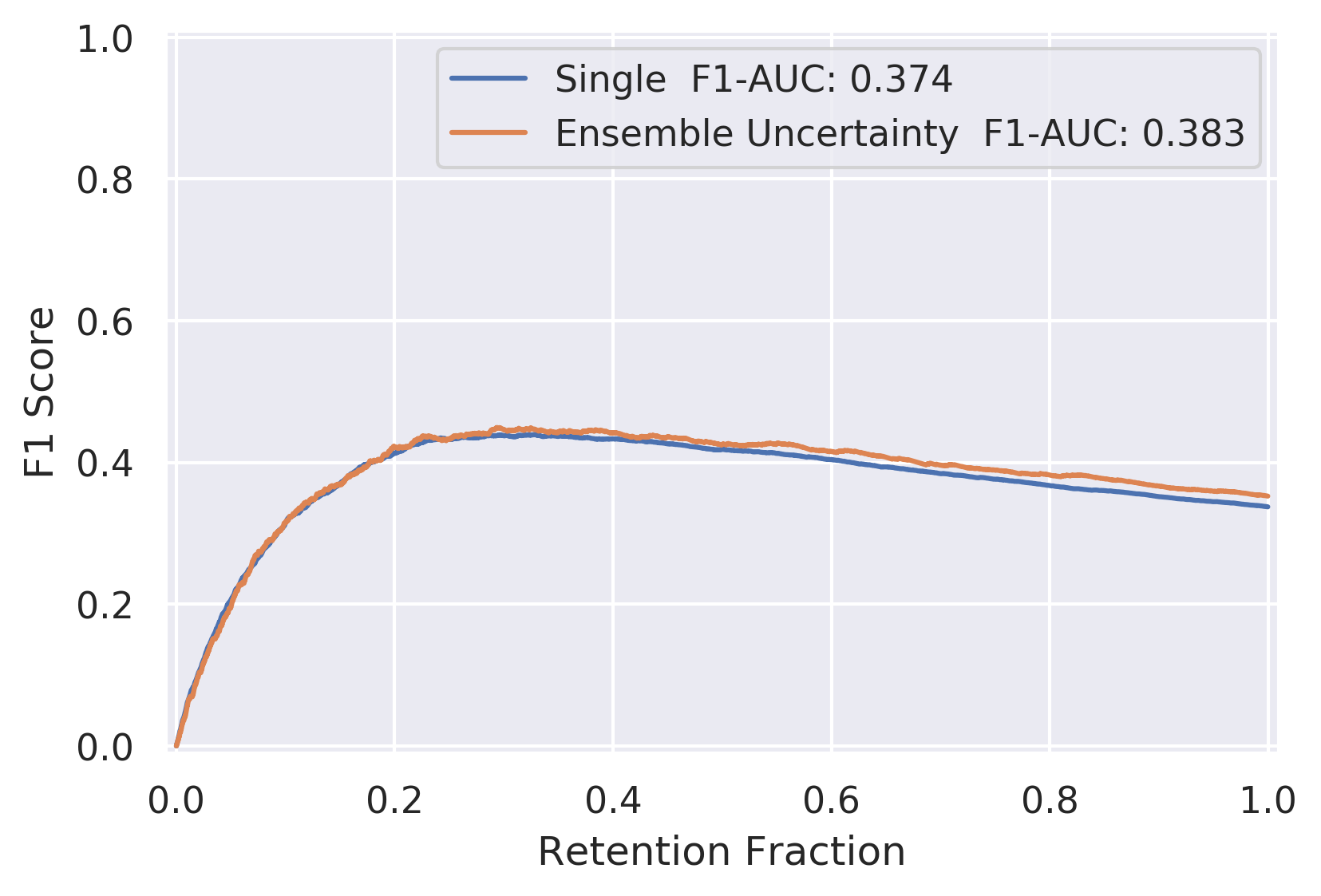}
         \caption{F1}
     \end{subfigure}
        \caption{Retention curves using eGLEU on eval data.}
        \label{fig:nmt_retention_curves}
\end{figure}

\section{Vehicle Motion Prediction}

We present the Shifts Vehicle Motion Prediction dataset to examine the implications of distributional shift in self-driving vehicles. The area of autonomous vehicle (AV) technology is highly relevant for uncertainty and robustness research, as the safety requirements and the risks associated with any errors are high. Furthermore, distributional shift is ubiquitous in the autonomous driving domain. During technology development, most self-driving companies concentrate their fleet in a limited number of locations and routes due to the large cost of operating in a new location. Therefore, fleets often face distributional shift when they begin operation in new locations. It is thus important to transfer as much knowledge as possible from the old locations to new ones. It is also critical for a planning model to recognize when this transferred knowledge is insufficient upon encountering unfamiliar data, which could risk unpredictable and unsafe behavior.\footnote{A case of knowledge, or epistemic uncertainty \cite{galthesis,malinin-thesis}.} Uncertainty quantification therefore has potentially life-critical application in this domain. For example, when the model's uncertainty is high, the vehicle can exercise extra caution or request assistance from a remote operator. 

Motion prediction is among the most important problems in the autonomous driving domain and has recently drawn significant attention from both academia and industry \cite{alahi2016social, lee2017desire, gupta2018social, chai2019multipath, casas2019spatially, cui2019multimodal, phan2020covernet, gao2020vectornet, biktairov2020prank, casas2020implicit, liang2020learning, liu2021multimodal}. It involves predicting the distribution over possible future states of agents around the self-driving car at a number of moments in time. A model of possible futures is needed because a self-driving vehicle needs a certain amount of time to change its speed, and sudden changes may be uncomfortable or even dangerous for its passengers. Therefore, in order to ensure a safe and comfortable ride, the motion planning module of a self-driving vehicle must reason about where other agents might end up in a few seconds to avoid planning a potential collision. This problem is complicated by the fact that \emph{the future is inherently uncertain}. For example, we cannot know the high-level navigational goals of other agents, or even their low-level tendency to turn right or left at a T-junction if they fail to indicate one way or another.\footnote{A case of data, or aleatoric uncertainty.} In order for the planning module to make the right decision, this uncertainty must be precisely quantified. Finally, motion prediction is also interesting because the predictions are both \emph{structured and continuous}. This poses further challenges in uncertainty estimation. Recently, ensemble-based uncertainty estimation for the related task of autonomous vehicle \emph{planning} was examined~\cite{filos2020can}, where a variance-based measure was proposed. However, there is still much potential for further development of informative measures of uncertainty in continuous structured prediction tasks such as motion prediction.

\paragraph{Dataset} The dataset for the Vehicle Motion Prediction task was collected by the Yandex Self-Driving Group (SDG) fleet and is the largest vehicle motion prediction dataset released to date, containing 600,000 scenes. These scenes span six locations, three seasons, three times of day, and four weather conditions.
Each scene includes information about the state of dynamic objects and an HD map. Each scene is 10 seconds long and is divided into 5 seconds of context features and 5 seconds of ground truth targets for prediction, separated by the time $T=0$. The goal is to predict the movement trajectory of vehicles at time $T \in \left(0, 5\right]$ based on the information available for time $T \in \left[-5, 0\right]$. The data contains training, development (\texttt{dev}) and evaluation (\texttt{eval}) sets. In order to study the effects of distributional shift, we partition the data such that the \texttt{dev} and \texttt{eval} sets have \emph{in-domain} partitions which match the location and precipitation type of the training set, and \emph{out-of-domain} or \emph{shifted} partitions which do not match the training data along one or more of those axes. As in the other Shifts tasks, we define a canonical partitioning which is used throughout benchmarking.\footnote{This partitioning is also the one used in the Shifts Challenge: \url{http://research.yandex.com/shifts}} The training set and in-domain partition of the \texttt{dev} and \texttt{eval} sets are taken from Moscow. Distributionally shifted \texttt{dev} data is taken from Skolkovo, Modiin, and Innopolis. Distributionally shifted \texttt{eval} data is taken from Tel Aviv and Ann Arbor. We also remove all cases of precipitation from the in-domain sets, while distributionally shifted datasets include precipitation. A full description of the dataset is available in Appendix~\ref{apn:sdc}, the support plan is detailed in Appendix~\ref{apn:datasheet}.

\paragraph{Metrics} Here we consider five different performance metrics --- minimum Average Displacement Error (minADE), minimum Final Displacement Error (minFDE),  confidence-weighed ADE and FDE, and corrected Negative Log-Likelihood (cNLL). cNLL is a new metric we introduce that is particilarly well-suited for assessing how models handle multi-modal situations. The minimum or weighting is done across up to 5 trajectories predicted by the baseline models. See \Cref{apn:sdc-metrics} for detailed explanations of the metrics.

\paragraph{Baselines} We consider two variants of Robust Imitative Planning (RIP)~\cite{filos2020can} as baselines. We use an ensemble of probabilistic models to stochastically generate multiple predictions for a given prediction request. Predictions are aggregated across ensemble members via a model averaging (MA) approach. We consider a simple RNN-based behavioral cloning network (RIP-BC) \cite{codevilla2018end} and autoregressive flow--based Deep Imitative Model (RIP-DIM) \cite{rhinehart2018deep} as backbone models. We adapt RIP to produce uncertainty estimates at two levels of granularity: per-trajectory and per--prediction request. Finally, we vary the number of ensemble members $K \in \{1, 3, 5\}$ and the uncertainty estimation method between Deep Ensembles \cite{deepensemble2017} and Dropout Ensembles \cite{Gal2016Dropout, smith2018understanding}. See Appendix~\ref{apn:sdc} for details on RIP, uncertainty estimation methods, backbone models, experimental setup, and full results. Additional results using Dropout Ensembles are provided in Appendix~\ref{apn:sdc-results}. 
\begin{table}[htbp!]
    \centering
    \caption{Predictive performance of BC \& DIM RIP on in-domain, shifted, and full \texttt{dev} \& \texttt{eval} data.}
    \label{table:sdc_performance}
\vspace{0.1in}
\resizebox{1\textwidth}{!}{%
    \begin{tabular}{@{}ll|ccc|ccc|ccc|ccc|ccc}
        \toprule
          \multirow{2}*{Dataset} &  \multirow{2}*{Model} & \multicolumn{3}{c|}{\text{cNLL} $\downarrow$} & \multicolumn{3}{c|}{\text{minADE} $\downarrow$} & \multicolumn{3}{c|}{\text{weightedADE} $\downarrow$} & \multicolumn{3}{c|}{ \text{minFDE} $\downarrow$} &  \multicolumn{3}{c}{\text{weightedFDE} $\downarrow$}\\
          & & In & Shifted & Full & In & Shifted & Full & In & Shifted & Full & In & Shifted & Full & In & Shifted & Full   \\
        \midrule
              \multirow{4}*{\texttt{Dev}}   
     & \text{BC, MA, K=1} & 59.64&  98.54& 64.29 & 0.818 & 0.960 & 0.835 & 1.088 & 1.245 & 1.107 & 1.718 & 2.113 & 1.765 & 2.368 & 2.777 & 2.417 \\
     & \text{BC, MA, K=5} & 56.86&  91.54& 61.01 & 0.765 & 0.887 & 0.779 & \bf{1.012} & \bf{1.133} & \bf{1.026} & 1.617 & 1.976 & 1.660 & \bf{2.210} & \bf{2.551} & \bf{2.251} \\
     & \text{DIM, MA, K=1} & \textbf{50.66}& 73.00 & \textbf{53.34} & 0.750 & 0.818 & 0.758 & 1.523 & 1.583 & 1.530 & 1.497 & 1.720 & 1.524 & 3.472 & 3.639 & 3.492 \\
     & \text{DIM, MA, K=5} &50.85 & \textbf{72.45} &  53.43& \textbf{0.719} & \textbf{0.786} & \textbf{0.727 }& 1.399 & 1.469 & 1.408 & \textbf{1.482} & \textbf{1.698} & \textbf{1.508} & 3.202 & 3.393 & 3.225 \\
     \midrule
  \multirow{4}*{\texttt{Eval}}
    & \text{BC, MA, K=1} & 60.20 & 98.82 & 67.93 & 0.829 & 1.084 & 0.880 & 1.104 & 1.407 & 1.164 & 1.733 & 2.420 & 1.870 & 2.394 & 3.197 & 2.555 \\
    & \text{BC, MA, K=5} & 57.75 & 95.00 & 65.20 & 0.777 & 1.014 & 0.824 & \bf{1.028} & \bf{1.299} & \bf{1.082} & 1.636 & 2.278 & 1.765 & \bf{2.238} & \bf{2.957} & \bf{2.382} \\
    & \text{DIM, MA, K=1} & \textbf{50.50} & \textbf{76.00} & \textbf{55.60} & 0.759 & 0.942 & 0.796 & 1.551 & 1.883 & 1.618 & 1.511 & \textbf{1.983} & 1.605 & 3.536 & 4.376 & 3.704 \\
    & \text{DIM, MA, K=5} & 51.19&  78.85& 56.73 & \textbf{0.728} & \textbf{0.918} & \textbf{0.766} & 1.424 & 1.754 & 1.490 & \textbf{1.493} & 2.000 & \textbf{1.595} & 3.256 & 4.093 & 3.424 \\
        \bottomrule
    \end{tabular}
}
\end{table}

%

Predictive performance results for the RIP variants are presented in \Cref{table:sdc_performance}. Performance is assessed on the in-distribution (In), distributionally shifted (Shifted), and combined (Full) \texttt{dev} and \texttt{eval} datasets. We observe that across all model configurations, performance on the shifted data is worse than that on the in-distribution data. We also observe that RIP-BC consistently outperforms RIP-DIM on the per-trajectory confidence weighted metrics (weightedADE and weightedFDE), and RIP (DIM) outperforms RIP (BC) on minADE and minFDE. This result might occur if DIM has higher predictive variance. In such a case, DIM might be more effective in modeling multimodality, and therefore would tend to produce at least one high accuracy trajectory on more scenes, improving performance on \texttt{min} aggregation metrics. This is supported by DIM models yielding the best cNLL, which is a metric particularly sensitive to correct treatment of multi-modal situations. In contrast, for ``obvious'' scenes, DIM might then produce unnecessarily complicated trajectories which would be reflected in poor performance on weightedADE. 
\begin{table}[htbp!]
\caption{Uncertainty and robustness performance for motion prediction. The error metric for computing the area under the F1 curve (F1-AUC) and F1 at 95\% retention rate (F1@95\%) is \textbf{cNLL}.}
    \label{table:sdc_uncertainty}
\centering
\vspace{0.1in}
\resizebox{1\textwidth}{!}{
    \begin{tabular}{lc|cc|cc|cc|cc|cc}
    \toprule
  \multirow{2}*{Data} & \multirow{1}*{Ensemble} &\multicolumn{2}{c|}{\text{R-AUC cNLL} $\downarrow$} &\multicolumn{2}{c|}{\text{R-AUC weightedADE} $\downarrow$} & \multicolumn{2}{c|}{\text{F1-AUC} (\%) $\uparrow$} & \multicolumn{2}{c|}{\text{F1@}$95$\% $\uparrow$} & \multicolumn{2}{c}{ROC-AUC (\%) $\uparrow$} \\
   & Size (K)& RIP-BC & RIP-DIM  & RIP-BC & RIP-DIM & RIP-BC & RIP-DIM& RIP-BC & RIP-DIM& RIP-BC & RIP-DIM \\
   \midrule
\multirow{2}*{\texttt{Dev}}  & 1 & 11.22 &  12.86& 0.268   & 0.419   & 65.1   &   63.8    &  89.3   &    87.4 & 51.0 &  \bf{51.8} \\
& 5 & \textbf{9.08}& 13.24 &  \bf{0.236}   &   0.376  &  \bf{65.2}   &  63.7   &  \bf{90.6}  &  89.7  &  49.2  & 51.4 \\
\midrule
\multirow{2}*{\texttt{Eval}}  & 1& 12.91 & 14.32 & 0.293   &   0.458  & 65.0   &  63.6   &  88.4   &  86.3   &  \bf{52.8}  &  51.8  \\ 
 & 5  & \textbf{10.57} & 15.16 &  \bf{0.258}  &  0.411  & \bf{65.1}    & 63.5   &  \bf{89.7}&  88.9   & 52.1  & 50.9 \\ 
\bottomrule
    \end{tabular}
}
\end{table}

\begin{figure}[htbp!]
     \centering
     \begin{subfigure}[b]{0.49\textwidth}
         \centering
         \includegraphics[width=\textwidth]{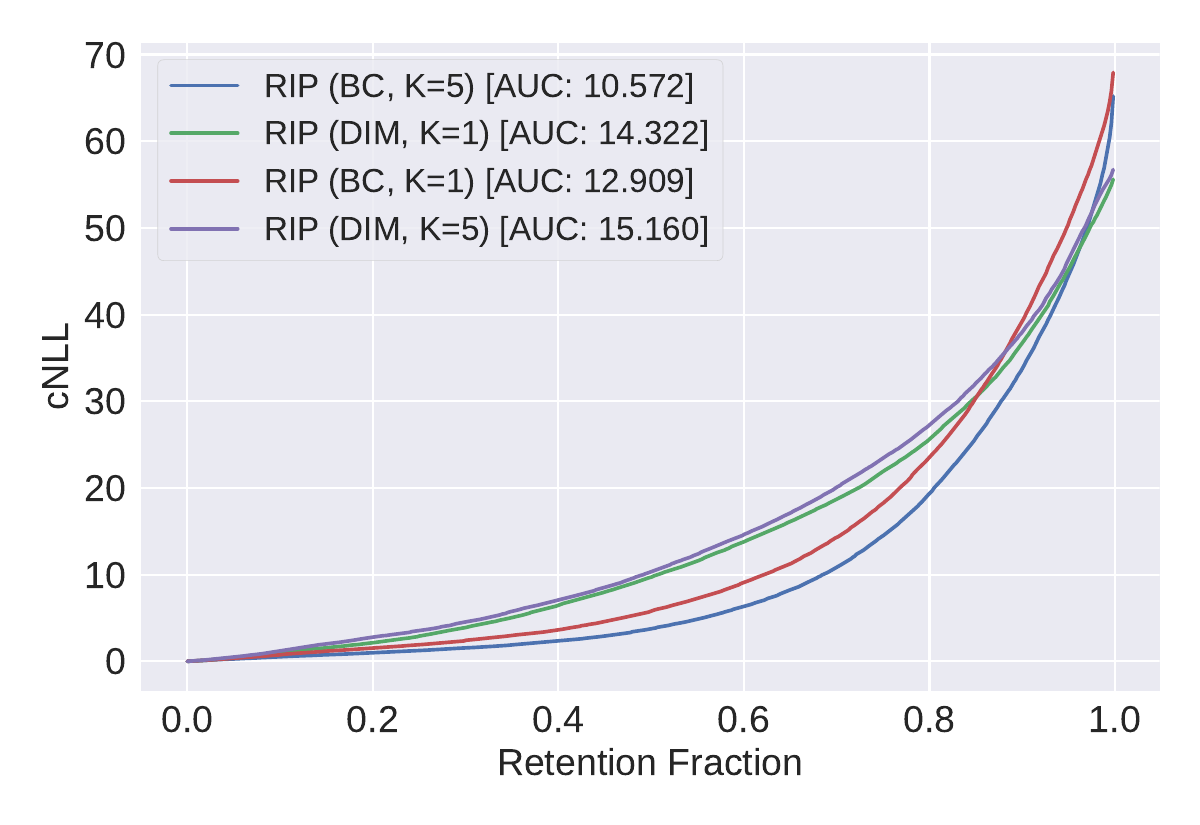}
         \caption{cNLL error-retention.}
         \label{fig:sdc_ret_eval}
     \end{subfigure}
     \hfill
     \begin{subfigure}[b]{0.49\textwidth}
         \centering
         \includegraphics[width=\textwidth]{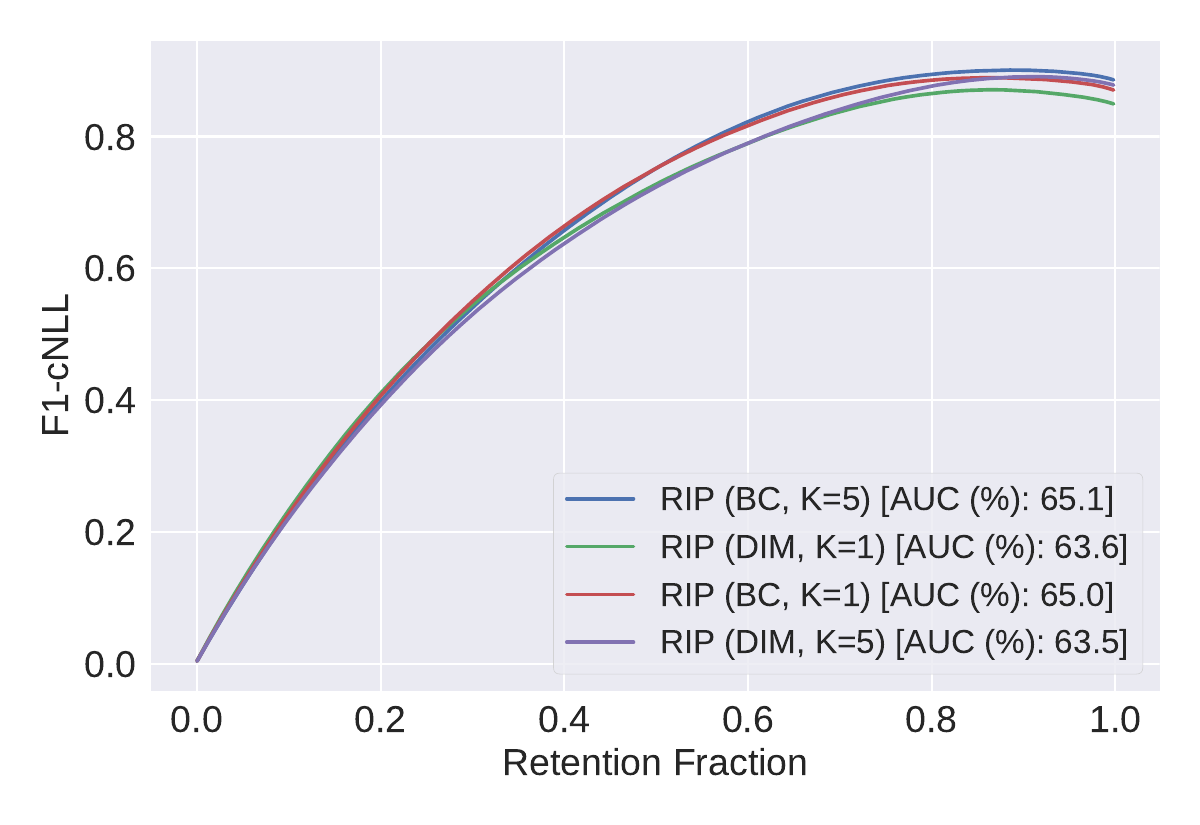}
         \caption{cNLL F1-retention.}
         \label{fig:sdc_fbeta_eval}
     \end{subfigure} 
        \caption{Retention curves for Vehicle Motion Prediction on full \texttt{eval} data.}
        \label{fig:sdc_retention}
\end{figure}

\cref{table:sdc_uncertainty} presents a joint evaluation of the uncertainty quantification and robustness of our baselines. We compute R-AUC with respect to cNLL and weightedADE, and the F1-AUC and F1$@95\%$ metrics with respect to the cNLL metric, as detailed in Appendices~\ref{apn:metrics} and ~\ref{apn:sdc-metrics}. We observe that an ensemble of RIP-BC models outperforms RIP-DIM on these metrics. These results strongly suggest that RIP-BC has more informative uncertainty estimates than RIP-DIM, because RIP-BC achieves better R-AUC cNLL despite having greater overall error in terms of cNLL (in addition to minADE and minFDE). Figure~\ref{fig:sdc_retention} depicts, for cNLL, error- and F1-retention curves on the full \texttt{eval} dataset which reflect the trends observed in Table~\ref{table:sdc_uncertainty}. Additionally, we find that across model configurations the per--prediction request uncertainty scores do not perform particularly well in detecting distribution shift (ROC-AUC). This may occur due to significant data uncertainty in all cases. Future work on detecting distributional shift on this dataset could, for example, inspect the distribution of log-likelihood scores on the in-distribution and shifted partitions in order to devise a metric for this task, aside from the uncertainty scores $U$ used for the retention analysis.

\section{Conclusion}

In this paper, we proposed the \textbf{Shifts Dataset}: a large, standardized dataset for evaluation of uncertainty estimates and robustness to realistic, curated distributional shift. The dataset --- sourced from industrial services --- is composed of three tasks, with each corresponding to a particular data modality: \emph{tabular weather prediction}, \emph{machine translation}, and self-driving car (SDC) \emph{vehicle motion prediction}. This paper describes this data and provides baseline results using ensemble methods. Given the current state of the field, where most methods are developed on small-scale classification tasks, we aim to draw the attention of the community to the evaluation of uncertainty estimation and robustness to distributional shift on large-scale industrial tasks across multiple modalities. We believe this work is a necessary step towards meaningful evaluation of uncertainty quantification methods, and hope for it to accelerate the development of this area and safe ML in general.

\newpage
\begin{ack}

We would like to thank Yandex for providing the data and resources necessary in benchmark creation. We thank Intel and the Turing Institute for funding the work of the OATML Group on this project. Finally, we thank Cambridge University Press and Cambridge Assessment for funding the work of the CUED Speech Group.

\end{ack}

\bibliographystyle{IEEEbib}
\bibliography{bibliography}

\newpage
\appendix
\appendixpage
\section{Assessment Metrics}\label{apn:metrics}

As discussed in \Cref{sec:metrics}, in this work we consider robustness and uncertainty estimation to be two equally important factors in assessing the reliability of a model. We assume that as the degree of distributional shift increases, so should a model's errors; in other words, a model's uncertainty estimates should be correlated with the degree of its error. This informs our choice of assessment metrics, which must \emph{jointly} assess robustness and uncertainty estimation. 

One standard approach to jointly assess robustness and uncertainty are \emph{error-retention curves}~\cite{malinin-thesis,deepensemble2017}, which plot a model's mean error over a dataset, as measured using a metric such as error-rate, MSE, eGLEU, cNLL, etc., with respect to the fraction of the dataset for which the model's predictions are used. These retention curves are traced by replacing a model's predictions with ground-truth labels obtained from an oracle in order of decreasing uncertainty, thereby decreasing error. Ideally, a model's uncertainty is correlated with its error, and therefore the most errorful predictions would be replaced first, which would yield the greatest reduction in mean error as more predictions are replaced. This represents a hybrid human-AI scenario, where a model can consult an oracle (human) for assistance in difficult situations and obtain from the oracle a perfect prediction on those examples. 

\begin{figure}[htbp!]
     \centering
     \begin{subfigure}[b]{0.32\textwidth}
         \centering
         \includegraphics[width=\textwidth]{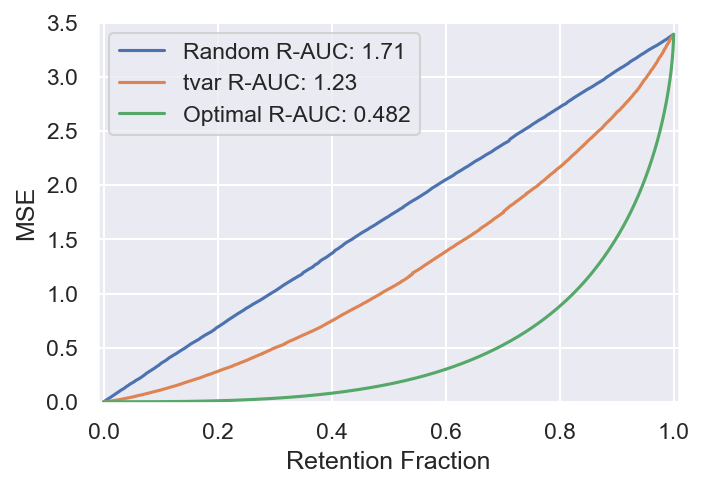}
         \caption{Weather Prediction}
     \end{subfigure}
     \begin{subfigure}[b]{0.32\textwidth}
         \centering
         \includegraphics[width=\textwidth]{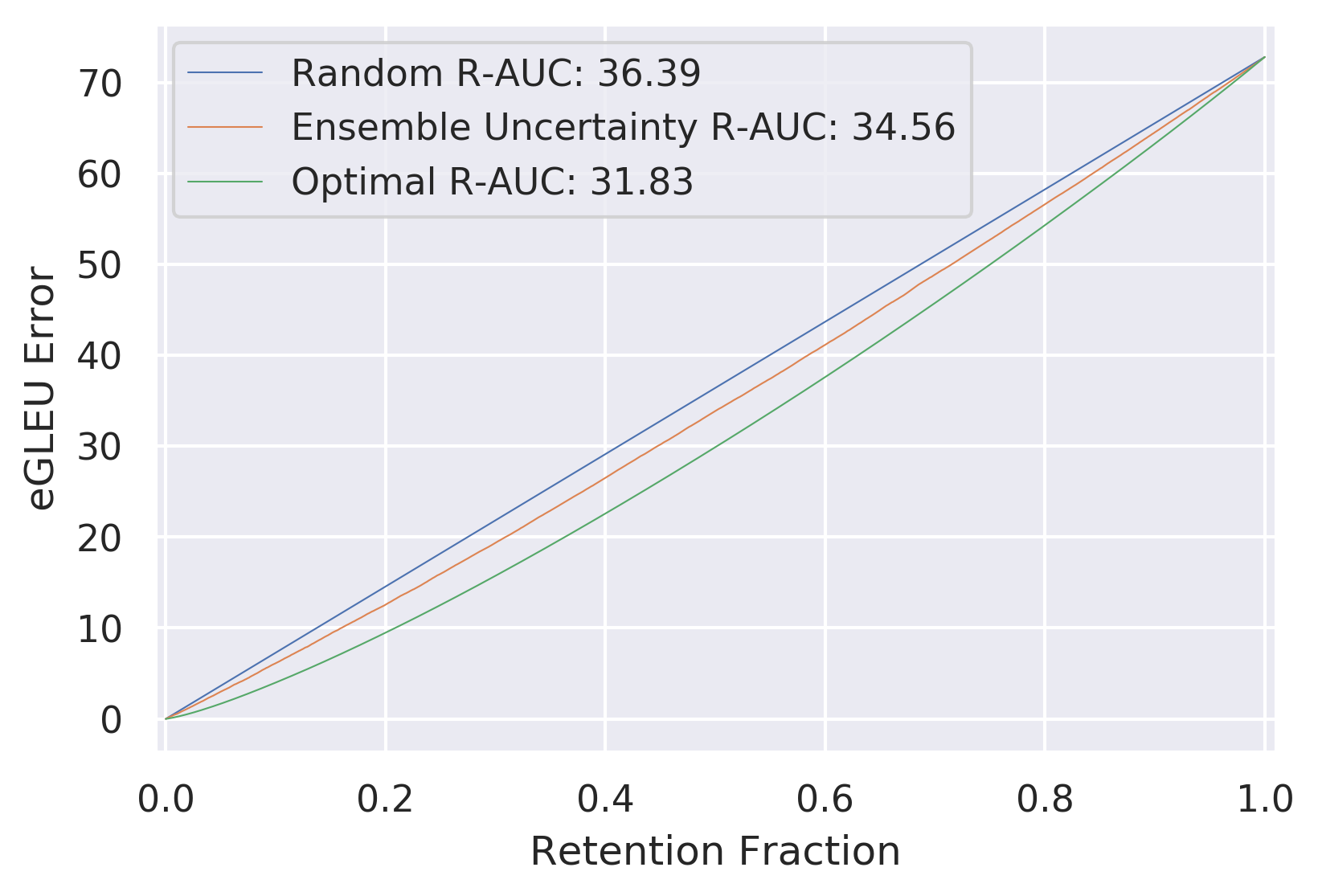}
         \caption{Machine Translation}
     \end{subfigure}
     \begin{subfigure}[b]{0.32\textwidth}
         \centering
         \includegraphics[width=\textwidth]{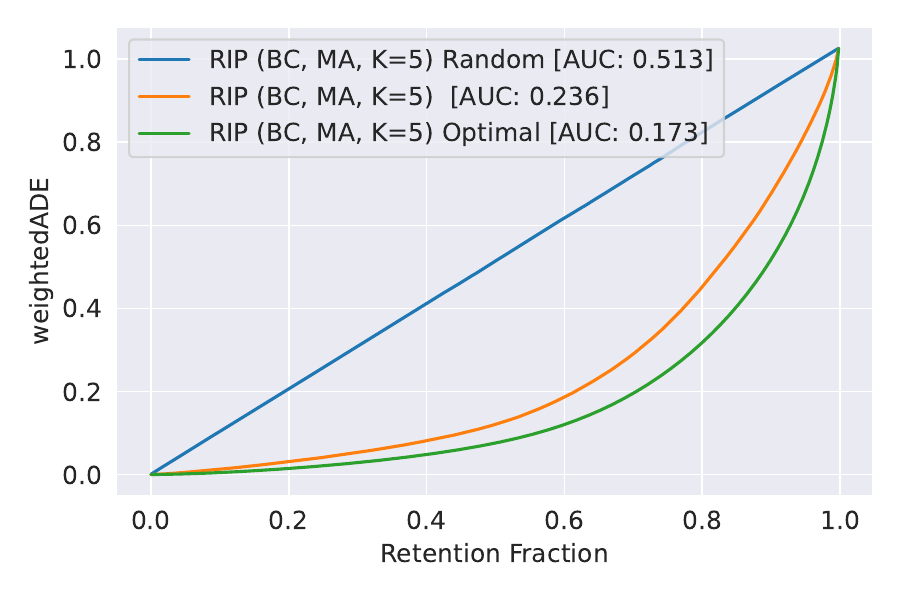}
         \caption{Vehicle Motion Prediction}
     \end{subfigure}
     \caption{Example error retention curves for the three tasks of the Shifts Dataset.}\label{fig:example_retention_curves}
\end{figure}

The area under the retention curve (R-AUC) is a metric for jointly assessing robustness to distributional shift and the quality of the uncertainty estimates. R-AUC can be reduced either by improving the predictions of the model, such that it has lower overall error at any given retention rate, or by providing estimates of uncertainty which better correlate with error, such that the most incorrect predictions are rejected first. It is important that the dataset in question contains both a subset ``matched'' to the training data, and a distributionally shifted subset.
Figure~\ref{fig:example_retention_curves} provides example retention curves for the three tasks of the Shifts Dataset. In each figure, in addition to the uncertainty-based ranking, we included curves which represent ``random'' ranking, where uncertainty estimates are entirely non-informative, and ``optimal'' ranking, where uncertainty estimates perfectly correlate with error. These represent the lower and upper bounds on R-AUC performance as a function of uncertainty quality.

While clearly interpretable and intuitive, one concern that can be raised regarding error-retention curves is that they can be more sensitive to predictive performance than to the quality of uncertainty estimates, which can be seen in Figure~\ref{fig:example_retention_curves}b.  This occurs on tasks where most errors have similar magnitude. Furthermore, for regression tasks, retention curves are dominated by noise in the targets (aleatoric uncertainty) at low retention fractions, when most systematic errors have already been detected. Therefore, in this work we propose another metric which jointly assesses robustness and uncertainty estimation.

First, we introduce the notion of an ``acceptable prediction'', which is a prediction whose error is acceptably small. This concept is natural for tasks with a non-binary notion of error, e.g., regression problems. For classification tasks, where predictions are already either correct or incorrect (acceptable/non-acceptable), this concept can be introduced by considering different levels of risk for different misclassifications. Formally, we say that a prediction is acceptable if an appropriate metric of error or risk $\mathcal{E}$ is below a \emph{fixed} task-dependent error threshold $T_e$. For example, if temperature is predicted to within a degree of the ground truth, then it is acceptable. This allows us to mitigate the issue of errors having similar magnitudes. This is expressed using via an indicator function as follows:
\begin{empheq}{align}
\mathcal{A}_{T_e}(\bm{x}) = \ 
\begin{cases} 
1,\  \mathcal{E}(\bm{x}) \leq T_e \\
0,\  \mathcal{E}(\bm{x}) > T_e \\
\end{cases}
\end{empheq}

For a given dataset $D$ and model, we first set an error threshold and determine which predictions are acceptable -- this yields a set of ``ground-truth'' acceptability labels $\mathcal{A}_{i=1}^N$. We can now use these acceptability labels to assess whether the model's \emph{estimates of uncertainty} $\mathcal{U}(\bm{x})$ can be used to indicate whether a prediction is acceptable. If the uncertainty score is greater than a threshold $T_u$, then we consider the prediction to be poor, if the uncertainty score is lower than this threshold, the prediction is considered to be acceptable.
\begin{empheq}{align}
\mathcal{\hat A}_{T_u}(\bm{x}) = \ 
\begin{cases} 
1,\  \mathcal{U}(\bm{x}) \leq T_u \\
0,\  \mathcal{U}(\bm{x}) > T_u \\
\end{cases}
\end{empheq}

Next, given the true acceptability labels $\{\mathcal{A}_{T_e}(\bm{x}_i)\}_{i=1}^N$ and the threshold-conditional indicators $\{\mathcal{\hat A}_{T_u}(\bm{x})\}_{i=1}^N$ we sweep through all uncertainty scores in a dataset $\{\mathcal{U}(\bm{x}_i)\}_{i=1}^N$ in decreasing order and use them as thresholds to F1 for classifying whether a prediction is actually acceptable or not based on the uncertainty. Formally, this is done as follows:
\begin{empheq}{align}
P_i  =&\ \frac{\sum_{j=1}^N \mathcal{A}_{T_e}(\bm{x}_j) \cdot \mathcal{\hat A}_{\mathcal{U}_i}(\bm{x}_j)}{N-i},\
R_i  =\ \frac{\sum_{j=1}^N \mathcal{A}_{T_e}(\bm{x}_j) \cdot \mathcal{\hat A}_{\mathcal{U}_i}(\bm{x}_j)}{\sum_{j=1}^N \mathcal{A}_{T_e}(\bm{x}_j)},\
\text{F1}_i =\ \frac{2\cdot P_i \cdot R_i}{P_i+R_i}
\end{empheq}
where we use $N-i$ because we sort uncertainties from largest ($\mathcal{U}_1$) to smallest ($\mathcal{U}_N$). We then plot $\{\text{F1}_i\}_{i=1}^N$ against $1 - \frac{i}{N}$, i.e., the fraction of data we are classifying as acceptable, which we refer to as the retention fraction. This yields the following curves for the three Shifts tasks:
\begin{figure}[htbp!]
     \centering
     \begin{subfigure}[b]{0.32\textwidth}
         \centering
         \includegraphics[width=\textwidth]{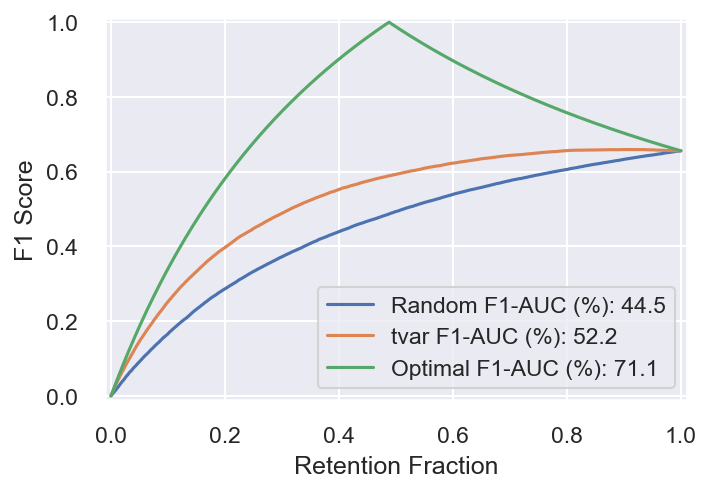}
         \caption{Weather Prediction}
     \end{subfigure}
     \begin{subfigure}[b]{0.32\textwidth}
         \centering
         \includegraphics[width=\textwidth]{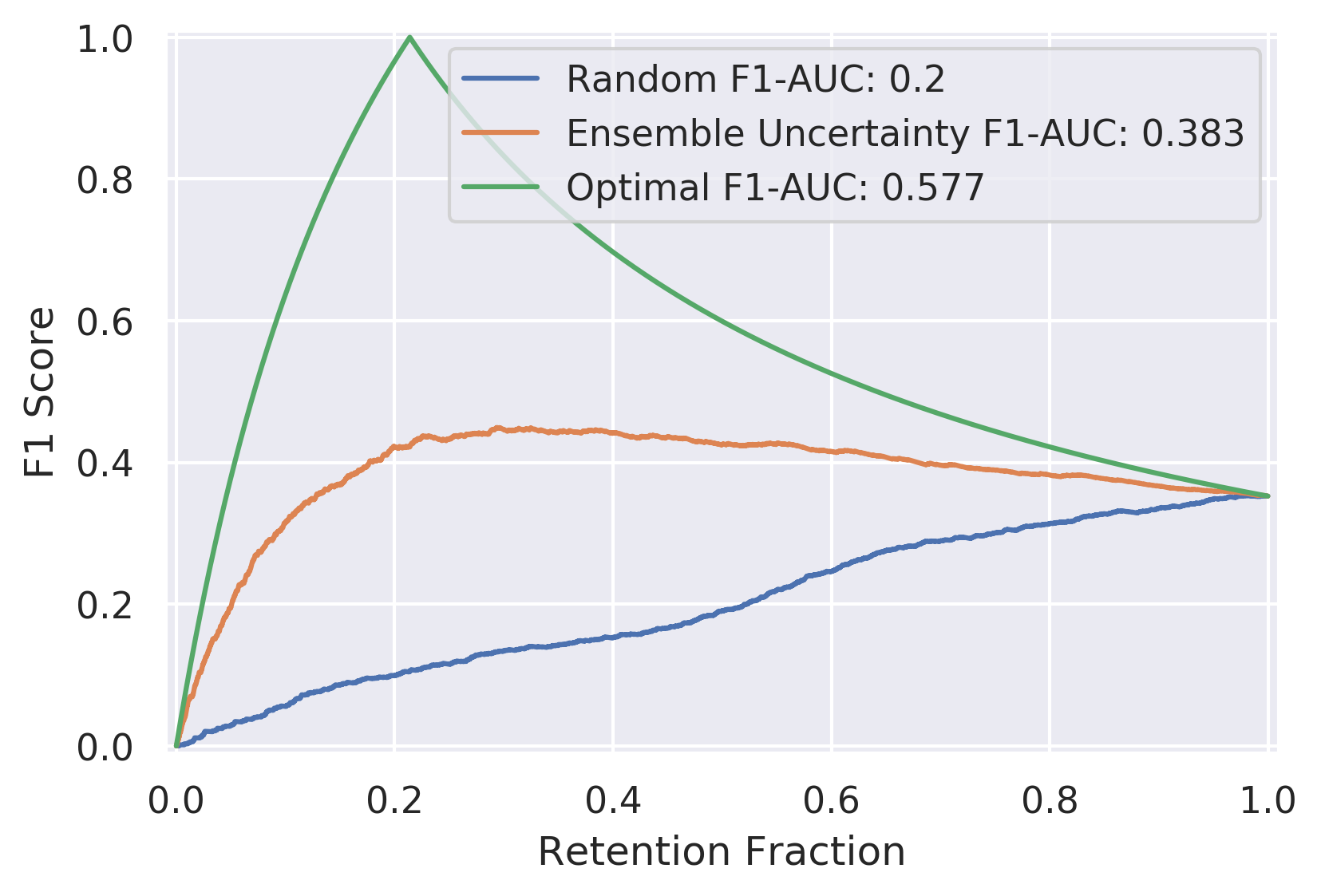}
         \caption{Machine Translation}
     \end{subfigure}
     \begin{subfigure}[b]{0.32\textwidth}
         \centering
         \includegraphics[width=\textwidth]{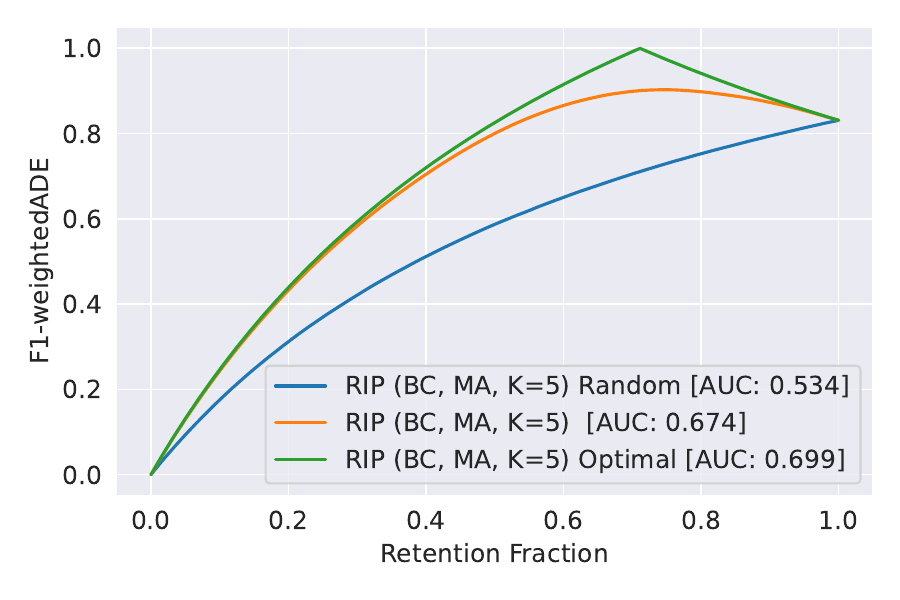}
         \caption{Motion Prediction}
     \end{subfigure}
     \caption{Examples of F1-Retention curves for the three tasks of the Shifts Dataset.} \label{fig:example_F1_retention_curves}
\end{figure}

Here we plot the uncertainty-based F1-retention curves for all datasets. On each figure, we plot both the uncertainty-derived curves as well as the ``random'' and ``optimal'' baselines, where uncertainties are either completely uncorrelated or perfectly correlated with errors, respectively. Better models have a higher area under this F1-retention curve (F1-AUC). The predictive performance of the model defines the starting point at 100\% retention -- better models start higher. Thus, area under the F1 curve can be increased by having a model which yield better predictions or by improving the correlation between uncertainty and error. Note, in contrast to Figure~\ref{fig:example_retention_curves}b, the quality of the ranking affects area under the curve far more than for error-retention curves. Thus, this metric is especially useful when errors have similar magnitudes.

Finally, it is necessary to point out that area under the error-retention curve and F1-retention curve is a \emph{summary statistic} which describes possible \emph{operating points}. We can specify a particular operating point, such as 95\% retention, and evaluate the error or F1 at that point for comparison. This is also an important figure, as all models work at a particular operating point which satisfies task-specific desiderata. In this work the desiderata for all tasks will be to not reject more than 5\% of the input data.  

\section{Shifts Dataset General Datasheet}\label{apn:datasheet}

Here we describe the motivation, uses, distribution as well as the maintenance and support plan for the Shifts Dataset as whole in the \emph{datasheet for datasets} format~\cite{gebru2018datasheets}. The details of the composition, collection and pre-prossessing of each component dataset are provided in appendices~\ref{apn:tab}-\ref{apn:sdc}

\paragraph{Motivation} As discussed at length in the main body of the paper, the primary goal for the creation of the Shifts Dataset was the evaluation of uncertainty quantification models and robustness to distributional shift on a range of large-scale, industrial tasks spanning multiple modalities. To this end, Yandex Research, in collaboration with the Yandex.Translate, Yandex.Weather services and Yandex Self-Driving Group created the Shifts Dataset. As the dataset creation was done by Yandex teams, it was therefore funded by Yandex.   

\paragraph{Uses} The dataset is used as part of the Shifts Challenge which was organized as part of NeurIPS2021, which was organized around this dataset\footnote{ \url{research.yandex.com/shifts}}. The Shifts Challenge consists of three tracks organized around each of the consituent datasets within Shifts. The dataset, baseline models and code to reproduce it all is provided in a GitHub repository\footnote{\url{https://github.com/yandex-research/shifts}}. Other than uncertainty and robustness research the dataset could be used for developing better models for each of the separate tasks - tabular data, translation and vehicle motion prediction.

\paragraph{Distribution} The parts of the dataset which were produced by Yandex are distributed under an open-source CC BY NC SA 4.0 license. All the code is available under an open-source Apache 2.0 licence. It is our intention that the dataset be freely available for research purposes. The dataset is available as a tarball download from GitHub. Currently, as the Shifts Challenge is still underway, only the training and development sets are available. However, the full dataset, with full accompanying metadata, will be available once the challenge concludes on November 1st, 2021. Licence details for each constituent dataset in Shifts are described in appendices~\ref{apn:tab}-\ref{apn:sdc}.

\paragraph{Maintenance}
The dataset is being actively maintained by Yandex Research, with support from the weather, translation and self-driving teams, and the teams can be contacted by raising an issue on GitHub and by writing to the first author of this paper. The dataset is currently hosted on Yandex S3 storage and will be hosted there permanently for the foreseeable future. The dataset can be updated at the discretion of the dataset creators, though regular updates are not planned. Updates which expand the evaluation sets or add new ones will mean that the previous dev/eval sets are supported. Updates which fix errors in dev/eval sets mean that the prior ones are obsolete and unsupported. If any update is to occur, we will make an announcement via GitHub, twitter, and the Shifts challenge mailing list.  Currently, as the data comes directly from Yandex, we do not allows other parties to update the Shifts Dataset. However, any issues found can be logged by raising an issue on GitHub or contacting the first author of this paper so that we can address them. Furthermore, as we are releasing the data under an open-source CC BY NC SA 4.0 license which allows modifications, we are happy for people to create derivative datasets using ours, provided the modifications are documents and the original dataset references.

\paragraph{Societal Consequences and Guidelines for Ethical Use} Research on uncertainty estimation and robustness aims to make AI safer and more reliable, and therefore has limited negative societal consequences overall. Users of this dataset are encouraged to use it for the purpose of improving the reliability and safety of large-scale applications of machine learning. Furthermore, we encourage users of out dataset to develop compute and memory efficient methods for improving safety and reliability. 

\paragraph{Responsibility} The authors confirm that, to the best of our knowledge, the released dataset does not violate any prior licenses or rights. However, if such a violation were to exist, we are responsible for resolving this issue. 
\section{Tabular Weather}\label{apn:tab}

The current appendix contains a description of the composition, collection, pre-processing and partitioning of the Shifts Tabular Weather Prediction dataset. Additionally, it contains a description of the metrics used for assessment and an expanded set of experimental results.

\subsection{Dataset Description}\label{apn:weather-description}

\paragraph{Composition} The data consists of pairs of meteorological features and target values at a particular latitude/longitude and time. The target value is air temperature measurements at 2 metres above the ground for regression and precipitation and cloudiness class from weather station measurements for classification. The feature vectors include both weather-related features such as sun evaluation at the current location, climate values of temperature,  pressure and topography, and meteorological parameters on different pressure and surface levels from \emph{weather forecast model predictions}. \emph{Weather forecast model predictions} are values produced by the following weather forecast models: Global Forecast System (GFS),\footnote{\url{https://www.ncdc.noaa.gov/data-access/model-data/model-datasets/global-forcast-system-gfs}} Global Deterministic Forecast System from the Canadian Meteorological Center (CMC),\footnote{\url{https://weather.gc.ca/grib/grib2_glb_25km_e.html}} and the Weather Research and Forecasting (WRF) Model.\footnote{\url{https://www.mmm.ucar.edu/weather-research-and-forecasting-model}} Each model returns the following predicted values: wind, humidity, pressure, clouds, precipitation, dew point, snow depth, air and soil temperature characteristics. Where applicable, the predictions are given at different isobaric levels from 50 hPa ($\approx$ 20 km above ground) to the ground level. The GFS and WRF models run 4 times a day (0, 6, 12 and 18 GMT), and the CMC model runs twice a day (0 and 12 GMT). Model spatial grid resolution is $0.25\degree \times 0.25\degree$ for GFS and $0.24\degree \times 0.24\degree$ for CMC. The WRF model is calculated for over 60 domains all over the globe, spatial resolution for each domain is 6 $\times$ 6 km.  Altogether, there are 123 features in total. It is important to note that the features are highly heterogeneous, i.e., they are of different types and scales. The target air temperature values at different locations are taken from about 8K weather stations located across the globe, each of which periodically ($\approx$ each 3 hours) reports a set of measurements. In total, the dataset has 129 columns: 123 features, 4 meta-data attributes including time, latitude, longitude, and 2 targets - temperature (target for regression task), precipitation class (target for classification task) and climate type. The full feature list is provided in Section~\ref{apn:tab-features}.

\paragraph{Collection Process}  The data for features from  GFS and CMC weather forecast models was downloaded in the GRIB file format from the web resources \url{https://www.ncdc.noaa.gov} and \url{https://weather.gc.ca/}, respectively. The GRIB files were decoded and collected by the production system of the Yandex Weather forecast service. MD5 hashes for files were checked after downloading the data. The parameters from WRF model were obtained from WRF model v3.6.1 computation on Yandex Weather servers. The data was checked for mistakes and outliers. Some parameters were converted to different units (for example degrees from K to C). We selected a subset of 123 weather parameters from the full dataset based on expertise and research of feature importance for weather forecasts of temperature and precipitation for the Yandex Weather production system. The data for weather station observations was downloaded from \url{https://www.ncdc.noaa.gov} and was decoded from SYNOP code. We filtered missed values and outliers by comparing with previous observations on the same weather station, and by comparing observation with nearby weather stations. Scripts and program codes for data collection and processing were prepared by in-house Yandex Weather software engineers. The period of data collection is from September 2018 to September 2019. 

\paragraph{Preprocessing, Cleaning and Labelling} The data was logged during applying trained CatBoost models for weather forecast prediction of the Yandex Weather service and was validated on Yandex Weather users by providing actual weather forecasts and accessing its mistakes on users and station measurements. We labeled data to match the timestamp of features and targets from these logs. Also we selected features only for latitudes and longitudes of weather observation stations to match with the measurements. Targets for air temperature were converted to degrees Celsius. Targets for precipitation class were constructed from cloudiness and precipitation measurements to create 9 classes and labeled as follows: 0 --- no precipitation, no clouds, 1 --- no precipitation, partly cloudy, 2 --- rain, partly cloudy, 3 --- sleet, partly cloudy, 4 --- snow, partly cloudy, 5 --- no precipitation, cloudy, 6 --- rain, cloudy, 7 --- sleet, cloudy, 8 --- snow, cloudy. The ``raw'' data was not saved, because it requires large amount of disk space. It was deleted after processing the data. 

\paragraph{Partitioning into train, development, and evaluation sets}

To analyze the robustness of learned models to \emph{climate shifts}, we use the Koppen climate classification~\cite{chen2013koppen} that provides publicly available data\footnote{Available to download from \url{http://hanschen.org/koppen}} that maps latitudes and longitudes at a $0.5^{\circ}$ resolution to one of five main climate types: \textit{Tropical}, \textit{Dry}, \textit{Mild Temperate}, \textit{Snow} and \textit{Polar}. This information is available over the years 1901 to 2010. The Weather Prediction dataset is augmented such that each sample has an associated climate type. The climate type is determined by minimizing the 1-norm between the longitudes/latitudes in the weather data and the Koppen climate classification for the most recent year available, 2010. The climate type is not used as a training feature.

\begin{figure}
  \centering
  \includegraphics[width=\linewidth]{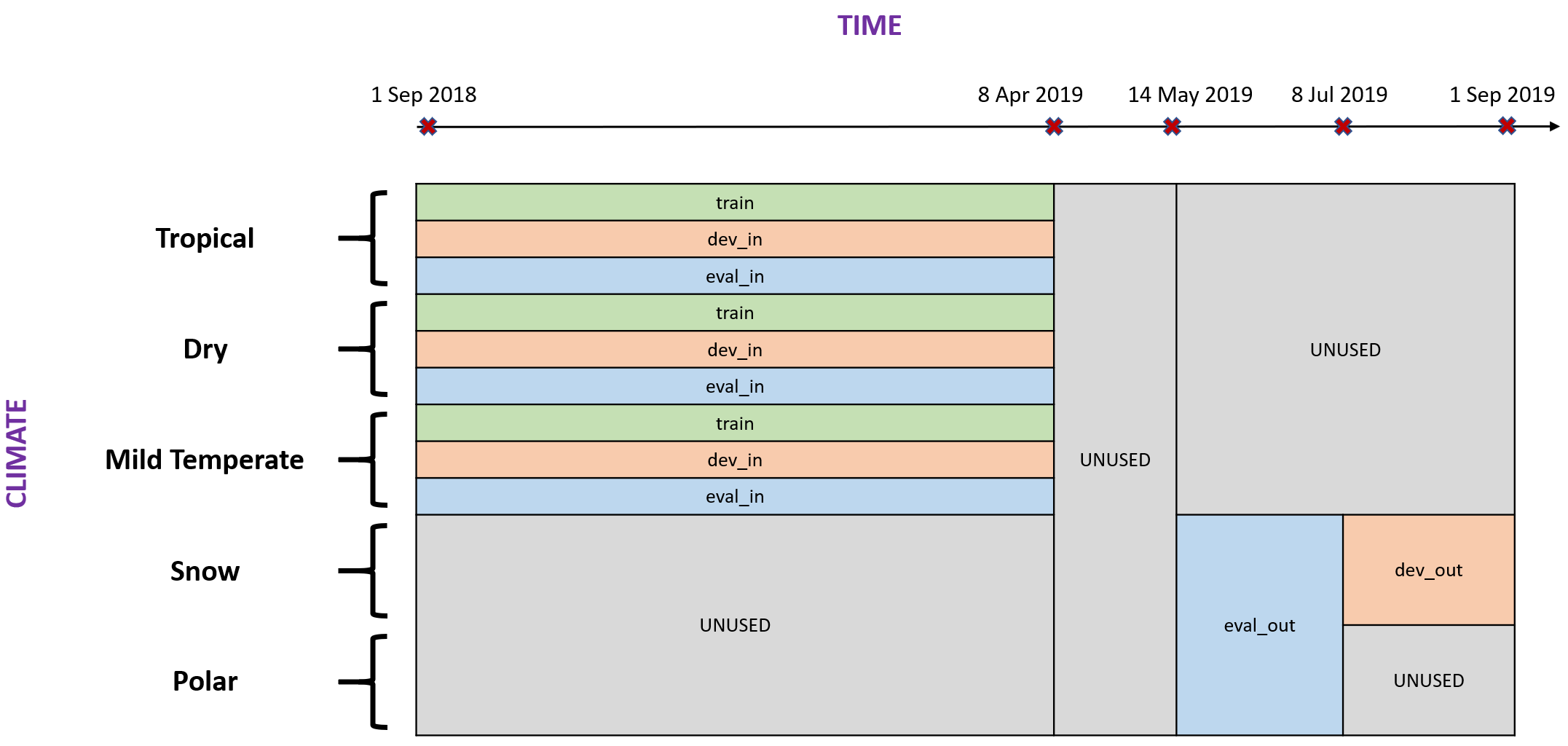}
  \caption{Canonical Partitioning of Weather Prediction dataset.}
  \label{fig:data}
\end{figure}

There are 10M records in the full dataset distributed uniformly between September $1^{\text{st}}$, 2018, and September $1^{\text{st}}$, 2019, with samples across all five climate types. To test the robustness of the models, we evaluate how well they perform on time-shifted and climate-shifted data. Model performance is expected to decrease with time and climate shifts. However, a robust model is expected to be stable with these shifts. 

In order to provide a standard benchmark which contains data which is both matched and shifted relative to the training set, we split the full dataset into `canonically partitioned'\footnote{Alternative partitioning can be made from the full data, but we will use the canonical partition throughout this work.} training, development, and evaluation datasets as follows (see Figure \ref{fig:data}):
\begin{itemize}
    \item The training data consists of measurements made from September 2018 till April $8^{\text{th}}$, 2019 for climate types \textit{Tropical}, \textit{Dry}, and \textit{Mild Temperate}. The training data includes two dummy rows in order to ensure there is at least one example of each of the precipitation classes (the targets for the classification task). The values for each of the features of the dummy examples are computed by averaging across the whole training dataset.
    \item The development data is composed of in-domain (\texttt{dev\_in}) and out-of-domain (\texttt{dev\_out}) data. The in-domain data corresponds to the same time range and climate types as the training data. The out-of-domain development data consists of measurements made from $8^{\text{th}}$ July till $1^{\text{st}}$ September 2019 for the climate type \textit{Snow}. 50K data points are subsampled for the climate type \textit{Snow} within this time range to construct \texttt{dev\_out}.
    \item The evaluation data is also composed of in-domain (\texttt{eval\_in}) and out-of-domain (\texttt{eval\_out}) data. As before, the in-domain data corresponds to the same time range and climate types as the training data. The out-of-domain evaluation data is further shifted than the out-of-domain development data; measurements are taken from $14^{\text{th}}$ May till $8^{\text{th}}$ July 2019, which is more distant in terms of the time of the year from the in-domain data compared to the out-of-domain development data. The climate types are restricted to \textit{Snow} and \textit{Polar}.
\end{itemize}

\begin{figure}
  \centering
  \includegraphics[width=0.8\linewidth]{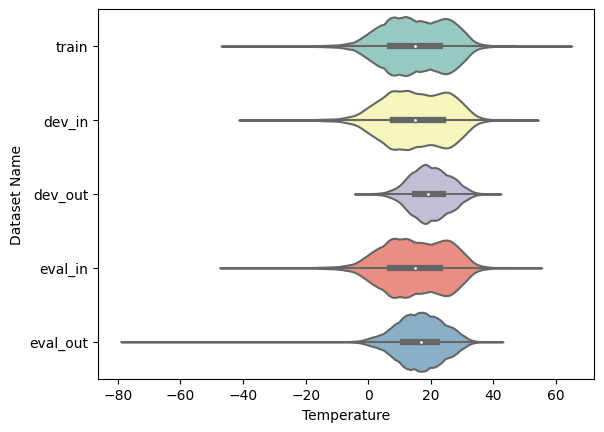}
  \caption{Temperature distributions on canonical partitions of Weather Prediction dataset.}
  \label{fig:violin}
\end{figure}

\begin{figure}[htbp!]
     \centering
     \begin{subfigure}[b]{0.32\textwidth}
         \centering
         \includegraphics[width=\textwidth]{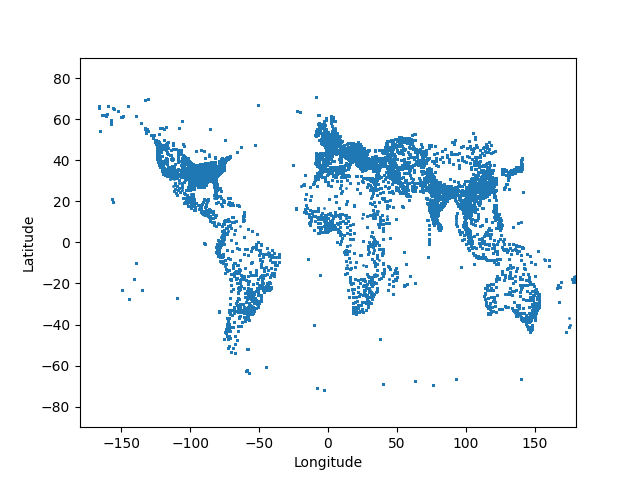}
         \caption{\texttt{train}.}
         \label{fig:l_train}
     \end{subfigure}
     \begin{subfigure}[b]{0.32\textwidth}
         \centering
         \includegraphics[width=\textwidth]{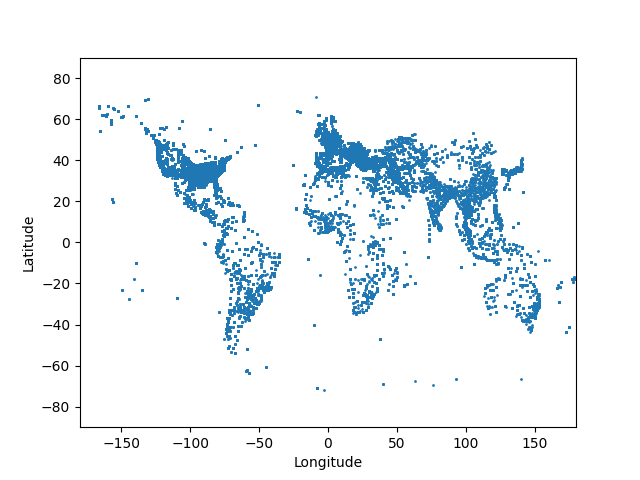}
         \caption{\texttt{dev\_in}.}
         \label{fig:l_dev_in}
     \end{subfigure}
     \begin{subfigure}[b]{0.32\textwidth}
         \centering
         \includegraphics[width=\textwidth]{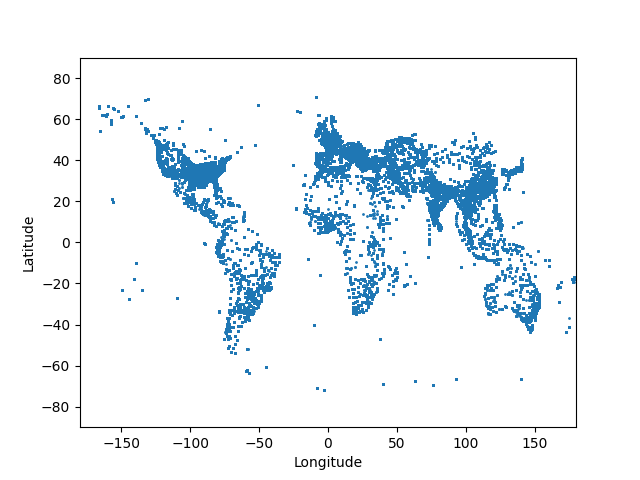}
         \caption{\texttt{eval\_in}.}
         \label{fig:l_eval_in}
     \end{subfigure}
     \\
     \begin{subfigure}[b]{0.32\textwidth}
         \centering
         \includegraphics[width=\textwidth]{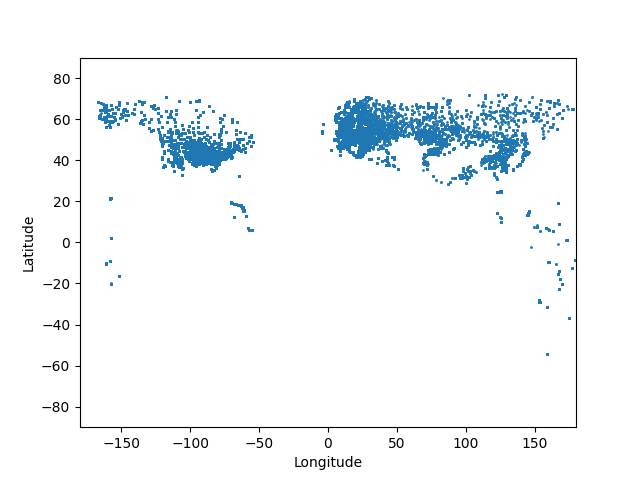}
         \caption{\texttt{dev\_out}.}
         \label{fig:l_dev_out}
     \end{subfigure} 
     \begin{subfigure}[b]{0.32\textwidth}
         \centering
         \includegraphics[width=\textwidth]{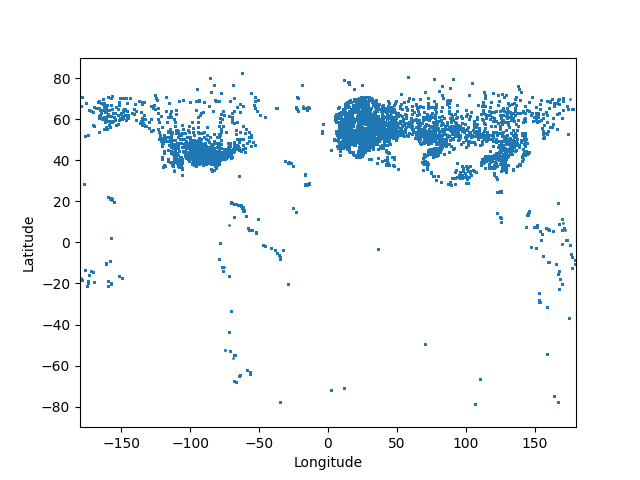}
         \caption{\texttt{eval\_out}.}
         \label{fig:l_eval_out}
     \end{subfigure} 
        \caption{Location of samples from canonical partitioning of Weather Prediction dataset.}
        \label{fig:weather_location}
\end{figure}

Table \ref{tab:tabular_data} details the number of samples in the selected partition of the data. It also details the number of samples for each climate type for each part of the dataset. The in-domain data is split in approximately 83.7-1.3-15\% ratio between training, development, and evaluation. Figure~\ref{fig:violin} depicts the shift in the target temperatures between the training, development, and evaluation datasets. It is clear that the temperature distribution is different for \texttt{dev\_out} and \texttt{eval\_out} compared to the in-domain sets. The higher average temperature in the out of domain sets is perhaps due to the out of domain data being sourced from the Summer regions (for the northern hemisphere) while the in-domain data is largely sourced from the Winter time period. Figure~\ref{fig:weather_location} further shows the shift in the samples' locations (latitudes/longitudes) between training, development, and evaluation datasets. The location shift is a natural result of the climate shifts present in the datasets where the training data tends to correspond to warmer parts of the world, whereas the development and evaluation datasets include colder climates too.

\begin{table}
\caption{Number of samples in the canonical partitioning of Weather Prediction dataset.}
\centering
\begin{small}
    \begin{tabular}{cl|rrrrrr}
    \toprule
  & \multirow{2}*{Data} & \multicolumn{6}{c}{\# of samples} \\ 
  & & \multicolumn{1}{c}{\textbf{Total}} & \multicolumn{1}{c}{Tropical} & \multicolumn{1}{c}{Dry} & \multicolumn{1}{c}{Mild Temperate} & \multicolumn{1}{c}{Snow} & \multicolumn{1}{c}{Polar} \\
  \midrule
\multirow{1}*{Training}  & \texttt{train} & \textbf{3,129,592} & 416,310 & 690,284 & 2,022,998 & 0 & 0 \\
  \midrule
\multirow{3}*{Development}  
  & \texttt{dev\_in} & \textbf{50,000} & 6,641 & 10,961 & 32,398 & 0 & 0 \\
  & \texttt{dev\_out} & \textbf{50,000} & 0 & 0 & 0 & 50,000 & 0 \\
  & \texttt{dev} & \textbf{100,000} & 6,641 & 10,961 & 32,398 & 50,000 & 0 \\
  \midrule
\multirow{3}*{Evaluation}  
  & \texttt{eval\_in} & \textbf{561,105} & 74,406 & 123,487 & 363,212 & 0 & 0 \\
  & \texttt{eval\_out} & \textbf{576,626} & 0 & 0 & 0 & 525,967 & 50,659 \\
  & \texttt{eval} & \textbf{1,137,731} & 74,406 & 123,487 & 363,212 & 525,967 & 50,659 \\
  \bottomrule
    \end{tabular}
    \end{small}
    \label{tab:tabular_data}
\end{table}

\begin{figure}[htbp!]
     \centering
     \begin{subfigure}[b]{0.32\textwidth}
         \centering
         \includegraphics[width=\textwidth]{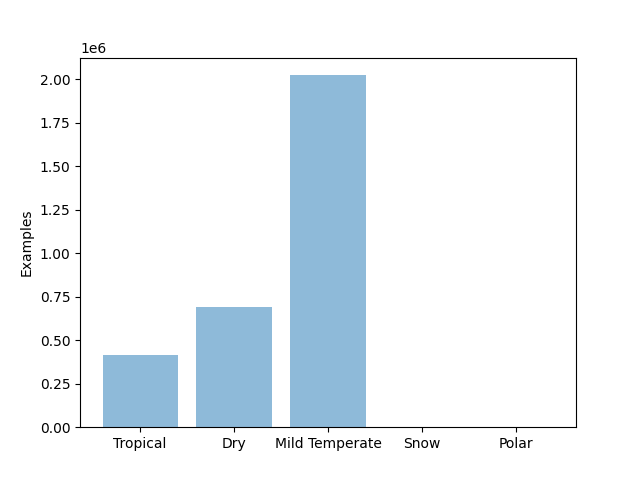}
         \caption{\texttt{train}.}
         \label{fig:c_train}
     \end{subfigure}
     \begin{subfigure}[b]{0.32\textwidth}
         \centering
         \includegraphics[width=\textwidth]{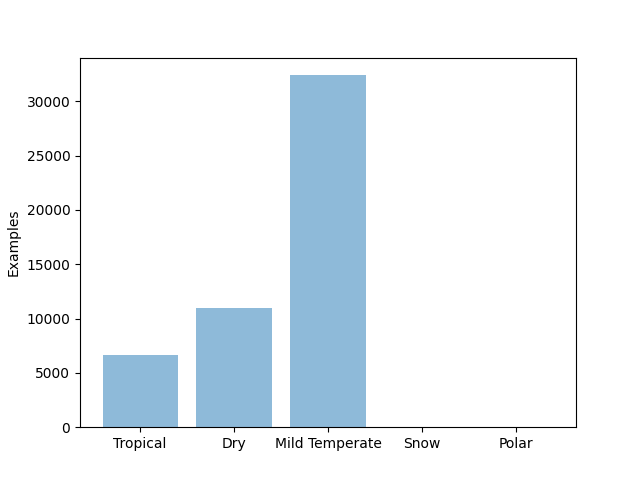}
         \caption{\texttt{dev\_in}.}
         \label{fig:c_dev_in}
     \end{subfigure}
     \begin{subfigure}[b]{0.32\textwidth}
         \centering
         \includegraphics[width=\textwidth]{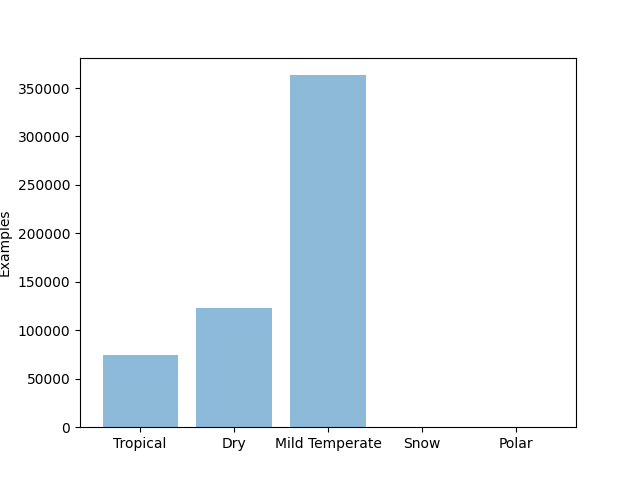}
         \caption{\texttt{eval\_in}.}
         \label{fig:c_eval_in}
     \end{subfigure}
     \\
     \begin{subfigure}[b]{0.32\textwidth}
         \centering
         \includegraphics[width=\textwidth]{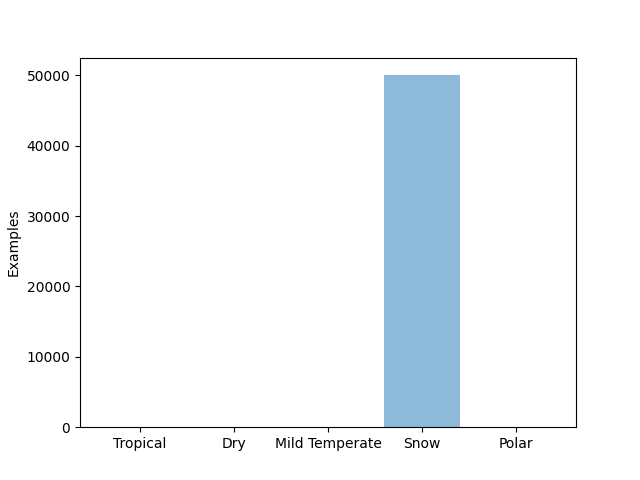}
         \caption{\texttt{dev\_out}.}
         \label{fig:c_dev_out}
     \end{subfigure} 
     \begin{subfigure}[b]{0.32\textwidth}
         \centering
         \includegraphics[width=\textwidth]{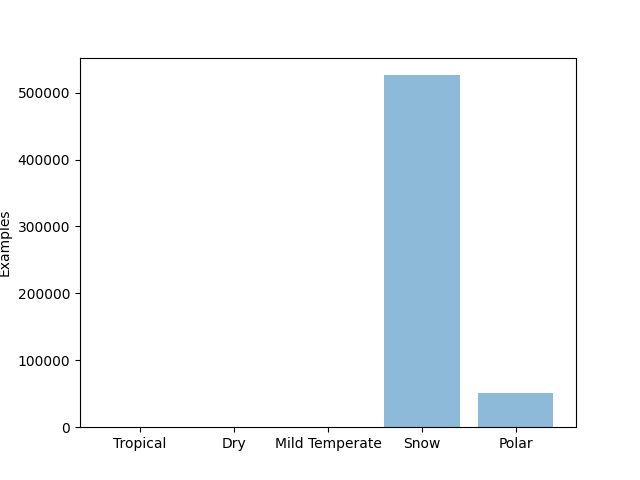}
         \caption{\texttt{eval\_out}.}
         \label{fig:c_eval_out}
     \end{subfigure} 
      \begin{subfigure}[b]{0.32\textwidth}
         \centering
         \includegraphics[width=\textwidth]{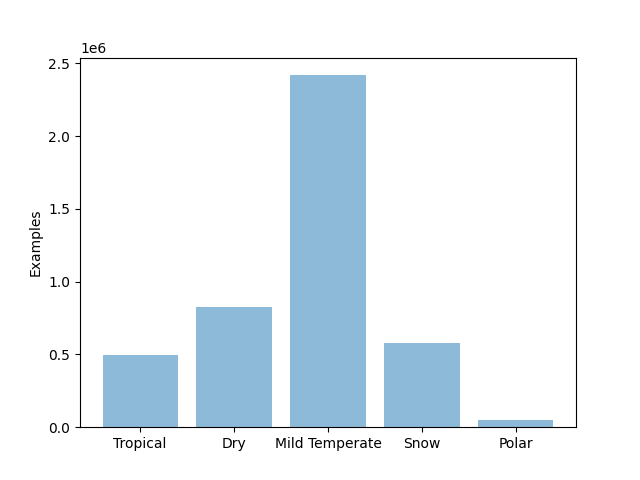}
         \caption{\texttt{all}.}
         \label{fig:c_all}
     \end{subfigure}
        \caption{Distribution of climate types from canonical partitioning of Weather Prediction dataset.}
        \label{fig-apn:weather_clim_typ}
\end{figure}

\begin{figure}[htbp!]
     \centering
     \begin{subfigure}[b]{0.32\textwidth}
         \centering
         \includegraphics[width=\textwidth]{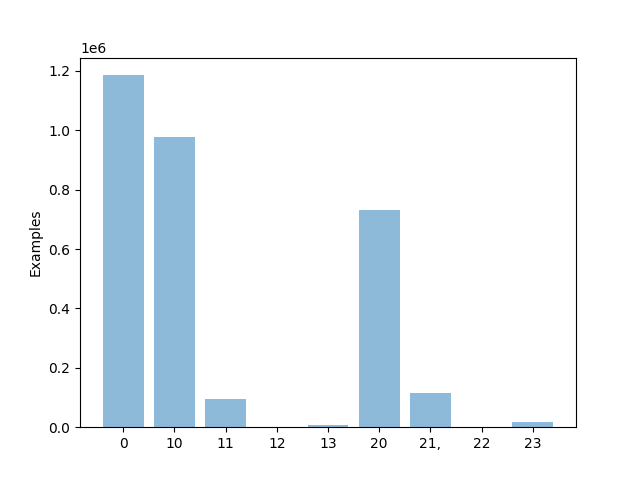}
         \caption{\texttt{train}.}
         \label{fig:d_train}
     \end{subfigure}
     \begin{subfigure}[b]{0.32\textwidth}
         \centering
         \includegraphics[width=\textwidth]{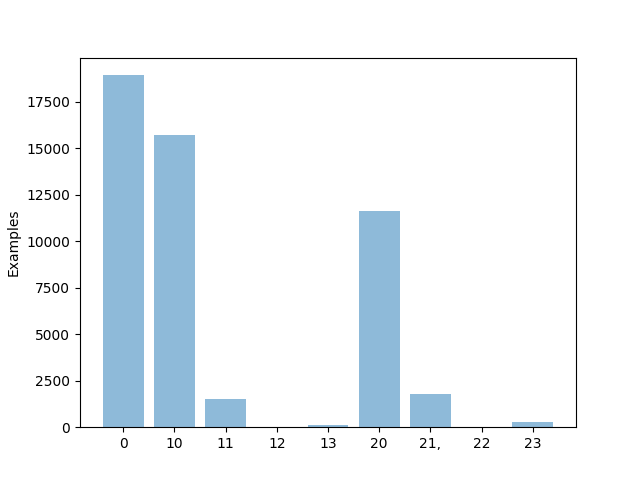}
         \caption{\texttt{dev\_in}.}
         \label{fig:d_dev_in}
     \end{subfigure}
     \begin{subfigure}[b]{0.32\textwidth}
         \centering
         \includegraphics[width=\textwidth]{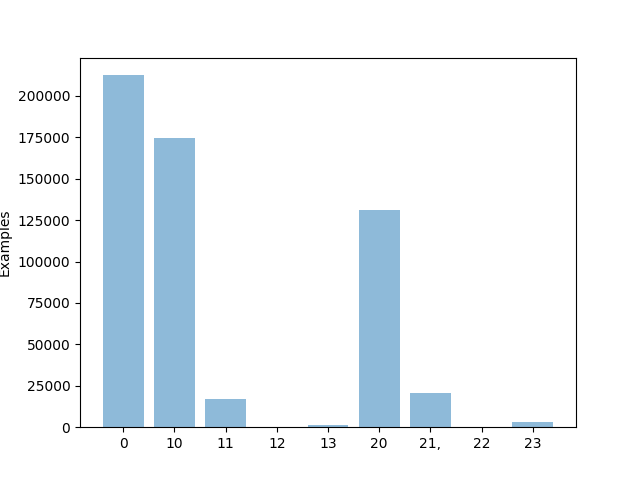}
         \caption{\texttt{eval\_in}.}
         \label{fig:d_eval_in}
     \end{subfigure}
     \\
     \begin{subfigure}[b]{0.32\textwidth}
         \centering
         \includegraphics[width=\textwidth]{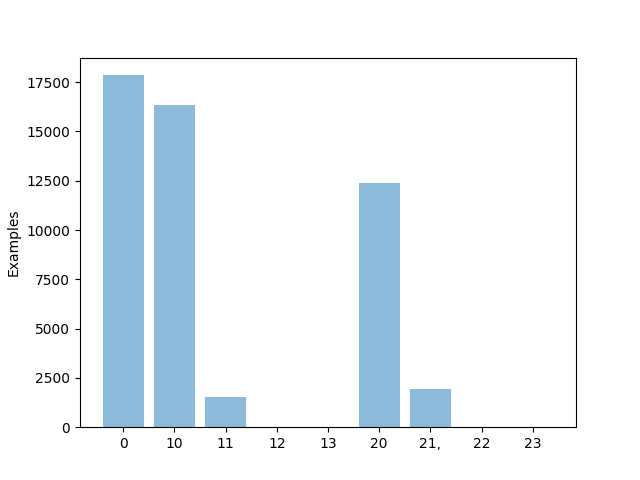}
         \caption{\texttt{dev\_out}.}
         \label{fig:d_dev_out}
     \end{subfigure} 
     \begin{subfigure}[b]{0.32\textwidth}
         \centering
         \includegraphics[width=\textwidth]{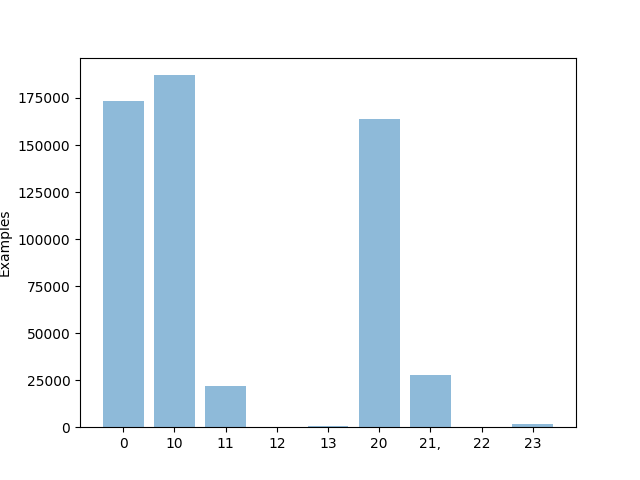}
         \caption{\texttt{eval\_out}.}
         \label{fig:d_eval_out}
     \end{subfigure} 
      \begin{subfigure}[b]{0.32\textwidth}
         \centering
         \includegraphics[width=\textwidth]{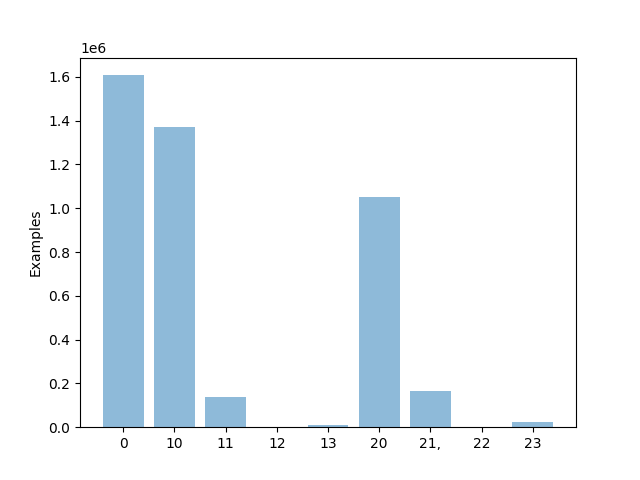}
         \caption{\texttt{all}.}
         \label{fig:d_all}
     \end{subfigure}
        \caption{Distribution of precipitation classes from canonical partitioning of Weather Prediction dataset.}
        \label{fig-apn:precip_class_typ}
\end{figure}

\paragraph{Format} This dataset is provided in CSV format. 

\paragraph{Licence} This dataset is provided under the CC BY NC SA 4.0 license. 

\subsection{Detailed description of features and targets}

\paragraph{Meta-Data Features}

\begin{enumerate}
    \item fact\_time --- timestamp  
    \item fact\_latitude --- geographical latitude, degrees
    \item fact\_longitude --- geographical longitude, degrees
    \item climate --- major climate type
\end{enumerate}

\paragraph{Targets}

\begin{enumerate}
    \item fact\_temperature --- air temperature 2m above the ground, C
    \item fact\_cwsm\_class - precipitation class
\end{enumerate}

\paragraph{Features}\label{apn:tab-features}

\begin{enumerate}
\item climate\_pressure --- climate pressure, mmHg 
\item climate\_temperature --- climate temperature, C 
\item cmc\_0\_0\_0\_1000 --- temperature at 1000 hPa isobaric level, K
\item cmc\_0\_0\_0\_2 --- temperature at 2m, K
\item cmc\_0\_0\_0\_2\_grad --- difference between temperatures on adjacent horizons at 2m, K
\item cmc\_0\_0\_0\_2\_interpolated --- temperature at 2m interpolated between horizons, K
\item cmc\_0\_0\_0\_2\_next --- temperature at 2m for next horizon, K
\item cmc\_0\_0\_0\_500 --- temperature at 500 hPa isobaric level, K
\item cmc\_0\_0\_0\_700 --- temperature at 700 hPa isobaric level, K
\item cmc\_0\_0\_0\_850 --- temperature at 850 hPa isobaric level, K
\item cmc\_0\_0\_0\_925 --- temperature at 925 hPa isobaric level, K
\item cmc\_0\_0\_6\_2 --- dew point temp at 2m, K
\item cmc\_0\_0\_7\_1000 --- dew point depression at 1000 hPa isobaric level, K
\item cmc\_0\_0\_7\_2 --- dew point depression at 2m, K
\item cmc\_0\_0\_7\_500 --- dew point depression at 500 hPa isobaric level, K
\item cmc\_0\_0\_7\_700 --- dew point depression at 700 hPa isobaric level, K
\item cmc\_0\_0\_7\_850 --- dew point depression at 850 hPa isobaric level, K
\item cmc\_0\_0\_7\_925 --- dew point depression at 925 hPa isobaric level, K
\item cmc\_0\_1\_0\_0 --- absolute humidity from 0 to 1
\item cmc\_0\_1\_11\_0 --- snow depth, m
\item cmc\_0\_1\_65\_0 --- rain accumulated from cmc gentime, mm
\item cmc\_0\_1\_65\_0\_grad --- rain accumulated from cmc gentime difference between adjacent horizons, mm
\item cmc\_0\_1\_65\_0\_next --- rain accumulated from cmc gentime for next horizon, mm
\item cmc\_0\_1\_66\_0 --- snow accumulated from cmc gentime, mm
\item cmc\_0\_1\_66\_0\_grad --- snow accumulated from cmc gentime difference between adjacent horizons, mm
\item cmc\_0\_1\_66\_0\_next --- snow accumulated from cmc gentime for next horizon, mm
\item cmc\_0\_1\_67\_0 --- ice rain accumulated from cmc gentime, mm
\item cmc\_0\_1\_67\_0\_grad --- ice rain accumulated from cmc gentime difference between adjacent horizons, mm
\item cmc\_0\_1\_67\_0\_next --- ice rain accumulated from cmc gentime for next horizon, mm
\item cmc\_0\_1\_68\_0 --- iced graupel accumulated from cmc gentime, mm
\item cmc\_0\_1\_68\_0\_grad --- iced graupel accumulated from cmc gentime difference between adjacent horizons, mm
\item cmc\_0\_1\_68\_0\_next --- iced graupel accumulated  from cmc gentime for next horizon, mm
\item cmc\_0\_1\_7\_0 --- instant precipitation intensity, mm/h
\item cmc\_0\_2\_2\_10 --- wind U component at 10m, m/s
\item cmc\_0\_2\_2\_1000 --- wind U component at 1000 hPa isobaric level, m/s
\item cmc\_0\_2\_2\_500 --- wind U component at 500 hPa isobaric level, m/s
\item cmc\_0\_2\_2\_700 --- wind U component at 700 hPa isobaric level, m/s
\item cmc\_0\_2\_2\_850 --- wind U component at 850 hPa isobaric level, m/s
\item cmc\_0\_2\_2\_925 --- wind U component at 925 hPa isobaric level, m/s
\item cmc\_0\_2\_3\_10 --- wind V component at 10m, m/s
\item cmc\_0\_2\_3\_1000 --- wind V component at 1000 hPa isobaric level, m/s
\item cmc\_0\_2\_3\_500 --- wind V component at 500 hPa isobaric level, m/s
\item cmc\_0\_2\_3\_700 --- wind V component at 700 hPa isobaric level, m/s
\item cmc\_0\_2\_3\_850 --- wind V component at 850 hPa isobaric level, m/s
\item cmc\_0\_2\_3\_925 --- wind V component at 925 hPa isobaric level, m/s
\item cmc\_0\_3\_0\_0 --- surface pressure, Pa
\item cmc\_0\_3\_0\_0\_next --- next horizon surface pressure, Pa
\item cmc\_0\_3\_1\_0 --- sea level pressure, Pa
\item cmc\_0\_3\_5\_1000 --- geopotential height at 1000 hPa isobaric level, gpm (geopotential meter)
\item cmc\_0\_3\_5\_500 --- geopotential height at 500 hPa isobaric level, gpm 
\item cmc\_0\_3\_5\_700 --- geopotential height at 700 hPa isobaric level, gpm
\item cmc\_0\_3\_5\_850 --- geopotential height at 850 hPa isobaric level, gpm
\item cmc\_0\_3\_5\_925 --- geopotential height at 925 hPa isobaric level, gpm
\item cmc\_0\_6\_1\_0 --- cloudiness, \% from 0 to 100
\item cmc\_available --- is there any data from cmc
\item cmc\_horizon\_h --- cmc horizon, h
\item cmc\_precipitations --- avg precipitations rate between adjacent horizons, mm/h
\item cmc\_timedelta\_s --- difference between cmc and forecast time, s
\item gfs\_2m\_dewpoint --- dew point temperature at 2m, C
\item gfs\_2m\_dewpoint\_grad --- dew point temperature at 2m difference between horizons, C
\item gfs\_2m\_dewpoint\_next --- dew point temperature on next horizon, C
\item gfs\_a\_vorticity --- absolute vorticity at height 1000 hPa, s-1
\item gfs\_available --- is there any data from gfs
\item gfs\_cloudness --- sum of 3 level cloudiness, from 0 to 3
\item gfs\_clouds\_sea --- Cloud mixing ratio at level 1000 hPa, kg/kg 0.0
\item gfs\_horizon\_h --- gfs horizon, h
\item gfs\_humidity --- relative humidity at 2m, \%
\item gfs\_precipitable\_water --- total precipitable water, kg m$^-2$
\item gfs\_precipitations --- avg precipitations rate between adjacent horizons, mm/h
\item gfs\_pressure --- surface pressure, mmHg
\item gfs\_r\_velocity --- vertical Velocity at 1000 hPa, Pa/s
\item gfs\_soil\_temperature --- soil temperature at 0.0-0.1 m, C
\item gfs\_soil\_temperature\_available --- is there gfs soil temp data
\item gfs\_temperature\_10000 --- temperature at vertical level at 100 hPa, C
\item gfs\_temperature\_15000 --- temperature at vertical level at 150 hPa, C
\item gfs\_temperature\_20000 --- temperature at vertical level at 200 hPa, C
\item gfs\_temperature\_25000 --- temperature at vertical level at 250 hPa, C
\item gfs\_temperature\_30000 --- temperature at vertical level at 300 hPa, C
\item gfs\_temperature\_35000 --- temperature at vertical level at 350 hPa, C
\item gfs\_temperature\_40000 --- temperature at vertical level at 400 hPa, C
\item gfs\_temperature\_45000 --- temperature at vertical level at 450 hPa, C
\item gfs\_temperature\_5000 --- temperature at vertical level at 50 hPa, C
\item gfs\_temperature\_50000 --- temperature at vertical level at 500 hPa, C
\item gfs\_temperature\_55000 --- temperature at vertical level at 550 hPa, C
\item gfs\_temperature\_60000 --- temperature at vertical level at 600 hPa, C
\item gfs\_temperature\_65000 --- temperature at vertical level at 650 hPa, C
\item gfs\_temperature\_7000 --- temperature at vertical level at 70 hPa, C
\item gfs\_temperature\_70000 --- temperature at vertical level at 700 hPa, C
\item gfs\_temperature\_75000 --- temperature at vertical level at 750 hPa, C
\item gfs\_temperature\_80000 --- temperature at vertical level at 800 hPa, C
\item gfs\_temperature\_85000 --- temperature at vertical level at 850 hPa, C
\item gfs\_temperature\_90000 --- temperature at vertical level at 900 hPa, C
\item gfs\_temperature\_92500 --- temperature at vertical level at 925 hPa, C
\item gfs\_temperature\_95000 --- temperature at vertical level at 950 hPa, C
\item gfs\_temperature\_97500 --- temperature at vertical level at 975 hPa, C
\item gfs\_temperature\_sea --- temperature at 2m, C
\item gfs\_temperature\_sea\_grad --- temperature difference adjacent horizons at 2m
\item gfs\_temperature\_sea\_interpolated --- gfs\_temperature\_sea\_interpolated between horizons, C
\item gfs\_temperature\_sea\_next --- next horizon temperature at 2m, C
\item gfs\_timedelta\_s --- difference between gfs and forecast time, s
\item gfs\_total\_clouds\_cover\_high --- cloud coverage (between horizons, divisible by 6) at high level, \%
\item gfs\_total\_clouds\_cover\_low --- cloud coverage (between horizons, divisible by 6) at low level, \%
\item gfs\_total\_clouds\_cover\_low\_grad --- difference between low level cloud coverage on adjacent horizons, \%
\item gfs\_total\_clouds\_cover\_low\_next --- next horizon cloud coverage (between horizons, divisible by 6) at low level, \%
\item gfs\_total\_clouds\_cover\_middle --- cloud coverage (between horizons, divisible by 6) at middle level, \%
\item gfs\_u\_wind --- 10 meter U wind component, m/s
\item gfs\_v\_wind --- 10 meter V wind component, m/s
\item gfs\_wind\_speed  --- wind velocity, sqrt(gfs\_u\_wind$^2$ + gfs\_v\_wind$^2$), m/s
\item sun\_elevation --- sun height proxy above horizon (without corrections for precision and diffraction)
\item topography\_bathymetry --- height above or below sea level, m
\item wrf\_available --- is there any data from wrf
\item wrf\_graupel --- avg graupel rate between two horizons, mm/h
\item wrf\_hail --- hail velocity on two horizons, mm/h
\item wrf\_psfc --- pressure, Pa
\item wrf\_rain --- avg rain rate between two horizons, mm/h
\item wrf\_rh2 --- relative humidity at 2m, from 0 to 1
\item wrf\_snow --- avg snow rate between two horizons, mm/h
\item wrf\_t2 --- temperature at 2m, K
\item wrf\_t2\_grad --- difference between temperatures at 2m on adjacent horizons, K
\item wrf\_t2\_interpolated --- wrf\_t2\_interpolated between horizons, K
\item wrf\_t2\_next --- next horizon temperature at 2m, K
\item wrf\_wind\_u --- wind U component, m/s
\item wrf\_wind\_v --- wind V component, m/s
\end{enumerate}

\subsection{Metrics}\label{subsection:performance_metrics}

 We aim at comparing different models in terms of uncertainty estimation and robustness to distributional shifts. Several performance metrics are considered. 

\textbf{Predictive Performance} For temperature prediction, predictive performance and robustness to distributional shifts are evaluated by measuring RMSE and MAE between predictions and targets: lower the RMSE/MAE score on the test sets, greater the robustness of the models to the distributional shift. For classification, we use accuracy and macro-averaged F1 (one-vs-all averaged with no weighting). More robust models are expected to have higher values of these metrics.

\textbf{Joint assessment of Uncertainty and Robustness} We jointly assess robustness and uncertainty estimation via error-retention and F1-retention curves, described in Section~\ref{sec:metrics} and detailed in Appendix~\ref{apn:metrics}. For regression, we use MSE as the error metric instead of RMSE as it is linear with respect to the error for each datapoint. For the F1-retention curve an acceptable prediction is defined as one where MSE < 1.0. This corresponds to an error of 1 degree or less, which most people cannot feel. Typically people are sensitive to differences in surrounding temperature of over a degree. These two performance metrics are respectively denoted as R-AUC and F1-AUC. For classification, we use the error rate to compute R-AUC. For both classification and regression, a good uncertainty measure is expected to achieve low R-AUC and high F1-AUC. Additionally, the F1 score at a retention rate of 95\% of the most certain samples is also quoted and is denoted as F1@95\%, which is a single point summary jointly of the uncertainty and robustness. Finally, ROC-AUC is used as a summary statistic for evaluating uncertainty-based out-of-distribution data detection.

\subsection{Training details}~\label{apn:weather-training}

The regression models are optimized with the loss function \texttt{RMSEWithUncertainty}~\cite{malinin2021gdbt} that predicts mean and variance of the normal distribution by optimizing the negative log-likelihood. Each model is constructed with a depth of 8 and then is trained for 20,000 iterations at a learning rate of 0.3. The classification models are optimized with the loss function \texttt{MultiClass} that predicts a discrete probability distribution over all classes. Each model is constructed with a depth of 6 and then is trained for 10,000 iterations at a learning rate of 0.4. Hyperparameter tuning is performed on the \texttt{dev\_in} data for both tasks. All models were trained within under 8 hours using a normal laptop.

\subsection{Additional experiments}\label{apn:tab-results}

In addition to considering ensembles of GBDT models implemented in CatBoost, we additionally consider ensembles of neural models. Specifically, we consider the FT-Transformer model~\cite{gorishniy2021revisiting}. We use FT-Transformers as the basis for Monte-Carlo Dropout Ensembles (MCDP)~\cite{Gal2016Dropout} as well as Deep Ensembles~\cite{deepensemble2017}. Additionally, we consider combining ensembles of CatBoost models with a Deep Ensemble of FT-Transformer models. Predictive performance figures are presented in table~\ref{tab:tabular_res_a_extra}. Here, we can see that ensembles of CatBoost models and Deep ensembles of FT-Transformer models have very similar performance, with the latter marginally outperforming the former. However, their combination yields the most competitive figures. These results are consistent for both the classification and regression tasks. 
\begin{table}[htbp!]
    \centering
    \caption{Predictive performance for Weather prediction. Mean is quoted for the single models.}
    \label{tab:tabular_res_a_extra}
\vspace{0.1in}
\resizebox{1\textwidth}{!}{%
    \begin{tabular}{@{}ll|ccc|ccc||ccc|ccc}
        \toprule
          \multirow{3}*{Dataset} &  \multirow{3}*{Model} & \multicolumn{6}{c||}{Regression} & \multicolumn{6}{c}{Classification} \\
          & & \multicolumn{3}{c|}{RMSE $\downarrow$} &  \multicolumn{3}{c||}{MAE $\downarrow$} & \multicolumn{3}{c|}{Accuracy (\%) $\uparrow$} &  \multicolumn{3}{c}{Macro F1 (\%) $\uparrow$} \\
          && In & Shifted & Full & In & Shifted & Full & In & Shifted & Full & In & Shifted & Full \\ 
        \midrule\midrule
              \multirow{5}*{\texttt{dev}}   
     & CatBoost, Single & 1.59 & 2.30&  1.98 & 1.18 & 1.75 & 1.47 & 67.0 & 47.5 & 57.2  & 42.2 & 20.2 & 36.8 \\
     & CatBoost, Ensemble & 1.51 & 2.12& 1.84 & 1.11 & 1.61& 1.36& 68.5 &50.3 & 59.4 & 42.3 & 21.3  & 37.2 \\
     & FT-Transformer, Single & 1.61 &  2.13& 1.89& 1.18 & 1.61& 1.39& 67.2 & 49.4  & 58.3 & 39.4 & 20.7 & 34.9 \\
     & FT-Transformer, MCDP & 1.59 & 2.09&  1.84& 1.16 &1.58 & 1.37 & 67.2 &  50.0  & 58.6 & 39.3 & 21.4 & 34.9 \\
     & FT-Transformer, Ensemble & 1.50 &  \textbf{2.01} & 1.77& 1.10 & \textbf{1.52}& 1.31 & 68.8 & \textbf{51.5} & 60.2 & 40.5 & \textbf{21.6}& 36.0\\
     & CatBoost $\oplus$ FT-Transformer & \textbf{1.47} &  \textbf{2.01} & \textbf{1.76}  & \textbf{1.08} & 1.53& \textbf{1.30} & \textbf{69.3} & \textbf{51.5}  &  \textbf{60.4}& \textbf{42.4} & 21.4& \textbf{37.3}\\
     \midrule
              \multirow{5}*{\texttt{eval}}   
     & CatBoost, Single & 1.60 & 2.60   &    2.16 & 1.19  &  1.91  & 1.56   & 66.7  &  44.5  & 55.5   & 42.9  &   21.5 &  34.4  \\
     & CatBoost, Ensemble & 1.52  &  2.37  & 2.00   & 1.11  &  1.75  &  1.44  & 68.2  &  46.7   & 57.3   & 44.1  & 22.2   &    35.5 \\
     & FT-Transformer, Single & 1.62  & 2.40   &   2.05  & 1.18  &  1.77  &  1.48  & 67.0  &  45.9  & 56.3   & 37.6  &  23.0   &   31.4  \\
     & FT-Transformer, MCDP & 1.59  & 2.34   & 2.01   & 1.17  & 1.73   &   1.45  & 67.0  &   46.4  &  56.6  & 37.6  &  23.2  &  31.6   \\
     & FT-Transformer, Ensemble & 1.51  &  \textbf{2.24}  & 1.92   & 1.10  &  \textbf{1.66}   & \textbf{1.38}   & 68.6  &  \textbf{48.0}  &  58.1  & 38.2  &  \textbf{23.8}  &  32.1  \\
     & CatBoost $\oplus$ FT-Transformer & \textbf{1.48}  &  2.25  &  \textbf{1.91}  & \textbf{1.08}  &  \textbf{1.66}  &  \textbf{1.38}  & \textbf{69.0}  & \textbf{48.0}   & \textbf{58.4}   & \textbf{44.2}  &  22.3   &    \textbf{35.6}  \\
        \bottomrule
    \end{tabular}
}
\end{table}

We jointly assess robustness and uncertainty quality for the additional baselines in the table below. Again, the result show that combining all models yields the best results. Curiously, the results also show that Monte-Carlo dropout ensembles are now competitive with CatBoost ensembles. This suggests that the uncertainty quality of MCDP is better than for CatBoost ensembles, even if CatBoost has the better raw predictive quality.
\begin{table}[htbp!]
    \centering
    \caption{Retention performance for Weather prediction. Mean is quoted for the single models.}
    \label{tab:tabular_res_b_extra}
\vspace{0.1in}
\resizebox{1\textwidth}{!}{%
    \begin{tabular}{@{}ll|ccc|ccc}
        \toprule
        Dataset &  Model & \multicolumn{3}{c|}{Regression} & \multicolumn{3}{c}{Classification} \\
          & & \text{R-AUC} $\downarrow$ &  \text{F1-AUC (\%)} $\uparrow$ &  \text{F1@}$95$\% $\uparrow$  & \text{R-AUC} $\downarrow$ &  \text{F1-AUC (\%)} $\uparrow$ &  \text{F1@}$95$\% $\uparrow$ \\
        \midrule\midrule
          \multirow{5}*{\texttt{dev}}   
     & CatBoost, Single & 1.894 & 44.35 & 62.72 & 0.1666 & 57.72 & 73.04 \\
     & CatBoost, Ensemble & 1.227 & 52.20 & 65.83  & 0.1522 & 59.07 & 74.86 \\
     & FT-Transformer, Single & 1.245 & 51.69 & 65.08 & 0.1592 & 58.51 & 73.80 \\
     & FT-Transformer, MCDP & 1.197 & 52.08 & 65.62 & 0.1565 & 58.80 & 74.16 \\
     & FT-Transformer, Ensemble & 1.051 & 53.66 & \textbf{67.56}  & 0.1472 & 59.54 & 75.38 \\
& CatBoost $\oplus$ FT-Transformer & \textbf{1.035} & \textbf{54.04} & 67.47 & \textbf{0.1453} & \textbf{59.71} & \textbf{75.58} \\
     \midrule
              \multirow{5}*{\texttt{eval}}   
     & CatBoost, Single & 2.320 & 43.41 & 61.89 & 0.1799 & 56.25 & 71.56 \\
     & CatBoost, Ensemble & 1.335 & 52.36 & 64.72 & 0.1640 & 58.22 & 73.17 \\
     & FT-Transformer, Single & 1.386 & 51.86 & 63.96 & 0.1705 & 57.72 & 72.17 \\
     & FT-Transformer, MCDP & 1.321 & 52.29 & 64.57 & 0.1676 & 58.04 & 72.55 \\
     & FT-Transformer, Ensemble & 1.168 & 53.77 & \textbf{66.40} & 0.1576 & 58.95 & 73.84 \\
     & CatBoost $\oplus$ FT-Transformer & \textbf{1.151} & \textbf{54.09} & 66.28 & \textbf{0.1561} & \textbf{59.07} & \textbf{74.02} \\
        \bottomrule
    \end{tabular}
}
\end{table}

Finally, we examine the quality of different uncertainty measures which are derivable from all of the baseline models. The results are provided in Table~\ref{tab:tabular_res_c_extra}. The results show an interesting trend, where the model which has the best joint uncertainty and robustness performance is a combination of CatBoost and FT-Transformer ensembles, and the best measure of uncertainty is total variance and confidence for regression and classification, respectively. Both are measures of \emph{total uncertainty}. At the same time, the best model for anomaly detection is a catboost ensemble using measures of \emph{knowledge uncertainty}. This highlights how the best model and uncertainty measure to use greatly depends on the task.
\begin{table}[htbp!]
\caption{Comparing ensembled F1-AUC and ROC-AUC for various uncertainty measures on the tests sets from the canonical partitioning of Weather Prediction dataset for regression and classification.}
\centering
\resizebox{1\textwidth}{!}{
    \begin{tabular}{c|ll|ccc|ccccc}
    \toprule
  \multirow{3}*{Data} & \multirow{3}*{Metric} & \multirow{3}*{Model} & \multicolumn{3}{c|}{Regression} & \multicolumn{5}{c}{Classification} \\
    &  & & \multicolumn{1}{c}{Total Unc.} & \multicolumn{2}{c|}{Knowledge Unc.} & \multicolumn{2}{c}{Total Unc.} & \multicolumn{3}{c}{Knowledge Unc.} \\ 
   & &   & \multicolumn{1}{c}{tvar} & \multicolumn{1}{c}{varm} & \multicolumn{1}{c|}{EPKL} &  \multicolumn{1}{c}{Conf} & \multicolumn{1}{c}{Entropy} & \multicolumn{1}{c}{MI} & \multicolumn{1}{c}{EPKL} & \multicolumn{1}{c}{RMI}\\
   \midrule
\multirow{6}*{\texttt{dev}}  
   & \multirow{3}*{F1-AUC (\%) $\uparrow$} & CatBoost& 52.20 & 50.12 & 50.51 & 59.07 & 58.86 &  57.72 & 57.69& 57.66 \\
    &  & FT-Transformer&  53.66 & 51.86 & 53.53 & 59.54 & 59.13 & 56.36  & 56.28 & 56.20 \\
      &  & CatBoost $\oplus$ FT-Transformer & \textbf{54.04} & 52.22 & 51.49 & \textbf{59.71} & 59.26 & 57.65 & 57.49 & 57.36 \\
      \cmidrule{2-11}
   & \multirow{3}*{ROC-AUC (\%) $\uparrow$} & CatBoost& 62.96 & 82.31 & \textbf{85.29}  & 63.98 & 65.00 &  83.75 & 83.96 & \textbf{84.12} \\
    & & FT-Transformer& 58.10 & 65.89 & 61.63 & 35.46 & 65.48 & 71.89  & 71.85 & 71.79 \\
    &  & CatBoost $\oplus$ FT-Transformer & 62.73 & 76.63 & 83.29 & 34.63 & 66.10 & 80.46  & 80.10 & 79.78 \\
   \midrule
\multirow{6}*{\texttt{eval}}  
   & \multirow{3}*{F1-AUC (\%) $\uparrow$} & CatBoost & 52.36 & 49.81 & 50.40  & 58.22 &  57.89 & 56.99   & 56.96& 56.93 \\
     & & FT-Transformer & 53.77 & 51.83 & 53.58 & 58.95 & 58.55 & 55.68 & 55.59 & 55.51 \\
     & &  CatBoost $\oplus$ FT-Transformer & \textbf{54.09} & 52.12 & 51.44 & \textbf{59.07} & 58.62 & 56.92 & 56.75 & 56.59 \\
     \cmidrule{2-11}
   & \multirow{3}*{ROC-AUC (\%) $\uparrow$} & CatBoost & 65.99 & 78.32 & \textbf{79.90} & 66.20 & 66.76 & 83.44  &83.59 &  \textbf{83.68}\\
   & & FT-Transformer & 65.03 & 68.78 & 67.67 & 30.68 & 70.37 &76.46 & 76.43 & 76.36 \\
   & & CatBoost $\oplus$ FT-Transformer& 67.78 & 75.43 & 79.29 & 30.86 & 69.92 & 82.49  & 82.16 & 81.85  \\
   \bottomrule
    \end{tabular}
}
    \label{tab:tabular_res_c_extra}
\end{table}

\begin{figure}[htbp!]
     \centering
    \begin{subfigure}[b]{0.49\textwidth}
         \centering
         \includegraphics[width=\textwidth]{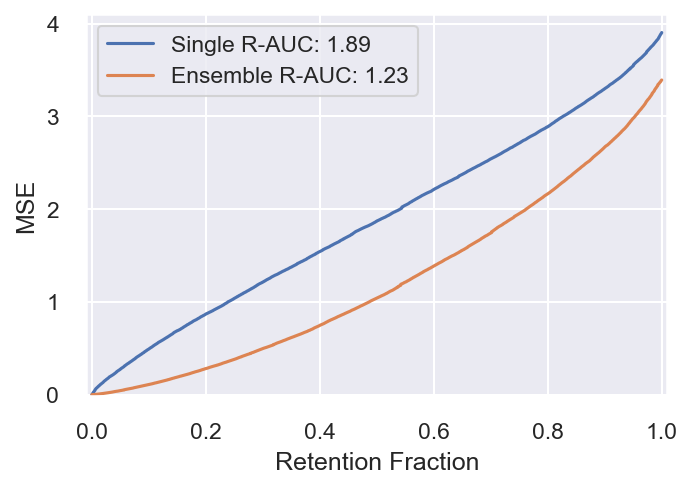}
         \caption{CatBoost, Regression, MSE.}
         \label{fig-apn:cat-eval_mse}
     \end{subfigure}
     \begin{subfigure}[b]{0.49\textwidth}
         \centering
         \includegraphics[width=\textwidth]{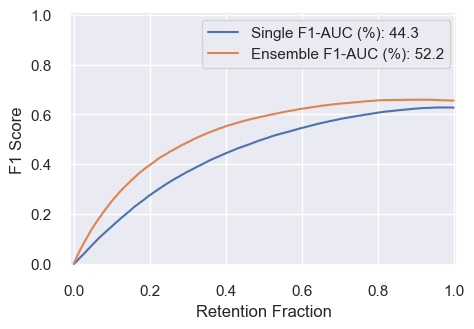}
         \caption{CatBoost, Regression, F1.}
         \label{fig-apn:cat-dev_f1}
     \end{subfigure}
     \\
     \begin{subfigure}[b]{0.49\textwidth}
         \centering
         \includegraphics[width=\textwidth]{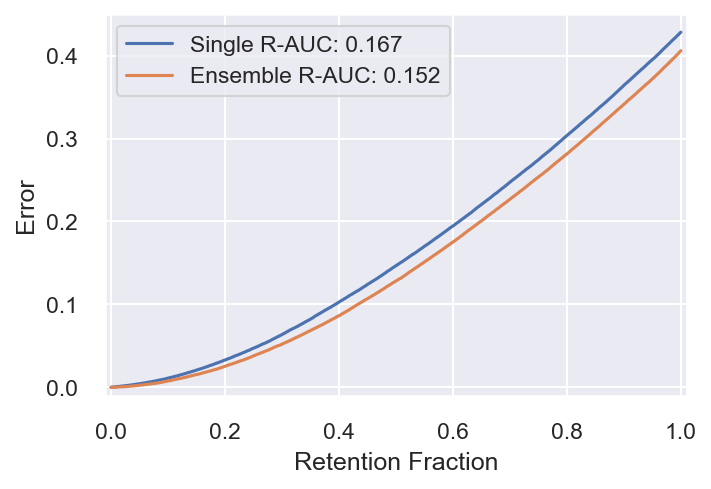}
         \caption{CatBoost, Classification, error rate.}
         \label{fig-apn:cat-dev_acc}
     \end{subfigure}
     \begin{subfigure}[b]{0.49\textwidth}
         \centering
         \includegraphics[width=\textwidth]{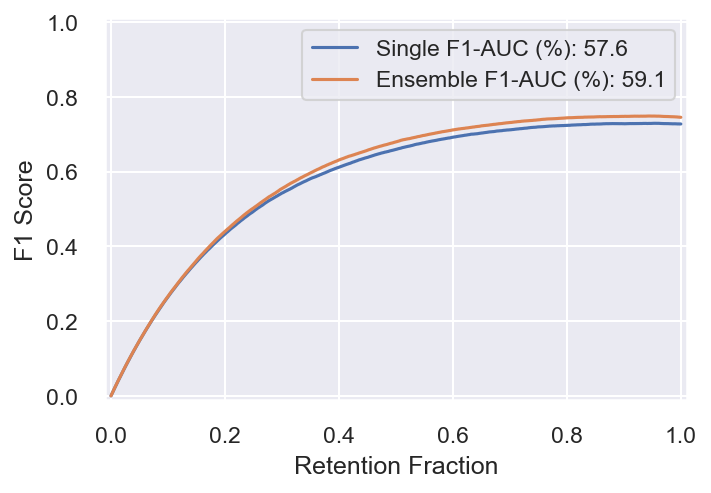}
         \caption{CatBoost, Classification, F1.}
         \label{fig-apn:dev_f1_class}
     \end{subfigure} 
     \begin{subfigure}[b]{0.49\textwidth}
         \centering
         \includegraphics[width=\textwidth]{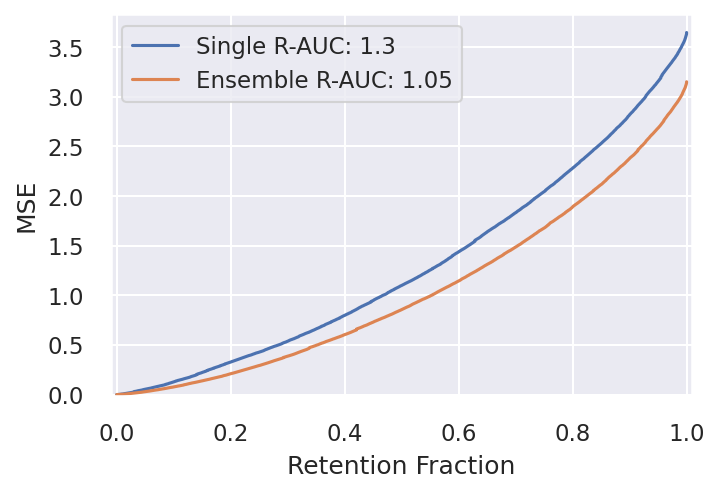}
         \caption{FT-Trans, Regression, MSE.}
         \label{fig-apn:dev_mse_ftt}
     \end{subfigure}
     \begin{subfigure}[b]{0.49\textwidth}
         \centering
         \includegraphics[width=\textwidth]{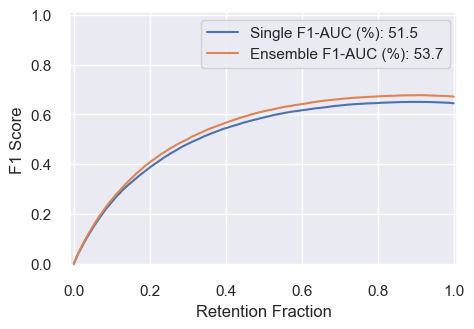}
         \caption{FT-Trans, Regression, F1.}
         \label{fig-apn:dev_f1_ftt}
     \end{subfigure}
     \\
     \begin{subfigure}[b]{0.49\textwidth}
         \centering
         \includegraphics[width=\textwidth]{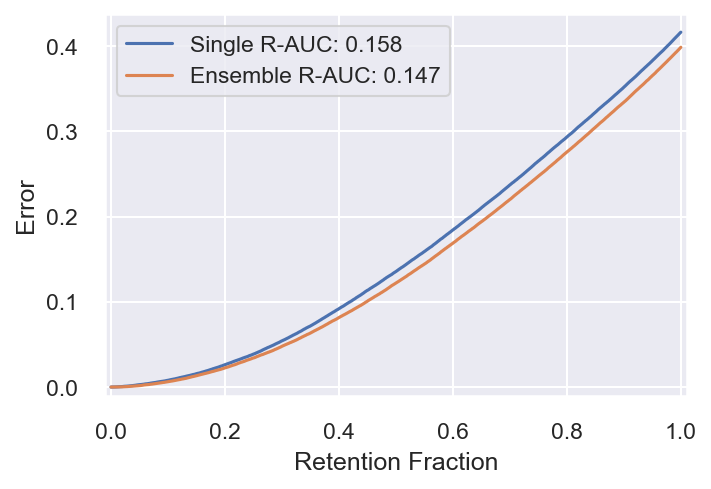}
         \caption{FT-Trans, Classification, error rate.}
         \label{fig-apn:dev_acc_ftt}
     \end{subfigure}
     \begin{subfigure}[b]{0.49\textwidth}
         \centering
         \includegraphics[width=\textwidth]{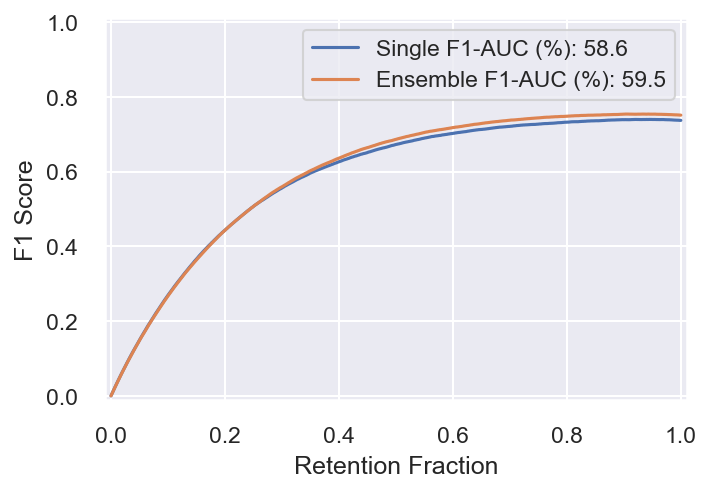}
         \caption{FT-Trans, Classification, F1.}
         \label{fig-apn:dev_f1_class_ftt}
     \end{subfigure} 
        \caption{Retention curves for CatBoost and FT-Transformer on \texttt{dev} for the canonical Weather prediction dataset.}
        \label{fig-apn:weather_retention}
\end{figure}

\begin{figure}[htbp!]
     \centering
          \begin{subfigure}[b]{0.49\textwidth}
         \centering
         \includegraphics[width=\textwidth]{figures/weather_single_retention_eval_mse.png}
         \caption{CatBoost, Regression, MSE.}
         \label{fig-apn:eval_mse}
     \end{subfigure}
     \begin{subfigure}[b]{0.49\textwidth}
         \centering
         \includegraphics[width=\textwidth]{figures/weather_single_retention_eval_f1.png}
         \caption{CatBoost, Regression, F1.}
         \label{fig-apn:eval_f1}
     \end{subfigure}
     \\
     \begin{subfigure}[b]{0.49\textwidth}
         \centering
         \includegraphics[width=\textwidth]{figures/weather_single_retention_eval_acc_class.png}
         \caption{CatBoost, Classification, error rate.}
         \label{fig-apn:eval_acc}
     \end{subfigure}
     \begin{subfigure}[b]{0.49\textwidth}
         \centering
         \includegraphics[width=\textwidth]{figures/weather_single_retention_eval_f1_class.png}
         \caption{CatBoost, Classification, F1.}
         \label{fig-apn:eval_accf1}
     \end{subfigure} 
     \begin{subfigure}[b]{0.49\textwidth}
         \centering
         \includegraphics[width=\textwidth]{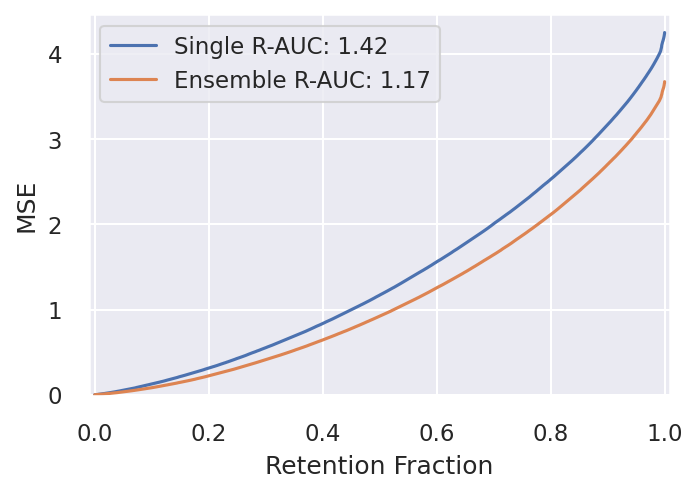}
         \caption{FT-Trans, Regression, MSE.}
         \label{fig-apn:eval_mse_ftt}
     \end{subfigure}
     \begin{subfigure}[b]{0.49\textwidth}
         \centering
         \includegraphics[width=\textwidth]{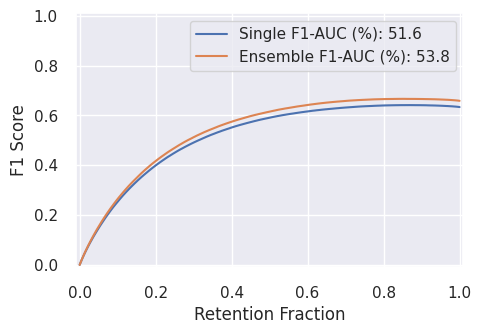}
         \caption{FT-Trans, Regression, F1.}
         \label{fig-apn:eval_f1_ftt}
     \end{subfigure}
     \\
     \begin{subfigure}[b]{0.49\textwidth}
         \centering
         \includegraphics[width=\textwidth]{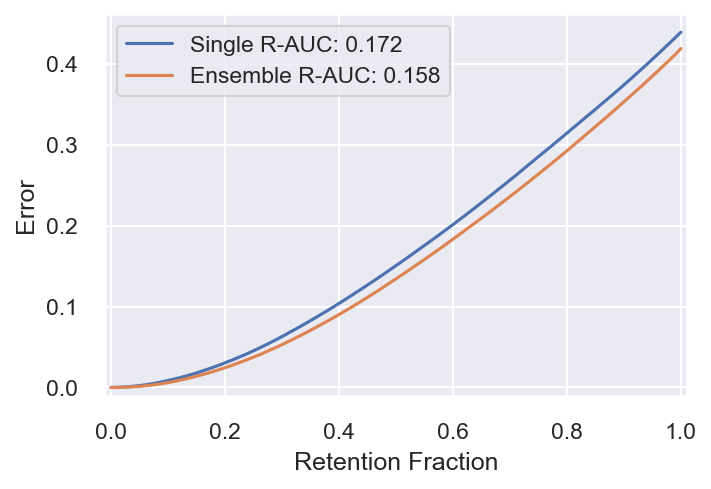}
         \caption{FT-Trans, Classification, error rate.}
         \label{fig-apn:eval_acc_ftt}
     \end{subfigure}
     \begin{subfigure}[b]{0.49\textwidth}
         \centering
         \includegraphics[width=\textwidth]{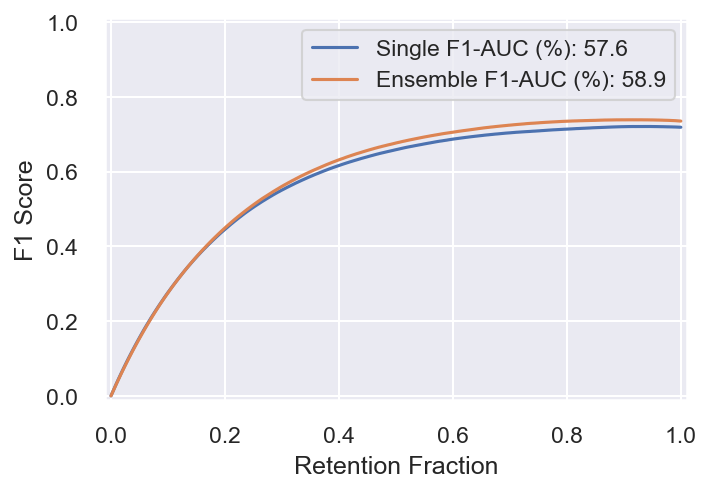}
         \caption{FT-Trans, Classification, F1.}
         \label{fig-apn:eval_accf1_ftt}
     \end{subfigure} 
        \caption{Retention curves with CatBoost and FT-Transformer on \texttt{eval} for the canonical Weather prediction dataset.}
        \label{fig-apn:weather_retention_eval}
\end{figure}

\subsubsection{Further experiments}

\begin{figure}[htbp!]
\centering
    \includegraphics[width=0.9\linewidth]{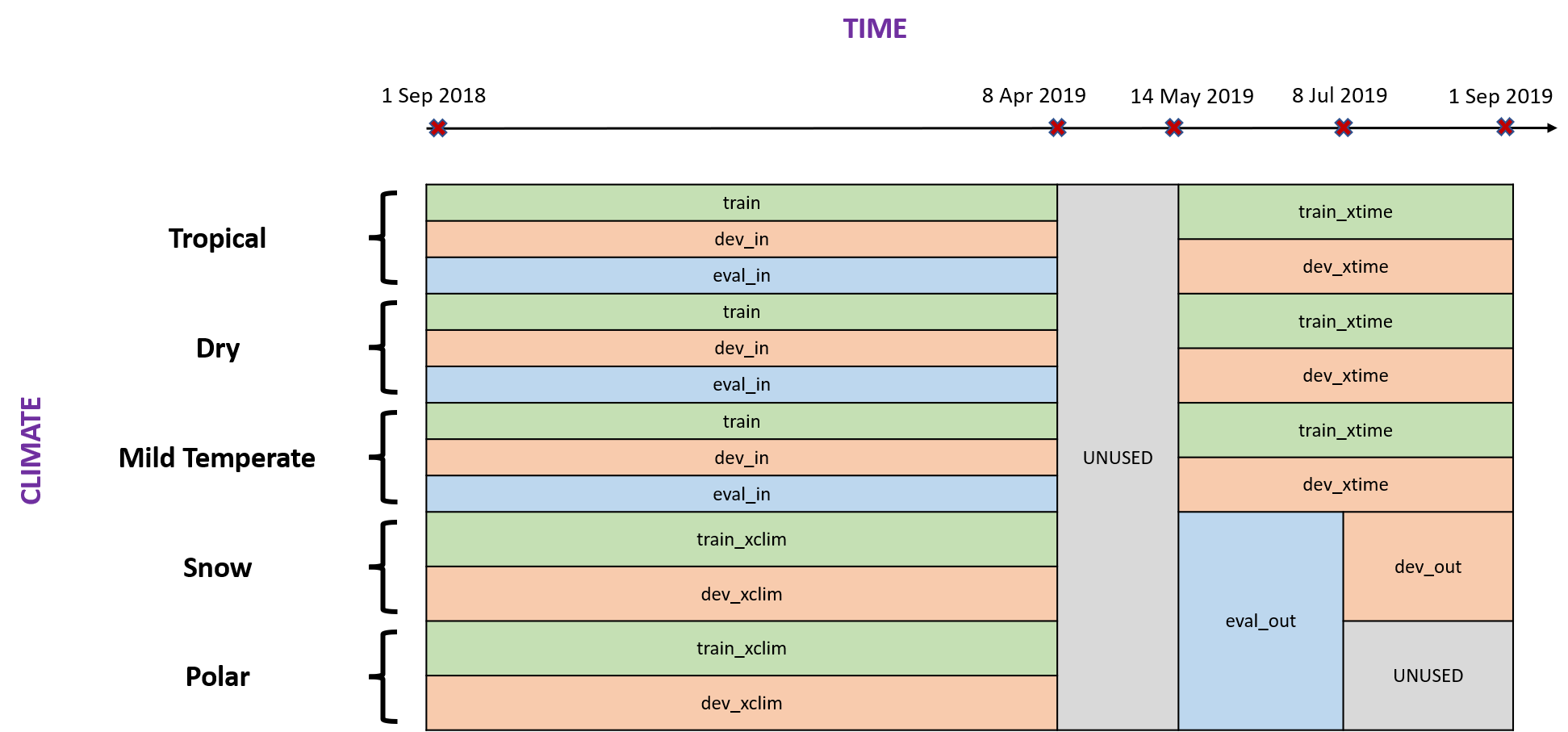}
  \caption{Extended splits of tabular weather data.}
  \label{tab:weather_splits_aug} 
\end{figure}

Figure \ref{tab:weather_splits_aug} depicts additional splits beyond the canonical partition of the tabular weather data. Table \ref{tab:weather_extra_exps} summarises the experiments to be performed with a brief description of what each experiment involves. All experiments are to be performed using CatBoost for both the regression and classification tasks. These experiments aim to better understand whether time or climate shift in the data leads to a greater performance drop from in-domain to shifted datasets. Hence, the focus here is on robustness only. The corresponding results for each experiment are given in Table \ref{tab:weather_res_a_extra}.

\begin{table}[htbp!]
\caption{Description of additional experiments.}
\centering
\begin{small}
    \begin{tabular}{c|l|l|l}
    \toprule
Exp & Training set & Development set & Description \\
\midrule
A & \texttt{train} & \texttt{dev\_in} & Time \& climate shifts \\ 
B & \texttt{train} $\oplus$ \texttt{train\_xclim} & \texttt{dev\_in} $\oplus$ \texttt{dev\_xclim} & Time shift \\
C & \texttt{train} $\oplus$ \texttt{train\_xtime} & \texttt{dev\_in} $\oplus$ \texttt{dev\_xtime} & Climate shift \\
D & \texttt{train} $\oplus$ \texttt{train\_xclim} $\oplus$ \texttt{train\_xtime} & \texttt{dev\_in} $\oplus$ \texttt{dev\_xclim} $\oplus$ \texttt{dev\_xtime} & No shift \\ 
   \bottomrule
    \end{tabular}
    \end{small}
    \label{tab:weather_extra_exps}
\end{table}

\begin{table}[htbp!]
    \centering
    \caption{Predictive performance for Weather prediction using different training sets. Mean is quoted for the single models.}
    \label{tab:weather_res_a_extra}
\vspace{0.1in}
\resizebox{1\textwidth}{!}{%
    \begin{tabular}{@{}ll|cccc|cccc||cccc|cccc}
        \toprule
          \multirow{3}*{Dataset} &  \multirow{3}*{Model} & \multicolumn{8}{c||}{Regression} & \multicolumn{8}{c}{Classification} \\
          & & \multicolumn{4}{c|}{RMSE $\downarrow$} &  \multicolumn{4}{c||}{MAE $\downarrow$} & \multicolumn{4}{c|}{Accuracy (\%) $\uparrow$} &  \multicolumn{4}{c}{Macro F1 (\%) $\uparrow$} \\
          && A & B & C & D & A & B & C & D & A & B & C & D & A & B & C & D  \\ 
        \midrule\midrule
              \multirow{2}*{\texttt{dev\_in}}   
     & CatBoost, Single & 1.59&1.62&1.61& 1.63 &1.18&1.21&1.20&1.21&67.0& 66.1&66.6 &  65.8&42.2&39.6&42.6&39.9 \\
     & CatBoost, Ensemble & 1.51&1.52&1.51& 1.54 &1.11&1.12&1.11&1.14&68.5&67.2&67.7&66.8&42.3&40.1&41.9&39.9 \\
     \midrule
              \multirow{2}*{\texttt{dev\_out}}   
     & CatBoost, Single & 2.30&2.30&2.04&1.95 &1.75&1.75&1.54&1.48&47.5&50.9&51.8&54.0&20.2&21.2&21.4& 22.7\\
     & CatBoost, Ensemble & 2.12&2.05&1.93&1.85 &1.61&1.55&1.45&1.40&50.3&53.4&53.0&55.4&21.3&22.2&21.8&22.9 \\
     \midrule
     \midrule
              \multirow{2}*{\texttt{eval\_in}}   
     & CatBoost, Single & 1.60 &1.63&1.62&1.64& 1.19&1.21&1.20&1.22&66.7&65.9&66.3&65.7&42.9&40.4&42.7&40.3 \\
     & CatBoost, Ensemble & 1.52&1.53&1.52&1.55 &1.11&1.12&1.12&1.14&68.2&67.1&67.6&66.7&44.1&42.0&43.8&41.3 \\
     \midrule
              \multirow{2}*{\texttt{eval\_out}}   
     & CatBoost, Single & 2.60 &2.62&2.28&2.15&1.91&1.93&1.69&1.62&44.5&48.3&48.6&51.5&21.5&23.8&23.5&25.6 \\
     & CatBoost, Ensemble & 2.37 &2.26&2.16&2.04&1.75&1.69&1.60&1.53&46.7&50.4&50.2&53.0&22.2&24.1&24.1&26.0 \\
        \bottomrule
    \end{tabular}
}
\end{table}

\FloatBarrier
\section{Machine Translation}\label{apn:nmt}

The current appendix contains a description of the composition, collection, pre-processing and partitioning of the Shifts Machine Translation dataset. Additionally, it contains a description of the metrics used for assessment and an expanded set of experimental results.

\subsection{Dataset Description} \label{apn:nmt-data}

\paragraph{Composition} The Shifts Machine Translation datasets consists of a training, development (dev) and evaluation (eval) set. Each set consists of pairs of source and target sentences in English and Russian, respectively. As most production NMT systems are built using a variety of general purpose corpora, we do not provide a new training corpus, rather, we will use the freely available WMT'20 English-Russian corpus. This data covers a variety of domains, but primarily focuses on parliamentary and news data. For the most part, this data is grammatically and orthographically correct and language use is formal. This is representative of the type of data used, for example, to build the Yandex.Translate NMT system. The composition of the WMT'20 En-Ru corpus is detailed on the workshop for machine translation website here: \url{http://www.statmt.org/wmt20/translation-task.html}. For simplicity of access and archiving purposes we downloaded the WMT'20 En-Ru training data set and also made it available on the Shifts Dataset and Challenge GitHub here: \url{https://github.com/yandex-research/shifts}.

The dev and eval datasets consist of an ``in-domain'' partition matched to the training data, and an ``out-of-distribution'', or shifted partition, which contains examples of atypical language usage. We select the English-Russian Newstest'19 as the in-domain \emph{development set} and will use a new corpus of news data collected from GlobalVoices News service~\cite{globalvoices} and manually annotated using expert human translators as the in-domain \emph{evaluation set}. For the shifted development and evaluation data we use the Reddit corpus prepared for the WMT'19 robustness challenge~\cite{michel2018mtnt}. This data contains examples of slang, acronyms, lack of punctuation, poor orthography, concatenations, profanity, and poor grammar, among other forms of atypical language usage. This data is representative of the types of inputs that machine translation services find challenging. As Russian target annotations are not available, we pass the data through a two-stage process, where orthographic, grammatical and punctuation mistakes are corrected, and the source-side English sentences are translated into Russian by expert in-house Yandex translators. The development set is constructed from the same 1400-sentence test-set used for the WMT'19 robustness challenge. For the heldout evaluation set we use the open-source MTNT crawler which connects to the Reddit API to collect a further set of 3,000 English sentences from Reddit, which is similarly corrected and translated. Note that the Reddit data has comments made by users, but no personal identification data (login, name, etc...) or other user identification data was recorded or stored - the dataset only only contains the raw comments made on a public discussion platform. In terms of size, these development and evaluation sets are comparable or larger to the ones used in the WMT challenges and for evaluating productions systems.
\begin{table}[htbp]
  \caption{NMT Data Description - All Data is English-Russian}
  \begin{center}
  \begin{small}
    \begin{tabular}{ l |c c c l }
      \toprule
      \multirow{2}*{Data Set} &  \multirow{2}*{N. Sentences} & \multicolumn{2}{c}{Avg. Sentence Length}  & \multirow{2}*{Type} \\ 
       & & En & Ru & \\
      \midrule
        WMT'20 & 62M & 23,9 & 20.9 & Train \\
        NWT'19 & 1997 & 24.5 & 24.7 & In-domain Dev \\
        GlobalVoices & 3,000 & 25.1 & 24.1& In-domain Eval \\
        WMT'19 MTNT Reddit & 1,362 & 17.2 & 16.5 &  Shifted Dev \\
        Shifts Reddit & 3,063 & 16.1 & 16.4 & Shifted Eval \\
      \bottomrule
    \end{tabular}
  \end{small}
  \end{center}
\end{table}

Both the development and evaluation Reddit data was manually annotated by members of the Yandex.Translate team with the following 7 non-exclusive anomaly flags:
\begin{itemize}
    \item \textbf{Punctuation anomalies}: Some punctuation marks are missed or used incorrectly or some formatting (like Wiki markup) is used in the sentence.
    \item \textbf{Spelling anomalies}: The sentence contains spelling errors, including incorrect concatenation of two words as well as incorrect use of hyphens.
    \item \textbf{Capitalization anomalies}: Words that should be capitalized according to the language rules are written in lower case or vice versa.
    \item \textbf{Fluency anomalies}: The sentence is non-fluent due to wrong or missing prepositions, pronouns or ungrammatical form choice. 
    \item \textbf{Slang anomalies}: In the sentence there are slang words of abbreviations like “idk” for “I don’t know” or “cuz” for “because”.
    \item \textbf{Emoji anomalies}: The sentence contains emojis either at the end of it, or instead of some words.
    \item \textbf{Tags anomalies}: The sentence contains markup for usernames or code like “r/username”.
\end{itemize}

\begin{figure}[htbp!]
     \centering
     \begin{subfigure}[b]{0.49\textwidth}
         \centering
         \includegraphics[width=\textwidth]{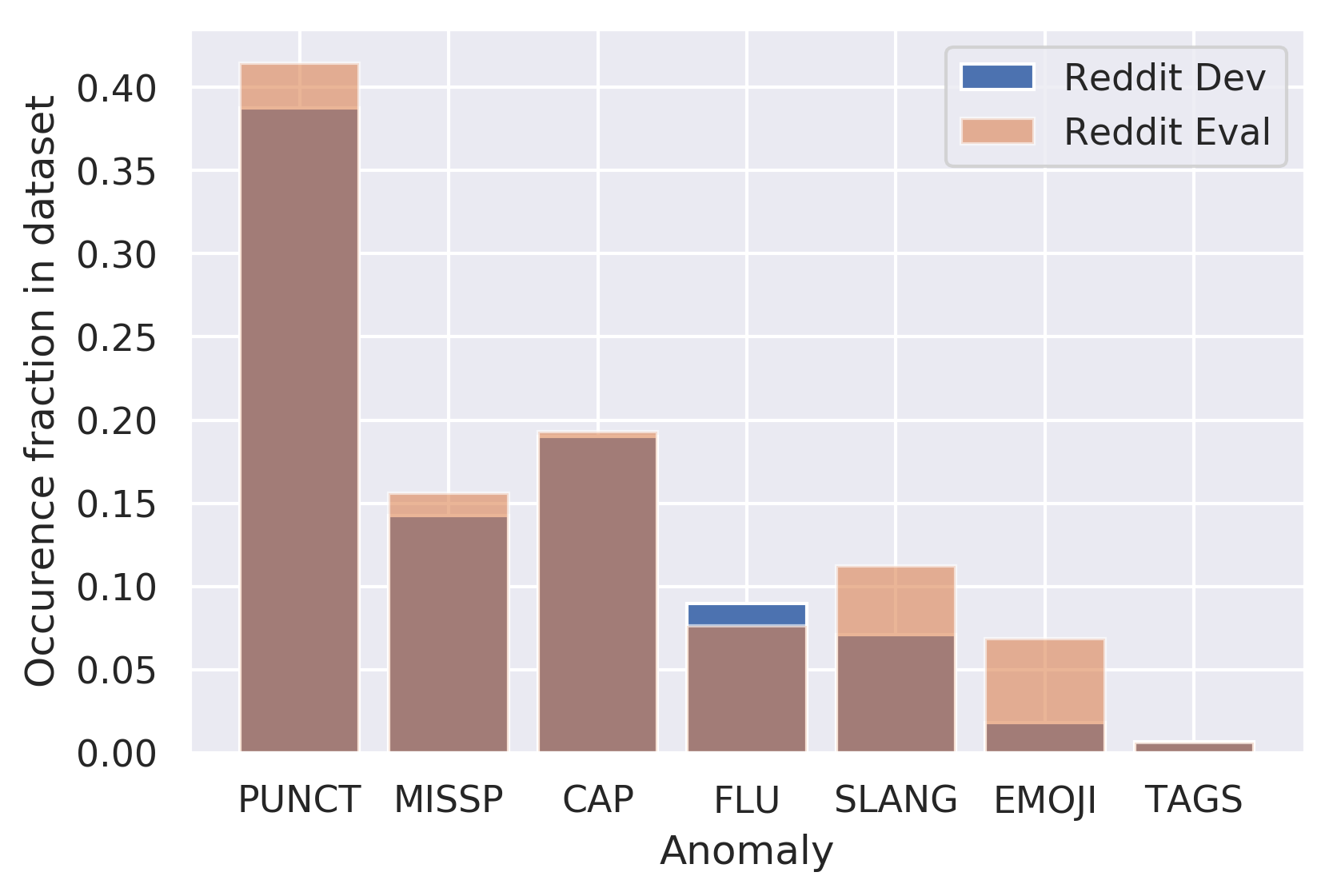}
     \end{subfigure} 
    \begin{subfigure}[b]{0.49\textwidth}
         \centering
         \includegraphics[width=\textwidth]{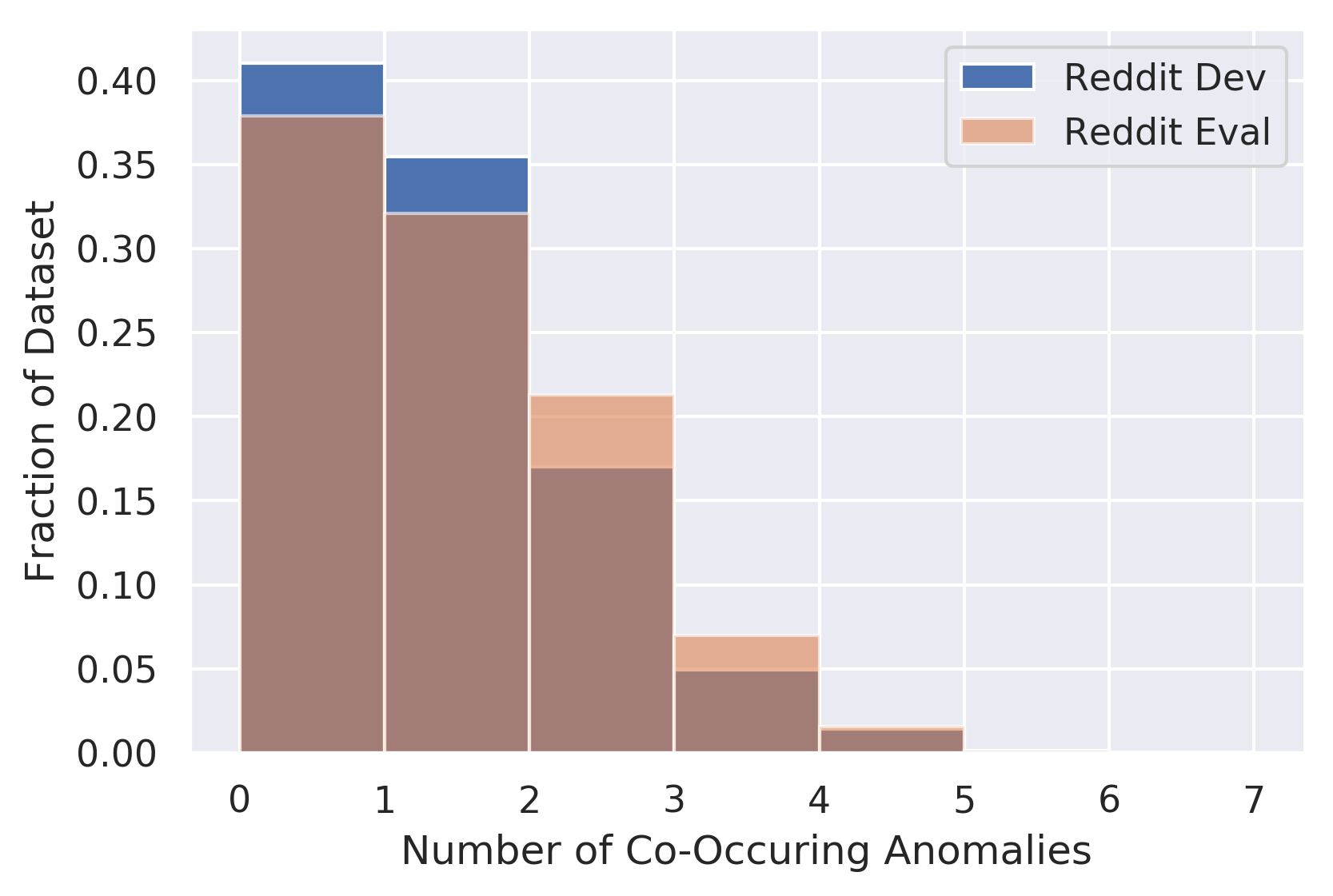}
     \end{subfigure} 
    \caption{Analysis of anomaly occurrence and co-occurrence in Reddit (shifted) development and evaluation data}
        \label{fig:nmt-anomalies}
\end{figure}

An analysis of the occurrence and co-occurrence of these anomalies is provided in figure~\ref{fig:nmt-anomalies}.

\paragraph{Collection Process} GlobalVoices\cite{globalvoices} data was crawled for parallel news articles in English and Russian using internal Yandex tools. The raw articles were manually split into sentence-pairs by in-house Yandex assessors. A full set of 30000 sentence pairs was produced, from which a subset of 3000 sentences was uniformly randomly sampled. Reddit data was crawled using the open-source MTNT~\cite{michel2018mtnt} crawler from \url{https://github.com/pmichel31415/mtnt}. This crawler links in with the Reddit API to allow mining and crawling Reddit for data. The crawler collected a set of 100K user comments which were then split into sentences using the NLTK toolkit. Then a set of 3500 sentences was randomly uniformly selected. After pre-processing and cleaning a set of 3065 sentences was produced. 

\paragraph{Preprocessing and Cleaning} For the GlobalVoices data parallel sentences markup was done manually by in-house Yandex assessors; non-parallel sentences were removed from dataset. For Reddit data 1-word phrases and sentences consisting only of non-alphabetical symbols were removed. Professional editors were used to manually correct grammatical and orthographic mistakes prior to translating into Russian, but were explicitly told to maintain the non-formal style as much as possible. This error correction was used only for obtaining target-side Russian translation.

\paragraph{Guidelines on ethical use} Users are discouraged from attempting to discover to which Reddit users the comments belong by manually or automatically crawling through Reddit to find the comments.

\paragraph{Format} This dataset is provided in raw text format and a TSV with metadata for the dev and eval reddit data.

\paragraph{License} The Shifts Machine Translation dataset is released under a mixed licence. GlobalVoices evaluation data is released under CC BY NC SA 4.0 . The source-side text for the Reddit development and evaluation datasets exist under terms of the Reddit API. The target side Russian sentences were obtained by Yandex via in-house professional translators and are released under CC BY NC SA 4.0. We highlight that the development set source sentences are the same ones as used in the MTNT dataset.

\subsection{Metrics}\label{apn:nmt-metrics}

To evaluate the performance of our models we will consider the following two metrics : corpus-level BLEU~\cite{sacrebleu} and sentence-level GLEU~\cite{napoles2015gleu,napoles2016gleu,wu2016google}. GLEU is an analogue of BLEU which is stable when computed at the level of individual sentences. Thus, it is far more useful at evaluating system performance on a per-sample basis, rather than at the level of an entire corpus. Note that GLEU correlates strongly with BLEU at the corpus level. 

Machine translation is inherently a multi-modal task, as a sentence can be translated in multiple equally valid ways. Furthermore, translation systems often yield multiple translation hypothesis. To account for this we will consider two GLEU-based metrics for evaluating translation quality. First is the \emph{expected GLEU} or \textbf{eGLEU} across all translation hypotheses returned by a translation models. Each hypothesis is assumed to be assigned a \emph{confidence score}, and confidences across each hypotheses by sum to one. This is our primary assessment metric:
\begin{equation}
    \text{eGLEU} = \frac{1}{N}\sum_{i=1}^N \sum_{h=1}^H \text{GLEU}_{i,h} \cdot w_{i,h},\quad w_{i,h} > 0, \sum_{h=1}^H w_h = 1
\label{eq:eGLEU}
\end{equation}
Additionally, we will consider the \emph{maximum GLEU} or \textbf{maxGLEU} across all hypothesis, which represents an upper bound on performance, given a model can appropriately rank it's hypotheses:
\begin{equation}
    \text{maxGLEU} = \frac{1}{N}\sum_{i=1}^N  \max_{h} \big[  \text{GLEU}_{i,h} \big]
\label{eq:maxGLEU}
\end{equation}
Finally, in order to calculate area under the error retention curve we need to introduce an \emph{error metric}, where lower error is better. This is trivially done by introducing \emph{eGLEU error}, which defined as:
\begin{equation}
    \text{eGLEU Error} = 100-\text{eGLEU}
\label{eq:eGLEU_Error}
\end{equation}

Thus, in section~\ref{sec:nmt}, area under the error retention curve (R-AUC), as well as the F1 metric for detecting `valid predictions' will be calculated using eGLEU Error.

\subsection{Training details} Training data was standard used the standard perl-based script provided in Fairseq~\cite{fairseq} examples. Duplicate sentence pairs as well as sentence pairs where source and target text matched were removed. Models were trained using Fairseq version 0.8. A full description and for from preprocessing and training is provided \href{https://github.com/yandex-research/shifts/tree/main/translation}{here}. All models were trained used 8xV100 GPUs over roughly 48 hours.

\subsection{Additional Results}

\begin{figure}[htbp!]
     \centering
     \begin{subfigure}[b]{0.49\textwidth}
         \centering
         \includegraphics[width=\textwidth]{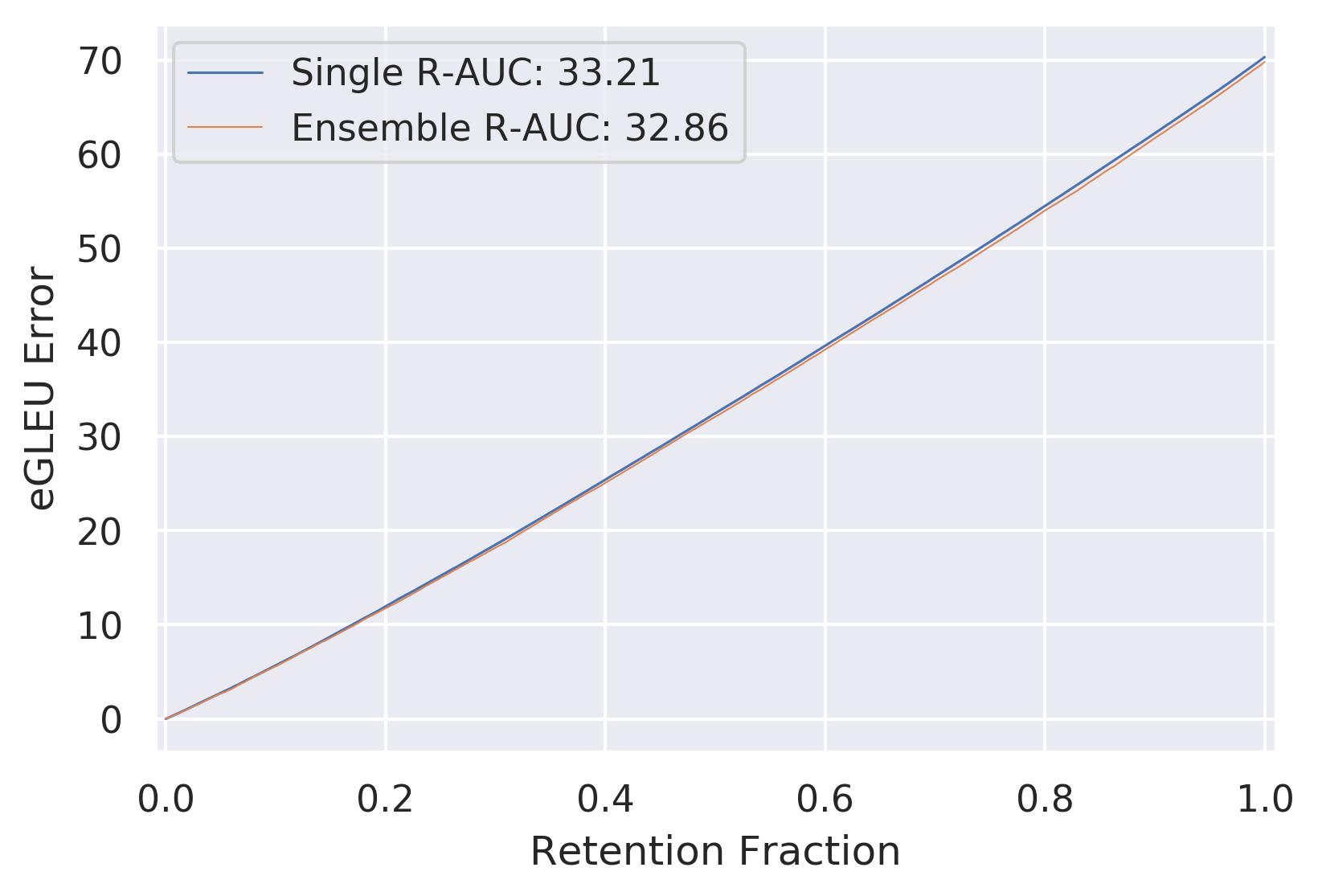}
         \caption{Dev}
     \end{subfigure}
     \begin{subfigure}[b]{0.49\textwidth}
         \centering
         \includegraphics[width=\textwidth]{figures/nmt_error_retention_eval.png}
         \caption{Eval}
     \end{subfigure}
     \\
     \begin{subfigure}[b]{0.49\textwidth}
         \centering
         \includegraphics[width=\textwidth]{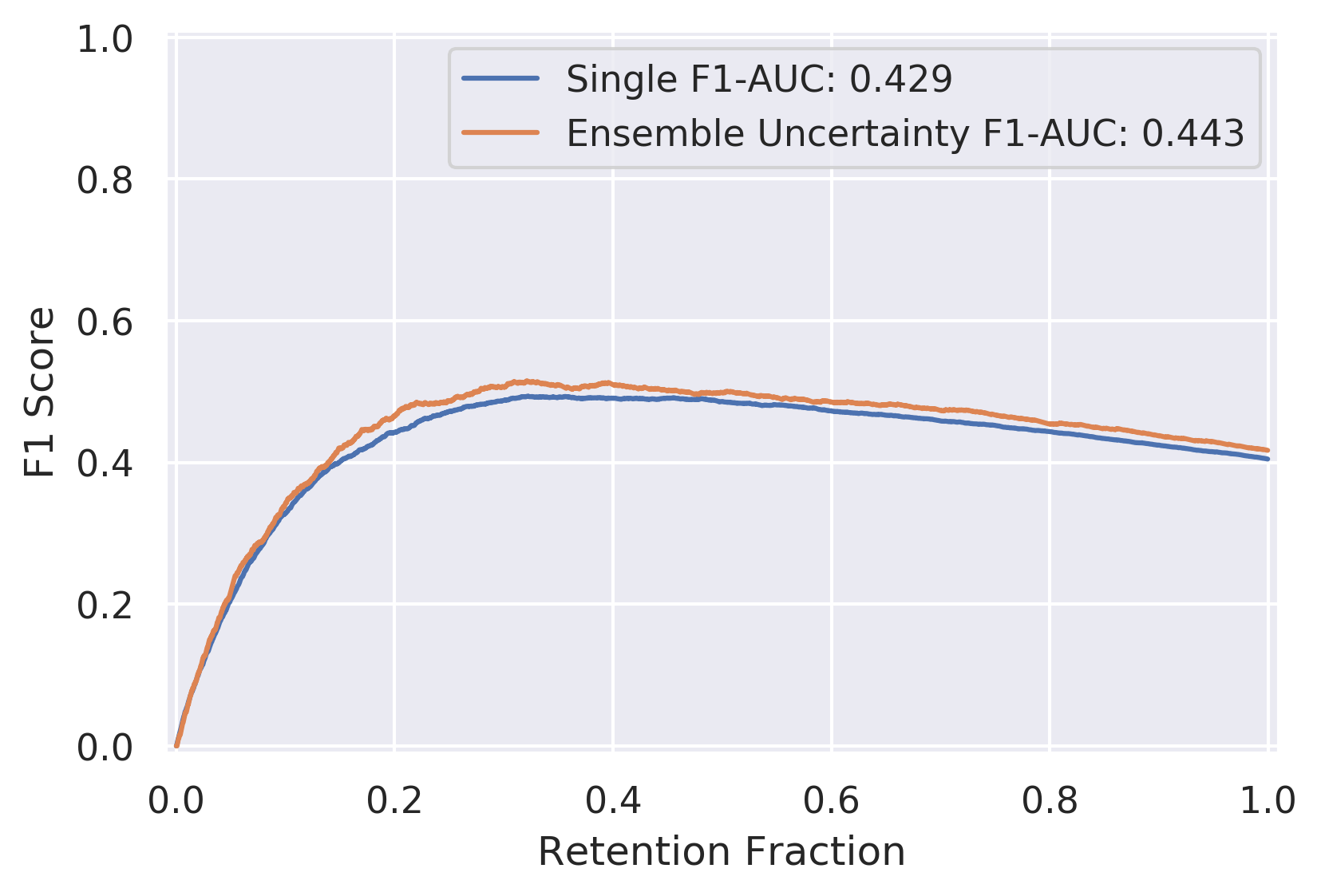}
         \caption{Dev}
     \end{subfigure}
     \begin{subfigure}[b]{0.49\textwidth}
         \centering
         \includegraphics[width=\textwidth]{figures/nmt_F1_retention_eval.png}
         \caption{Eval}
     \end{subfigure}
        \caption{Location of samples from canonical partitioning of Weather Prediction dataset.}
        \label{fig-apn:nmt_retention_curves}
\end{figure}

\newpage
\FloatBarrier
\section{Vehicle Motion Prediction}\label{apn:sdc}

The current appendix contains a description of the composition, collection, pre-processing and partitioning of the Shifts Vehicle Motion Prediction dataset. Additionally, it contains a description of the metrics used for assessment and an expanded set of experimental results.

\subsection{Dataset Description}

\begin{table}[htbp]
  \caption{A comparison of various motion prediction datasets. The Shifts Vehicle Motion Prediction dataset is the largest by number of scenes and total size in hours.}
  \label{table:sdc_dataset_comparison}
  \begin{center}
  \begin{small}
    \begin{tabular}{ l |cccccc }
      \toprule
       \multirow{2}{*}{Dataset} & \multirow{2}{*}{Scene Length (s)} &  \multicolumn{3}{c}{\# Scenes} & \multirow{2}{*}{Total Size (h)} & \multirow{2}{*}{Avg. \# Actors} \\ 
       & & Train & Dev & Eval & &  \\
       \midrule
            Argoverse  & 5 & 205,942 & 39,472 & 78,143 & 320 & 50 \\
            Lyft & 25 & 134,000 & 11,000 & 16,000 & 1,118 & 79 \\
            Waymo & 20 & 72,347 & 15,503 & 15,503 & 574 & - \\
            Shifts & 10 & 500,000 & 50,000 & 50,000 & 1,667 & 29 \\
      \bottomrule
    \end{tabular}
  \end{small}
  \end{center}
\end{table}

\paragraph{Composition} The dataset for the Vehicle Motion Prediction task was collected by the Yandex Self-Driving Group (SDG) fleet.
This is the largest vehicle motion prediction dataset released to date, containing 600,000 scenes (see \cref{table:sdc_dataset_comparison} for a comparison to other public datasets). The dataset consists of scenes spanning six locations, three seasons, three times of day, and four weather conditions (cf. \cref{table:sdc_dataset_loc_season} and \ref{table:sdc_dataset_prec_tod}). Each of these conditions is available in the form of tags associated with every scene. Each scene is 10 seconds long and is divided into 5 seconds of context features and 5 seconds of ground truth targets for prediction, separated by the time $T=0$. The goal of the task is to predict the movement trajectory of vehicles at time $T \in \left(0, 5\right]$ based on the information available for time $T \in \left[-5, 0\right]$. 
\begin{table}[htbp]
  \caption{The number of scenes in the Vehicle Motion Prediction dataset by location and season.}
  \label{table:sdc_dataset_loc_season}
  \begin{center}
  \begin{small}
    \begin{tabular}{ l |ccc }
      \toprule
Location & Train & Dev & Eval \\
      \midrule
        Moscow & 450,504 & 30,505 & 30,534 \\
        Skolkovo & 6,283 & 2,218 & 2,956 \\
        Innopolis & 15,086 & 5,164 & 5,016 \\
        Ann Arbor & 19,349 & 8,290 & 6,617 \\
        Modiin & 3,502 & 2,262 & 1,555 \\
        Tel Aviv & 5,276 & 1,561 & 3,322 \\
            \bottomrule
    \end{tabular}
  \end{small}
  \quad
    \begin{small}
    \begin{tabular}{ l |ccc }
    \toprule
        Season & Train & Dev & Eval \\
        \midrule
        Summer & 85,698 & 10,634 & 10,481 \\
        Autumn & 126,845 & 15,290 & 15,840 \\
        Winter & 287,457 & 24,076 & 23,679 \\
        Spring & 0 & 0 & 0 \\
    \bottomrule
    \end{tabular}
  \end{small}
  \end{center}
\end{table}

Each scene includes information about the state of dynamic objects (i.e., vehicles, pedestrians) and an HD map.
Each vehicle is described by its position, velocity, linear acceleration, and orientation (yaw, known up to $\pm\pi$).
A pedestrian state consists of a position vector and a velocity vector. All state components are represented in a common coordinate frame and sampled at 5Hz frequency by the perception stack running on the Yandex SDG fleet. The HD map includes lane information (e.g., traffic direction, lane priority, speed limit, traffic light association), road boundaries, crosswalks, and traffic light states, which are also sampled at 5Hz. To facilitate easy use of this dataset, we provide utilities to render scene information as a feature map, which can be used as an input to a standard vision model (e.g., a ResNet \cite{resnet}). Our utilities represent each scene as a birds-eye-view image with each channel corresponding to a particular feature (e.g., a vehicle occupancy map) at a particular timestep. We also provide pre-rendered feature maps for every prediction request (cf. \cref{subsection:task_setup}) in the dataset, which are used to train the baseline models. The maps are $128 \times 128$ pixels in size with each pixel covering 1 square meter, have $17$ channels describing both HD map information and dynamic object states at time $T = 0$, and are centered with respect to the agent for which a prediction is being made. Researchers working with the dataset are free to use these feature maps, use the provided utilities to render another set of feature maps at different (earlier) timesteps, or construct their own scene representations from the raw data.

The ground truth part of a scene contains future states of dynamic objects sampled at 5Hz for a total of $25$ state samples. Some objects might not have all $25$ states available due to occlusions or imperfections of the on-board perception system.
\begin{table}[htbp]
  \caption{The number of scenes in the Vehicle Motion Prediction dataset by precipitation and time of day.}
  \label{table:sdc_dataset_prec_tod}
  \begin{small}
    \begin{tabular}{ l |ccc }
      \toprule
Precipitation Type & Train & Dev & Eval \\
\midrule
No & 432,598 & 44,799 & 44,274 \\
Rain & 15,618 & 1,857 & 1,751 \\
Sleet & 15,210 & 1,082 & 990 \\
Snow & 36,574 & 2,262 & 2,985 \\
            \bottomrule
    \end{tabular}
  \end{small}
  \quad
    \begin{small}
    \begin{tabular}{ l |ccc }
      \toprule
Sun Phase & Train & Dev & Eval \\
\midrule
Astronomical Night & 171,867 & 13,164 & 13,113 \\
Daylight & 299,065 & 33,879 & 33,979 \\
Twilight & 29,068 & 2,957 & 2,908 \\
\bottomrule
    \end{tabular}
  \end{small}
\end{table}

A number of vehicles in the scene are labeled as \emph{prediction requests}. These are the vehicles that are visible at the most recent time $T=0$ in the context features part of a scene, and therefore would call for a prediction in a deployed system. 
For such vehicles we provide not only their future trajectories, but also a number of non--mutually exclusive tags (detailed in \cref{tab:trajectory_tags}) describing the associated maneuver in more detail -- whether the vehicle is turning, accelerating, slowing down, etc. -- for a total of $10$ maneuver types. Note that some prediction requests may not have all $25$ state samples available.
We call prediction requests with fully-observed state \emph{valid} prediction requests and propose to evaluate predictions only on those.
\begin{table}[h]
  \caption{Number of actor maneuvers of the respective type.}
  \begin{center}
  \begin{small}
    \begin{tabular}{l | ccc}
\toprule
Maneuver Type & Train & Dev & Eval \\
\midrule
Move Left & 254,843 & 25,049 & 25,820 \\
Move Right & 322,231 & 30,074 & 30,633 \\
Move Forward & 5,032,724 & 395,467 & 413,920 \\
Move Back & 54,677 & 4,811 & 4,891 \\
Acceleration & 2,473,750 & 206,977 & 215,009 \\
Deceleration & 2,050,186 & 168,550 & 174,477 \\
Uniform Movement & 6,369,920 & 566,083 & 573,033 \\
Stopping & 441,619 & 38,411 & 39,336 \\
Starting & 739,143 & 64,986 & 65,759 \\
Stationary & 4,620,678 & 433,161 & 433,576 \\
\bottomrule
    \end{tabular}
  \end{small}
  \end{center}\label{tab:trajectory_tags}
\end{table}

\begin{figure}[h!]
  \centering
  \includegraphics[width=\linewidth]{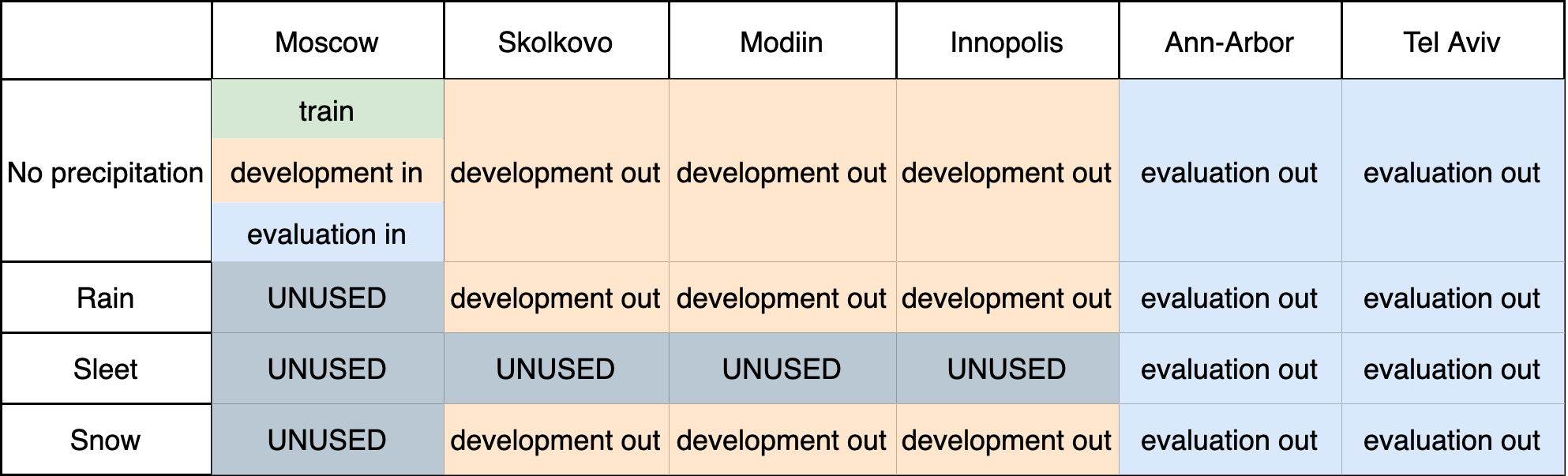}
  \caption{The canonical partitioning of the Vehicle Motion Prediction dataset.}
  \label{fig:sdc_partitioning}
\end{figure}

In order to study the effects of distributional shift, as well as assess the robustness and uncertainty estimation of baseline models, we divide the Vehicle Motion Prediction dataset such that there are \emph{in-domain} partitions which match the location and precipitation type of the training set, and \emph{out-of-domain} or \emph{shifted} partitions which do not match the training data along one or more of those axes. Furthermore, we provide a \emph{development} set which acts as a validation set, and an \emph{evaluation} set which acts as the test set. For standardized benchmarking we define a \emph{canonical partitioning} of the full dataset (cf. Figure~\ref{fig:sdc_partitioning}, \cref{table:sdc_partitioning}) as the following. The training, in-domain development, and in-domain evaluation data are taken from Moscow. Distributionally shifted development data is taken from Skolkovo, Modiin, and Innopolis. Distributionally shifted evaluation data is taken from Tel Aviv and Ann Arbor.
In addition, we remove all cases of precipitation from the in-domain training, development, and evaluation sets, while distributionally shifted datasets include precipitation.
The canonical partitioning is fully described in Figure~\ref{fig:sdc_partitioning}.
This partitioning is also the one used in the Shifts Challenge.

\begin{table}[htbp]
  \caption{The number of scenes in the canonical dataset partitioning.}
  \label{table:sdc_partitioning}
  \begin{center}
  \begin{small}
    \begin{tabular}{ l | cc }
      \toprule
      Dataset Partition & In-Distribution & Distributionally Shifted  \\
      \midrule
            Train & 388,406 & - \\
            Development & 27,036 & 9,569 \\
            Evaluation & 26,865 & 9,939 \\
      \bottomrule
    \end{tabular}
  \end{small}
  \end{center}
\end{table}

\paragraph{Collection Process} The Vehicle Motion Prediction data was collected by the perception system running onboard a number of self-driving vehicles equipped with LiDAR sensors, radars, and cameras. This perception system consists of a number of neural network--based detectors followed by an object tracker that fuses detections across sensor modalities and time. 
The provided HD map for each location has been constructed and validated by cartographers employed by Yandex SDG.
The provided dataset was sampled from a much larger dataset collected over a course of $8$ months. The sampling procedure was biased towards sampling scenes on which the motion prediction system currently used by the SDC fleet makes mistakes, as well as sampling more scenes from locations where the fleet drives less frequently.

\paragraph{Preprocessing and Cleaning} The collected dataset has been cleaned from scenes in which:
\begin{itemize}
    \item any kind of onboard system failure was detected, as the perception system output can potentially be unreliable in such scenes;
    \item the perception system has produced outputs that clearly violate physical constraints, such as actors having unrealistic acceleration or colliding with one other.
\end{itemize}

\paragraph{Format} This dataset is provided in protobuf format.

\paragraph{License}  We release this dataset under the CC BY NC SA 4.0 license.

\subsection{Task Setup} \label{subsection:task_setup}
\begin{figure}
  \centering
  \includegraphics[width=\linewidth]{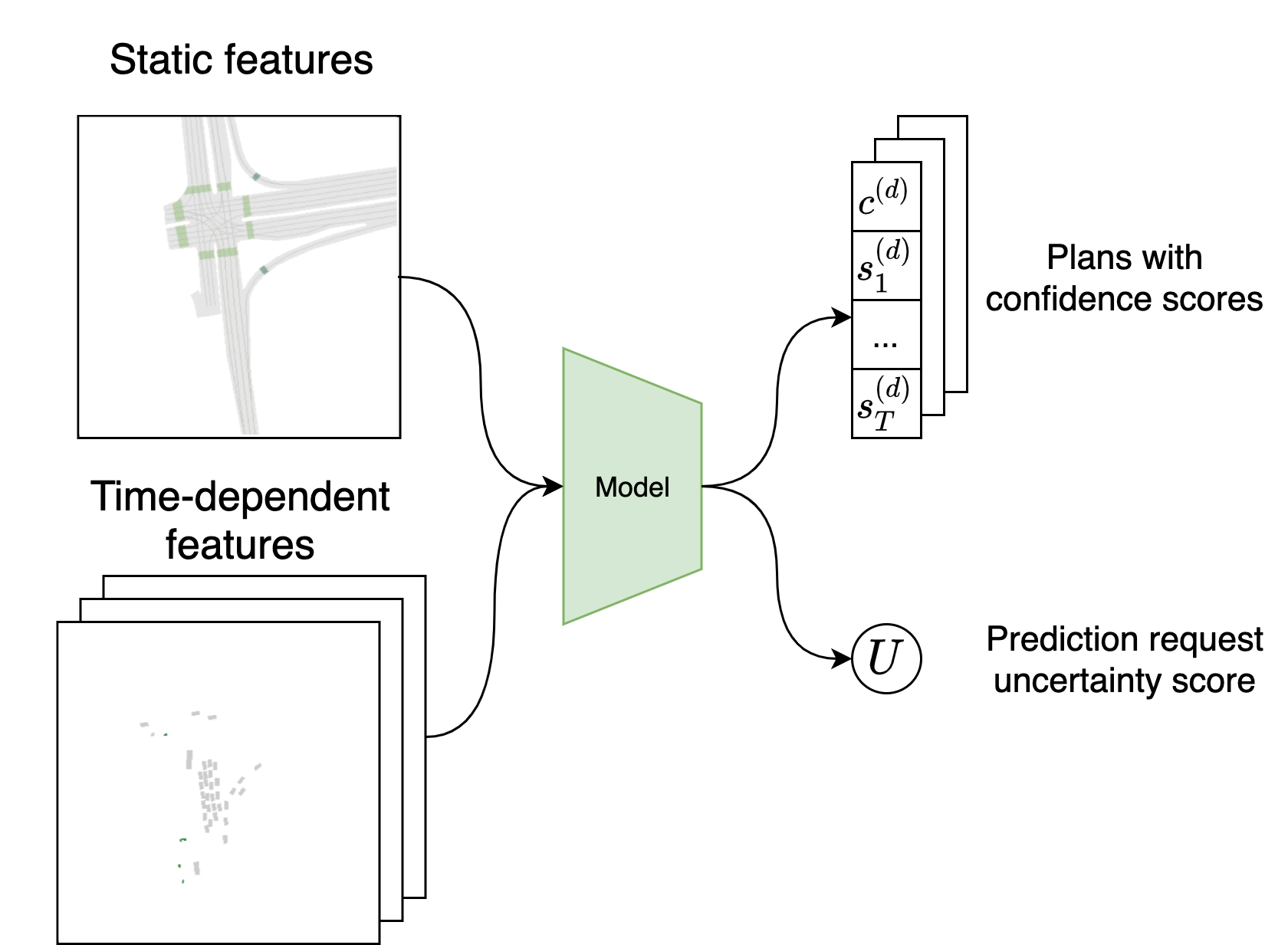}
  \caption{Diagram of the Vehicle Motion Prediction task. Models take as input a single scene context $\bm{x}$ composed of static (HD map) and time-dependent input features, and predict trajectories $\{\bm{y}^{(d)} \mid d \in 1, \dots, D\}$ with corresponding per-trajectory confidence scores $\{\bm{c}^{(d)} \mid d \in 1, \dots, D\}$, as well as a single per--prediction request uncertainty score $U$.}
  \label{fig:sdc_model}
\end{figure}

Vehicle Motion Prediction is a complex task and therefore must be described in detail. We provide a training dataset $\mathcal{D}_\textup{train}=\{(\bm{x}_i, \bm{y}_i)\}_{i = 1}^{N}$ of time-profiled ground truth trajectories (i.e., plans) $\bm{y}$ paired with high-dimensional observations (features) $\bm{x}$ of the corresponding scenes. Each $\bm{y} = (s_1, \dots, s_T)$ corresponds to the trajectory of a given vehicle observed through the SDG perception stack. Each state $s_t$ corresponds to the x- and y-displacement of the vehicle at timestep $t$, s.t. $\bm{y} \in \mathbb{R}^{T \times 2}$. We consider the performance of models on development and evaluation datasets $\mathcal{D}^{j}_\textup{dev}=\{(\bm{x}_i, \bm{y}_i)\}_{i = 1}^{M_j}$. and $\mathcal{D}^{j}_\textup{eval}=\{(\bm{x}_i, \bm{y}_i)\}_{i = 1}^{M_j}$. See Figure~\ref{fig:sdc_model} for a depiction of the task. 

\paragraph{Prediction Requests.} 
There are $N$ ($M_j$) prediction requests in the training dataset (evaluation datasets), with many requests for each scene corresponding to the many different vehicle trajectories observed. For example, in the canonical partition of the data, there are 388,406 scenes in the training dataset (Moscow, no precipitation), and 5,649,675 valid prediction requests.

Models can be trained to make use of ground truth trajectories that contain occlusions (i.e., prediction requests that are not valid) during training, such as through linear interpolation of missing steps. However, for the baseline methods considered in this work, both training and evaluation are done using only the fully observed ground truth trajectories.

Next, we describe the two levels of uncertainty quantification that we consider for each prediction request in the proposed task: per-trajectory and per--prediction request uncertainty scores.

\paragraph{Per-Trajectory Confidence Scores.} Like machine translation, motion prediction is an inherently multimodal task. A motion prediction model can produce a different number of sampled trajectories (plans) $D_i$ for each input $\bm{x}_i$; in other words, for two inputs $\bm{x}_i, \bm{x}_j$ with $i \neq j$, $D_i$ and $D_j$ can differ.
As a justification, consider that in a certain context, multiple trajectories may be desirable to capture multimodality (e.g., the vehicle of interest is at a T-junction), and in others a single or fewer trajectories would be sufficient (e.g., the vehicle is clearly proceeding straight). In our task, we expect a stochastic model to accompany its $D_i$ predicted trajectories on a given input $\bm{x}_i$ with scalar per-trajectory confidence scores $c_i^{(d)}, d \in \{1, \dots D_i\}$.  These provide an ordering of the plausibility of the various trajectories predicted for a given input. The scores must be non-negative and sum to $1$ (i.e., form a valid probability distribution). 

\paragraph{Per--Prediction Request Uncertainty Score.}
We also expect models to produce scalar uncertainty estimates corresponding to each prediction request input $\bm{x}_i$.
For example, on evaluation dataset $\mathcal{D}^{j}_\textup{eval}$, we have $M_j$ per--prediction request uncertainty scores $\{U_i \mid i \in 1, \dots, M_j\}$.
These correspond to the model's uncertainty in making any trajectory prediction for the agent of interest.
In a real-world deployment setting, a self-driving vehicle would associate a high per--prediction request uncertainty score with a scene context that is particularly unfamiliar or high-risk.

Next, we will describe standard motion prediction performance metrics, followed by confidence-aware metrics which reward models with well-calibrated uncertainty.

\subsection{Performance Metrics}\label{apn:sdc-metrics}

\paragraph{Standard Performance Metrics.}
We assess the performance of a motion prediction system using several standard metrics.

The average displacement error (ADE) measures the quality of a predicted trajectory $\bm{y}$ with respect to the ground truth trajectory $\bm{y}^*$ as
\begin{equation}
    \text{ADE}(\bm{y}) \coloneqq \frac{1}{T} \sum_{t = 1}^T \left\lVert s_t - s^*_t \right\rVert_2,
\label{eq:ade}
\end{equation}
where $\bm{y} = (s_1, \dots, s_T)$. Analogously, the final displacement error
\begin{equation}
    \text{FDE}(\bm{y}) \coloneqq \left\lVert s_T - s^*_T \right\rVert_2,
\label{eq:fde}
\end{equation}
measures the quality at the last timestep.

Stochastic models define a predictive distribution $q(\bm{y} \mid \bm{x}; \bm{\theta})$, and can therefore be evaluated over the $D$ trajectories sampled for a given input $\bm{x}$. For example, we can measure an aggregated ADE over $D$ samples with
\begin{equation}
    \text{aggADE}_D(q) \coloneqq \underset{\{\bm{y}\}_{d = 1}^{D} \sim q(\bm{y} \mid \bm{x})}{\oplus} \text{ADE}(\bm{y}^{d}),
\label{eq:agg_ade}
\end{equation}
where $\oplus$ is an aggregation operator, e.g., $\oplus = \min$ recovers the minimum ADE ($\text{minADE}_{D}$) commonly used in evaluation of stochastic motion prediction models \cite{filos2020can, phan2020covernet}. We consider minimum and mean aggregation of the average displacement error (minADE, avgADE), as well as of the final displacement error (minFDE, avgFDE).

\textbf{Per-Trajectory Confidence-Aware Metrics.}
A stochastic model used in practice for motion prediction must ultimately \emph{decide} on a particular predicted trajectory for a given prediction request. We may make this decision by selecting for evaluation the predicted trajectory with the highest per-trajectory confidence score. In other words, given per-trajectory confidence scores $\{c^{(d)} \mid d \in 1, \dots, D\}$ we select the top trajectory $y^{(d^{*})}, d^* = \underset{d}{\arg \max}\ c^{(d)}$, and measure the decision quality using \emph{top1} ADE and FDE metrics, e.g.,
\begin{equation}
    \text{top1ADE}_D(q) \coloneqq \text{ADE}(\bm{y}^{(d^*)}).
\label{eq:top1_ade}
\end{equation}

We may also wish to assess the quality of the relative weighting of the $D$ trajectories with their corresponding per-trajectory confidence scores $c^{(d)}$. For this the following weighted metric can be considered:
\begin{equation}
    \text{weightedADE}_D(q) \coloneqq \sum_{d \in D} c^{(d)} \cdot \text{ADE}(\bm{y}^{(d)}).
\label{eq:weighted_ade}
\end{equation}
The top1FDE and weightedFDE metrics follow analogously to the above. Unfortunately, these metrics, while highly intuitive, have a conceptual limitation. 
Consider the following loss:
\begin{empheq}{align}
    &\mathcal{L}\left({\tt p}(\bm{y}|\bm{x}), \{\hat c_i^{(1:D)},\bm{\hat y}^{(1:D)}\}\right) = \mathbb{E}_{{\tt p}(\bm{y}|\bm{x})}\left[\sum_{d=1}^D c^d\text{ADE}(\bm{\hat y}^d,\bm{y}) \right], \  \{\hat c_i^{(1:D)},\bm{\hat y}^{(1:D)}\} = \bm{f}(\bm{x};\bm{\theta})
\end{empheq}
which is the expected weightedADE given a set of trajectories and weights from a model. If we wish to minimize this loss with respect to the predicted trajectories and weights, then:

\begin{empheq}{align}
    \begin{split}
     \mathcal{L}_{\text{min}}=&\  \min_{\{\hat c_i^{(1:D)},\bm{\hat y}^{(1:D)}\}}\left\{\mathbb{E}_{{\tt p}(\bm{y}|\bm{x})}\left[\sum_{d=1}^D c^d\text{ADE}(\bm{\hat y}^d,\bm{y}) \right]\right\} \\
    =&\ \min_{\{\hat c_i^{(1:D)}\}} \left\{\sum_{d=1}^D \hat c^d\left(\min_{\{\hat y^{(d)}\}}\left\{\mathbb{E}_{{\tt p}(\bm{y}|\bm{x})}\left[\text{ADE}(\bm{\hat y}^d,\bm{y}) \right]\right\}\right)\right\} \\
    =&\ \min_{\{\hat c_i^{(1:D)}\}} \left\{\mathbb{E}_{{\tt p}(\bm{y}|\bm{x})}\left[\text{ADE}(\bm{\hat y}^*,\bm{y}) \right] \sum_{d=1}^D \hat c^d\right\}  \\
    =&\ \mathbb{E}_{{\tt p}(\bm{y}|\bm{x})}\left[\text{ADE}(\bm{\hat y}^*,\bm{y}) \right]
    \end{split}
\label{eq:optimalWADE}
\end{empheq}
where $ \bm{\hat y}^{*}$ is the \emph{weighted geometric median}
\begin{empheq}{align}
     \bm{\hat y}^{*} =& \ \arg_{\{ \bm{\hat y}\}}\min\left\{\mathbb{E}_{{\tt p}(\bm{y}|\bm{x})}\left[\text{ADE}(\bm{\hat y},\bm{y}) \right]\right\}
\end{empheq}
Thus, the optimal model would suffer from \emph{mode-collapse} and always yields the weighted geometric median of the modes of the true distribution of trajectories. To put this concretely, at a T-junction, where trajectories can go either left or right, the optimal model will yield a trajectory going straight, which is clearly a fundamentally undesirable behaviour. Mathematically, the problem lies in the additive nature of the metric -- each mode can be optimized independently of the others. This can be avoided by instead considering a likelihood based metric, such as the following one:
\begin{equation}
    \text{cNLL}(\mathcal{D}) \coloneqq \frac{1}{N}\sum_{n=1}^N\left\{-\ln\left[\sum_{d=1}^D c^{(d)} \prod_{t=1}^T \mathcal{N}(\bm{y}_{t,i}^{*};\ \bm{s}_t^{(d)}(\bm{x}_i; \bm{\theta}), \bm{\Sigma}=\bm{1})   \right]\right\} - T\ln2\pi
\label{eq:corrected_NLL}
\end{equation}
Under the following metric, which assumes that each mode is modelled using a Normal distribution of fixed variance, an optimal model would place a Normal over each mode and weight them appropriately. This can be clearly demonstrated using the following numerical example: 
\begin{empheq}{align}
    &y \sim {\tt p}(y) = 0.5\cdot \mathcal{N}(x, 10, 1) + 0.5\cdot \mathcal{N}(x, -10, 1) \\
    \begin{split}
    &\mathbb{E}_{\tt p}(y) [\text{wADE}(y, \bm{s}^{(1:2)}=[10,-10], \bm{c} = [0.5,0.5])] = 201.5 \\
    &\mathbb{E}_{\tt p}(y) [\text{wADE}(y, \bm{s}^{(1:2)}=[0,0], \bm{c} = [0.5,0.5])] = 101.50
    \end{split} \\
    &\mathbb{E}_{\tt p}(y) [\text{cNLL}(y, \bm{s}^{(1:2)}=[10,-10], \bm{c} = [0.5,0.5])] = 1.09 \\
    &\mathbb{E}_{\tt p}(y) [\text{cNLL}(y, \bm{s}^{(1:2)}=[0,0], \bm{c} = [0.5,0.5])] = 50.75 
\end{empheq}
Where we have a bimodal Gaussian mixture distribution with modes at -10, 10. We assume we have a model which predicts the means of two trajectories with equal weight. We have two situations: either the model yields two distinct modes at -10, 10 or a collapsed mode at 0 (the median). We can see that predicting the median will yield a lower weightedADE and correctly predicting two distinct modes will yield the lower cNLL. It is important to highlight that this argument holds \emph{in expectation} and is relevant to situations which contain inherent ambiguity and multi-modality. Note that the offset $T\ln2\pi$ is used to make assure that the minimal value of this metric is 0, so that it can be used for error-retention and F1-retention plots.

\paragraph{Per-Prediction Request Confidence-Aware Metrics.}
In addition to making a decision amongst many possible trajectories in a particular situation, a motion planning agent should know when, in general, any trajectories it predicts will be inaccurate (e.g., due to unfamiliarity of the setting, or inherent ambiguity in the path of the vehicle for which a prediction is requested). We evaluate the quality of uncertainty quantification jointly with robustness to distributional shift using the retention-based metrics described in \cref{sec:metrics}, with the per--prediction request uncertainty scores determining retention order. Note that each retention curve is plotted with respect to a particular error metric above (e.g., we consider AUC for retention with respect to the cNLL metric introduced above, written as R-AUC). Additionally, we also assess whether the per--prediction uncertainty scores can be used to discriminate between in-domain and shifted scenes. In this case, quality is assessed via area under a ROC curve (ROC-AUC).

\subsection{Experimental Setup}\label{sec:sdc-setup}

\textbf{Robust Imitative Planning.}  In detail, we use the following approach for trajectory and confidence score generation.
\begin{enumerate}[label=\textbf{\arabic*})]
    \item \textbf{Trajectory Generation.} Given a scene input $\bm{x}$, $K$ ensemble members generate $G$ trajectories.\footnote{In practice, each ensemble member generates the same number of trajectories $Q$, s.t. $G = K \cdot Q$.}
    \item \textbf{Trajectory Scoring.} We score each of the $G$ trajectories by computing a log probability under each of the $K$ trained likelihood models. 
    \item \textbf{Per-Trajectory Confidence Scores.} We aggregate the $G \cdot K$ resulting log probabilities to $G$ scores using a per-trajectory aggregation operator $\oplus_{\text{trajectory}}$.\footnote{For example, applying a $\texttt{min}$ aggregation is informed by robust control literature \cite{wald1939contributions} in which we aim to optimize for the worst-case scenario, as measured by the log-likelihood of the ``most pessimistic'' model for a given trajectory.} 
    By aggregating over the log-likelihood estimates sampled from the model posterior (i.e., contributed by each ensemble member), we obtain a robust score for each of the $G$ trajectories \cite{filos2020can}.
    \item \textbf{Trajectory Selection.} Among the $G$ trajectories, the RIP ensemble produces the top $D$ trajectories as determined by their corresponding $G$ per-trajectory confidence scores, where $D$ is a hyperparameter.
    \item \textbf{Per--Prediction Request Uncertainty Score.} We aggregate the $D$ top per-trajectory confidence scores to a single uncertainty score $U$ using the aggregator $\oplus_{\text{pred-req}}$.\footnote{In practice, this is done by applying the aggregation (e.g., $\oplus_{\text{pred-req}} = \texttt{mean}$) to the confidences $c^{(d)}$, and then \emph{negating} to obtain the uncertainty score $U$.} This value conveys the ensemble's estimated uncertainty for a given scene context and a particular prediction request.
   \item \textbf{Confidence Reporting.} We obtain scores $c^{(d)}$ by applying a \texttt{softmax} to the $D$ top per-trajectory  confidence scores. We report these $c^{(d)}$ and $U$ (computed in step \textbf{5}) as our final per-trajectory  confidence scores and per--prediction request uncertainty score, respectively.
\end{enumerate}

To summarize, our implementation of RIP for motion prediction produces $D$ trajectories and corresponding normalized per-trajectory scores $\{c^{(d)} \mid d \in 1, \dots, D\}$, as well as an aggregated uncertainty score $U$ for the overall prediction request. 

\textbf{Backbone Likelihood Model.} We consider two different model classes as ensemble members: a simple behavioral cloning agent with a Gated Recurrent Unit decoder (BC) \cite{cho2014learning, codevilla2018end} and a Deep Imitative Model (DIM) \cite{rhinehart2018deep} with an autoregressive flow decoder \cite{rezende2016variational}, following \cite{filos2020can}.
In both cases, we model the likelihood of a trajectory $\bm{y}$ in context $\bm{x}$ to come from an expert (i.e., from the distribution of ground truth trajectories), with learnable parameters $\bm{\theta}$, as
\begin{equation}
\begin{split}
q(\bm{y} \mid \bm{x}; \bm{\theta}) & = \prod_{t = 1}^T p(s_t \mid \bm{y}_{<t}, \bm{x}; \bm{\theta})\ = \prod_{t = 1}^{T} \mathcal{N}(s_t; \mu(\bm{y}_{<t}, \bm{x}; \bm{\theta}), \Sigma(\bm{y}_{<t}, \bm{x}; \bm{\theta})),
\end{split}
\label{eq:sdc_likelihood_model}
\end{equation}
where $\mu(\cdot; \bm{\theta})$ and $\Sigma(\cdot; \bm{\theta})$ are two heads of a recurrent neural network with shared torso. 
Hence we assume that the conditional densities are normally distributed, and learn those parameters through maximum likelihood estimation.
Notably, for the BC model, we found that conditioning on samples $\hat{\bm{y}}_{<t}$ instead of ground truth values $\bm{y}_{<t}$ (where usage of ground truth is often referred to as teacher forcing in RNN literature) significantly improved performance across all datasets and metrics.

\textbf{Uncertainty Estimation Methods.} The above ensembling is done using multiple stochastic models trained with different random seeds, as introduced in Deep Ensembles \cite{deepensemble2017}.
For each ensemble member, we generate $Q$ trajectories.
We can also use a Monte Carlo Dropout \cite{Gal2016Dropout} approach for each ensemble member, in which we sample new dropout masks \emph{at test time} during each of the $Q$ forward passes (and corresponding trajectory generations). 
Following \cite{smith2018understanding} we refer to the combination of this uncertainty estimation method with ensembling as Dropout Ensembles. 
Previous work has investigated the benefits of Deep Ensembles from a loss landscape perspective \cite{fort2020deep}, and found that Deep Ensembles tend to explore diverse modes in function space, whereas approximate variational methods such as Monte Carlo Dropout explore around a particular mode. 
Dropout Ensembles are hence motivated as ensembles of variational methods which aim to consider a diverse set of modes, with local exploration around each mode.





\textbf{Setup.} We report performance of RIP across the two backbone models -- Behavioral Cloning (BC) \cite{codevilla2018end} and Deep Imitative Model (DIM) \cite{rhinehart2018deep} -- as well as the two uncertainty estimation methods -- Deep Ensembles \cite{deepensemble2017} and Dropout Ensembles \cite{Gal2016Dropout, smith2018understanding}.
We evaluate RIP on development (\texttt{dev}) and evaluation (\texttt{eval}) datasets in in-distribution (In), distributionally shifted (Shifted), and combined in-distribution and shifted (Full) settings.
With both backbone model classes we vary the number of ensemble members $K \in \{1, 3, 5\}$, train with learning rate 1e-4, use a cosine annealing LR schedule with 1 epoch warmup, and use gradient clipping at 1. 
We sample $Q = 10$ trajectories from each of the ensemble members.
We consider two types of aggregation: ``Lower Quartile'' in which we compute the mean minus the standard deviation $\mu - \sigma$ of the input scores, and ``Model Averaging'' (MA) in which we compute the mean $\mu$ of the input scores.
LQ reflects the intuition to assign a high score to a trajectory when the ensemble members assign it a high score on average, and tend to be certain (have a low standard deviation) in their scoring; MA reflects only the prior intuition.
This aggregation strategy (LQ or MA) is used as both the per-trajectory aggregation operator $\oplus_{\text{trajectory}}$ and the per--prediction request aggregation operator $\oplus_{\text{pred-req}}$ (where the latter is followed by negation to obtain an uncertainty, as opposed to a confidence).
We fix the RIP ensemble at all $K$ to produce the top $D = 5$ trajectories as ranked by their per-trajectory confidence score.

\subsection{Additional Results}\label{apn:sdc-results}

Below, we report predictive performance using standard-metrics, robustness and uncertainty quantification metrics, and retention plots across the RIP variants.

\begin{table}[p]
    \centering
    \caption{\emph{Predictive performance} of RIP, across model backbones (behavioral cloning (BC) \cite{codevilla2018end} and Deep Imitative Model (DIM) \cite{rhinehart2018deep}) and uncertainty estimation methods (Deep Ensembles \cite{deepensemble2017} and Dropout Ensembles \cite{smith2018understanding}). Each section contains losses computed over the in-distribution (In), distributionally shifted (Shifted), and combined (Full) development and evaluation datasets. Altogether, we vary the backbone model, uncertainty estimation method, aggregation strategy (applied for both the per-trajectory aggregation operator $\oplus_{\text{trajectory}}$ and the per--prediction request aggregation operator $\oplus_{\text{pred-req}}$), and the number of ensemble members $K$.
    See \cref{sec:sdc-setup} for setup details.}
    \label{table-apn:sdc_performance}
\vspace{0.1in}
\resizebox{1\textwidth}{!}{%
    \begin{tabular}{@{}lll|ccc|ccc|ccc|ccc|ccc}
        \toprule
          \multirow{2}*{Dataset} & \multirow{2}*{Method} & \multirow{2}*{Model} & \multicolumn{3}{c|}{\text{minADE} $\downarrow$} & \multicolumn{3}{c|}{\text{weightedADE} $\downarrow$} & \multicolumn{3}{c|}{ \text{minFDE} $\downarrow$} &  \multicolumn{3}{c}{\text{weightedFDE} $\downarrow$} &  \multicolumn{3}{c}{\text{cNLL} $\downarrow$}
          \\
          & & & In & Shifted & Full & In & Shifted & Full & In & Shifted & Full & In & Shifted & Full & In & Shifted & Full \\
        \midrule
        \multirow{24}*{\texttt{Dev}} & \multirow{12}*{\shortstack[l]{Deep\\ Ensemble}}
     & \text{BC, LQ, K=1} & 0.818 & 0.960 & 0.835 & 1.088 & 1.245 & 1.107 & 1.718 & 2.113 & 1.765 & 2.368 & 2.777 & 2.417 & 59.64 & 98.54 & 64.29 \\
     & & \text{BC, LQ, K=3} & 0.780 & 0.909 & 0.795 & 1.040 & 1.170 & 1.056 & 1.638 & 2.018 & 1.683 & 2.254 & 2.609 & 2.297 & 54.78 & 87.81 & 58.73 \\
     & & \text{BC, LQ, K=5} & 0.766 & 0.888 & 0.780 & 1.017 & 1.138 & 1.031 & 1.618 & 1.980 & 1.661 & 2.214 & 2.552 & 2.254 & 56.45 & 90.25 & 60.49 \\
     & & \text{BC, MA, K=1} & 0.818 & 0.960 & 0.835 & 1.088 & 1.245 & 1.107 & 1.718 & 2.113 & 1.765 & 2.368 & 2.777 & 2.417 & 59.64 & 98.54 & 64.29 \\
     & & \text{BC, MA, K=3} & 0.780 & 0.908 & 0.795 & 1.034 & 1.166 & 1.050 & 1.641 & 2.018 & 1.686 & 2.249 & 2.611 & 2.292 & 55.00 & 88.45 & 59.00 \\
     & & \text{BC, MA, K=5} & 0.765 & 0.887 & 0.779 & 1.012 & 1.133 & 1.026 & 1.617 & 1.976 & 1.660 & 2.210 & 2.551 & 2.251 & 56.86 & 91.54 & 61.01 \\
     & & \text{DIM, LQ, K=1} & 0.750 & 0.818 & 0.758 & 1.523 & 1.583 & 1.530 & 1.497 & 1.720 & 1.524 & 3.472 & 3.639 & 3.492 & 50.66 & 73.00 & 53.34 \\
     & & \text{DIM, LQ, K=3} & 0.717 & 0.787 & 0.725 & 1.407 & 1.470 & 1.415 & 1.467 & 1.687 & 1.493 & 3.219 & 3.397 & 3.240 & 48.88 & \bf{70.93} & 51.52 \\
     & & \text{DIM, LQ, K=5} & 0.720 & 0.787 & 0.728 & 1.399 & 1.470 & 1.407 & 1.487 & 1.704 & 1.513 & 3.202 & 3.397 & 3.225 & 51.12 & 72.87 & 53.72 \\
     & & \text{DIM, MA, K=1} & 0.750 & 0.818 & 0.758 & 1.523 & 1.583 & 1.530 & 1.497 & 1.720 & 1.524 & 3.472 & 3.639 & 3.492 & 50.66 & 73.00 & 53.34 \\
     & & \text{DIM, MA, K=3} & 0.717 & \bf{0.785} & 0.725 & 1.410 & 1.475 & 1.418 & 1.466 & \bf{1.685} & 1.492 & 3.226 & 3.409 & 3.248 & \bf{48.74} & 71.30 & \bf{51.44} \\
     & & \text{DIM, MA, K=5} & 0.719 & 0.786 & 0.727 & 1.399 & 1.469 & 1.408 & 1.482 & 1.698 & 1.508 & 3.202 & 3.393 & 3.225 & 50.85 & 72.45 & 53.43 \\
    \arrayrulecolor{black!25}\cmidrule(lr{1em}){2-18}
    & \multirow{12}*{\shortstack[l]{Dropout\\ Ensemble}}
    & \text{BC, LQ, K=1} & 0.803 & 0.908 & 0.815 & 1.116 & 1.236 & 1.130 & 1.649 & 1.952 & 1.685 & 2.409 & 2.718 & 2.446 & 55.98 & 82.49 & 59.15 \\
    & & \text{BC, LQ, K=3} & 0.741 & 0.853 & 0.754 & 1.013 & 1.132 & 1.028 & 1.542 & 1.873 & 1.581 & 2.209 & 2.545 & 2.249 & 53.01 & 83.93 & 56.71 \\
    & & \text{BC, LQ, K=5} & 0.759 & 0.878 & 0.773 & \bf{1.008} & 1.127 & \bf{1.023} & 1.605 & 1.960 & 1.648 & \bf{2.204} & \bf{2.538} & \bf{2.244} & 55.58 & 88.78 & 59.55 \\
    & & \text{BC, MA, K=1} & 0.803 & 0.908 & 0.815 & 1.116 & 1.236 & 1.130 & 1.649 & 1.952 & 1.685 & 2.409 & 2.718 & 2.446 & 55.98 & 82.49 & 59.15 \\
    & & \text{BC, MA, K=3} & 0.739 & 0.850 & 0.752 & 1.020 & 1.135 & 1.033 & 1.534 & 1.864 & 1.574 & 2.223 & 2.553 & 2.263 & 53.09 & 83.81 & 56.76 \\
    & & \text{BC, MA, K=5} & 0.757 & 0.877 & 0.771 & 1.010 & \bf{1.126} & 1.024 & 1.597 & 1.952 & 1.640 & 2.209 & 2.539 & 2.248 & 55.82 & 89.57 & 59.86 \\
    & & \text{DIM, LQ, K=1} & 0.750 & 0.831 & 0.759 & 1.498 & 1.587 & 1.509 & 1.510 & 1.757 & 1.539 & 3.432 & 3.662 & 3.459 & 52.57 & 76.54 & 55.44 \\
    & & \text{DIM, LQ, K=3} & \bf{0.716} & 0.786 & 0.725 & 1.412 & 1.473 & 1.419 & 1.466 & 1.687 & 1.493 & 3.234 & 3.408 & 3.254 & 49.69 & 72.58 & 52.43 \\
    & & \text{DIM, LQ, K=5} & 0.723 & 0.793 & 0.731 & 1.409 & 1.475 & 1.417 & 1.494 & 1.717 & 1.521 & 3.224 & 3.408 & 3.246 & 51.25 & 73.47 & 53.91 \\
    & & \text{DIM, MA, K=1} & 0.750 & 0.831 & 0.759 & 1.498 & 1.587 & 1.509 & 1.510 & 1.757 & 1.539 & 3.432 & 3.662 & 3.459 & 52.57 & 76.54 & 55.44 \\
    & & \text{DIM, MA, K=3} & \bf{0.716} & 0.786 & \bf{0.724} & 1.414 & 1.479 & 1.422 & \bf{1.465} & \bf{1.685} & \bf{1.491} & 3.238 & 3.420 & 3.260 & 49.38 & 71.86 & 52.07 \\
    & & \text{DIM, MA, K=5} & 0.721 & 0.793 & 0.729 & 1.409 & 1.474 & 1.417 & 1.489 & 1.717 & 1.516 & 3.224 & 3.405 & 3.246 & 50.99 & 73.64 & 53.70 \\
 \arrayrulecolor{black}\midrule
  \multirow{24}*{\texttt{Eval}} & \multirow{12}*{\shortstack[l]{Deep\\ Ensemble}}
    & \text{BC, LQ, K=1} & 0.829 & 1.084 & 0.880 & 1.104 & 1.407 & 1.164 & 1.733 & 2.420 & 1.870 & 2.394 & 3.197 & 2.555 & 60.20 & 98.82 & 67.93 \\
    & & \text{BC, LQ, K=3} & 0.792 & 1.026 & 0.839 & 1.056 & 1.326 & 1.110 & 1.658 & 2.297 & 1.786 & 2.284 & 3.005 & 2.429 & 55.97 & 90.54 & 62.89 \\
    & & \text{BC, LQ, K=5} & 0.777 & 1.015 & 0.825 & 1.032 & 1.303 & 1.086 & 1.636 & 2.283 & 1.765 & 2.242 & 2.964 & 2.386 & 57.26 & 93.92 & 64.60 \\
    & & \text{BC, MA, K=1} & 0.829 & 1.084 & 0.880 & 1.104 & 1.407 & 1.164 & 1.733 & 2.420 & 1.870 & 2.394 & 3.197 & 2.555 & 60.20 & 98.82 & 67.93 \\
    & & \text{BC, MA, K=3} & 0.792 & 1.025 & 0.838 & 1.050 & 1.319 & 1.104 & 1.661 & 2.294 & 1.788 & 2.278 & 2.997 & 2.422 & 55.94 & 90.53 & 62.87 \\
    & & \text{BC, MA, K=5} & 0.777 & 1.014 & 0.824 & 1.028 & 1.299 & 1.082 & 1.636 & 2.278 & 1.765 & 2.238 & 2.957 & 2.382 & 57.75 & 95.00 & 65.20 \\
    & & \text{DIM, LQ, K=1} & 0.759 & 0.942 & 0.796 & 1.551 & 1.883 & 1.618 & 1.511 & 1.983 & 1.605 & 3.536 & 4.376 & 3.704 & 50.50 & 76.00 & 55.60 \\
    & & \text{DIM, LQ, K=3} & 0.726 & 0.914 & 0.764 & 1.433 & 1.756 & 1.498 & 1.481 & 1.972 & 1.579 & 3.277 & 4.094 & 3.440 & 49.45 & 76.66 & 54.89 \\
    & & \text{DIM, LQ, K=5} & 0.729 & 0.921 & 0.768 & 1.422 & 1.757 & 1.489 & 1.498 & 2.007 & 1.600 & 3.253 & 4.098 & 3.422 & 51.61 & 79.71 & 57.24 \\
    & & \text{DIM, MA, K=1} & 0.759 & 0.942 & 0.796 & 1.551 & 1.883 & 1.618 & 1.511 & 1.983 & 1.605 & 3.536 & 4.376 & 3.704 & 50.50 & 76.00 & 55.60 \\
    & & \text{DIM, MA, K=3} & 0.726 & 0.912 & 0.763 & 1.437 & 1.759 & 1.502 & 1.478 & 1.967 & 1.576 & 3.286 & 4.101 & 3.449 & 49.09 & 76.07 & 54.49 \\
    & & \text{DIM, MA, K=5} & 0.728 & 0.918 & 0.766 & 1.424 & 1.754 & 1.490 & 1.493 & 2.000 & 1.595 & 3.256 & 4.093 & 3.424 & 51.19 & 78.85 & 56.73 \\
 \arrayrulecolor{black!25}\cmidrule(lr{1em}){2-18}
  & \multirow{12}*{\shortstack[l]{Dropout\\ Ensemble}}
    & \text{BC, LQ, K=1} & 0.812 & 1.038 & 0.857 & 1.128 & 1.410 & 1.184 & 1.664 & 2.267 & 1.784 & 2.430 & 3.170 & 2.578 & 56.57 & 86.28 & 62.52 \\
    & & \text{BC, LQ, K=3} & 0.751 & 0.972 & 0.795 & 1.029 & \bf{1.297} & 1.082 & 1.558 & 2.154 & 1.677 & 2.238 & \bf{2.948} & 2.380 & 53.94 & 86.68 & 60.49 \\
    & & \text{BC, LQ, K=5} & 0.770 & 1.008 & 0.817 & \bf{1.024} & \bf{1.297} & \bf{1.079} & 1.623 & 2.268 & 1.752 & \bf{2.233} & 2.957 & \bf{2.378} & 56.49 & 92.77 & 63.75 \\
    & & \text{BC, MA, K=1} & 0.812 & 1.038 & 0.857 & 1.128 & 1.410 & 1.184 & 1.664 & 2.267 & 1.784 & 2.430 & 3.170 & 2.578 & 56.57 & 86.28 & 62.52 \\
    & & \text{BC, MA, K=3} & 0.749 & 0.970 & 0.794 & 1.036 & 1.305 & 1.090 & 1.551 & 2.147 & 1.670 & 2.253 & 2.963 & 2.395 & 54.07 & 86.94 & 60.65 \\
    & & \text{BC, MA, K=5} & 0.768 & 1.004 & 0.815 & 1.027 & 1.299 & 1.081 & 1.615 & 2.253 & 1.743 & 2.239 & 2.958 & 2.383 & 56.90 & 93.27 & 64.18 \\
    & & \text{DIM, LQ, K=1} & 0.739 & 0.924 & 0.776 & 1.478 & 1.815 & 1.546 & 1.474 & \bf{1.949} & 1.569 & 3.380 & 4.239 & 3.552 & 49.90 & 75.31 & 54.98 \\
    & & \text{DIM, LQ, K=3} & 0.722 & 0.910 & 0.760 & 1.431 & 1.763 & 1.497 & 1.470 & 1.967 & 1.569 & 3.266 & 4.112 & 3.435 & 49.30 & 75.24 & 54.49 \\
    & & \text{DIM, LQ, K=5} & 0.729 & 0.929 & 0.769 & 1.430 & 1.769 & 1.497 & 1.497 & 2.027 & 1.603 & 3.268 & 4.126 & 3.440 & 50.77 & 80.02 & 56.63 \\
    & & \text{DIM, MA, K=1} & 0.739 & 0.924 & 0.776 & 1.478 & 1.815 & 1.546 & 1.474 & \bf{1.949} & 1.569 & 3.380 & 4.239 & 3.552 & 49.90 & 75.31 & 54.98 \\
    & & \text{DIM, MA, K=3} & \bf{0.720} & \bf{0.907} & \bf{0.758} & 1.432 & 1.760 & 1.497 & \bf{1.465} & 1.960 & \bf{1.564} & 3.267 & 4.107 & 3.435 & \bf{48.74} & \bf{74.70} & \bf{53.93} \\
    & & \text{DIM, MA, K=5} & 0.728 & 0.925 & 0.767 & 1.431 & 1.766 & 1.498 & 1.494 & 2.017 & 1.599 & 3.269 & 4.120 & 3.439 & 50.51 & 79.30 & 56.28 \\
        \bottomrule
    \end{tabular}
}
\end{table}

\begin{table}[p]
    \centering
    \caption{\emph{Uncertainty and robustness performance} of RIP across the two backbone models (BC and DIM) and uncertainty estimation methods (Deep Ensemble and Dropout Ensemble). The error metric for computing the area under the rejection curve (R-AUC) and area under the F1 curve (F1-AUC) is \textbf{cNLL}. We use a threshold of 25 for the F1 metrics, which approximately corresponds to a 1 meter deviation on all trajectories. See \cref{sec:sdc-setup} for setup details.}
    \label{table-apn:sdc_uncertainty}
\vspace{0.1in}
\resizebox{1\textwidth}{!}{
    \begin{tabular}{lll|ccc|ccc|ccc|ccc|c}
        \toprule
         \multirow{2}*{Dataset} & \multirow{2}*{Method} & \multirow{2}*{Model} & \multicolumn{3}{c|}{\text{R-AUC} $\downarrow$} & \multicolumn{3}{c|}{\text{F1-AUC} (\%) $\uparrow$} & \multicolumn{3}{c|}{\text{F1@}$95$\% $\uparrow$} & \multirow{2}*{ROC-AUC (\%) $\uparrow$} \\
         & & & In & Shifted & Full & In & Shifted & Full & In & Shifted & Full &  \\
        \midrule
    \multirow{24}*{\texttt{Dev}} & \multirow{12}*{\shortstack[l]{Deep\\ Ensemble}}
        & \text{BC, LQ, K=1} & 11.06 & 13.91 & 11.22 & 64.9 & 66.7 & 65.1 & 89.1 & 90.2 & 89.3 & 51.0 \\ 
        & & \text{BC, LQ, K=3} & 11.26 & 11.69 & 11.18 & 63.4 & 66.0 & 63.8 & 88.5 & 90.3 & 88.8 & 46.7 \\
        & & \text{BC, LQ, K=5} & 9.68 & 10.38 & 9.62 & 64.3 & 66.4 & 64.6 & 89.7 & 91.0 & 90.0 & 47.3 \\
        & & \text{BC, MA, K=1} & 11.06 & 13.91 & 11.22 & 64.9 & 66.7 & 65.1 & 89.1 & 90.2 & 89.3 & 51.0 \\
        & & \text{BC, MA, K=3} & 9.31 & 10.73 & 9.31 & 64.8 & 66.5 & 65.0 & 90.3 & 91.3 & 90.6 & 48.6 \\
        & & \text{BC, MA, K=5} & 9.07 & 10.47 & 9.08 & 64.9 & 66.5 & 65.2 & 90.4 & 91.3 & 90.6 & 49.2 \\
        & & \text{DIM, LQ, K=1} & 12.54 & 15.28 & 12.86 & 63.6 & 64.8 & 63.8 & 87.2 & 88.8 & 87.4 & \bf{51.8} \\
        & & \text{DIM, LQ, K=3} & 12.30 & 14.51 & 12.57 & 63.7 & 64.9 & 63.8 & 89.3 & 89.9 & 89.3 & 51.4 \\
        & & \text{DIM, LQ, K=5} & 12.87 & 15.01 & 13.14 & 63.5 & 64.8 & 63.7 & 89.7 & 90.2 & 89.7 & 51.4 \\
        & & \text{DIM, MA, K=1} & 12.57 & 15.10 & 12.86 & 63.7 & 64.9 & 63.8 & 87.2 & 88.8 & 87.4 & \bf{51.8} \\
        & & \text{DIM, MA, K=3} & 12.38 & 14.46 & 12.64 & 63.7 & 64.9 & 63.8 & 89.2 & 89.9 & 89.3 & 51.4 \\
        & & \text{DIM, MA, K=5} & 12.97 & 15.10 & 13.24 & 63.5 & 64.8 & 63.7 & 89.6 & 90.2 & 89.7 & 51.4 \\
    \arrayrulecolor{black!25}\cmidrule(lr{1em}){2-13}
    & \multirow{12}*{\shortstack[l]{Dropout\\ Ensemble}}
        & \text{BC, LQ, K=1} & 8.87 & 10.00 & 8.87 & \bf{65.3} & \bf{67.1} & \bf{65.6} & 89.7 & 90.4 & 89.9 & 51.2 \\
        & & \text{BC, LQ, K=3} & \bf{8.11} & \bf{9.53} & \bf{8.14} & 64.9 & 66.5 & 65.1 & 90.6 & 91.3 & \bf{90.8} & 50.9 \\
        & & \text{BC, LQ, K=5} & 8.28 & 9.60 & 8.28 & 65.0 & 66.6 & 65.2 & 90.5 & 91.3 & 90.7 & 50.7 \\
        & & \text{BC, MA, K=1} & 8.87 & 9.99 & 8.87 & \bf{65.3} & \bf{67.1} & \bf{65.6} & 89.7 & 90.4 & 89.9 & 51.2 \\
        & & \text{BC, MA, K=3} & 8.53 & 9.79 & 8.54 & 64.9 & 66.5 & 65.1 & \bf{90.7} & \bf{91.4} & \bf{90.8} & 50.3 \\
        & & \text{BC, MA, K=5} & 8.89 & 10.23 & 8.90 & 64.9 & 66.5 & 65.2 & 90.5 & \bf{91.4} & 90.7 & 50.2 \\
        & & \text{DIM, LQ, K=1} & 12.57 & 16.41 & 13.03 & 63.8 & 64.7 & 63.9 & 87.6 & 89.1 & 87.8 & 51.5 \\
        & & \text{DIM, LQ, K=3} & 12.37 & 14.91 & 12.69 & 63.7 & 64.8 & 63.8 & 89.2 & 90.0 & 89.3 & 51.3 \\
        & & \text{DIM, LQ, K=5} & 12.94 & 15.18 & 13.22 & 63.6 & 64.8 & 63.7 & 89.6 & 90.2 & 89.7 & 51.4 \\
        & & \text{DIM, MA, K=1} & 12.61 & 16.30 & 13.06 & 63.8 & 64.8 & 63.9 & 87.6 & 89.1 & 87.7 & 51.6 \\
        & & \text{DIM, MA, K=3} & 12.49 & 14.80 & 12.79 & 63.6 & 64.8 & 63.8 & 89.2 & 90.0 & 89.3 & 51.4 \\
        & & \text{DIM, MA, K=5} & 13.05 & 15.20 & 13.33 & 63.5 & 64.8 & 63.7 & 89.5 & 90.2 & 89.6 & 51.4 \\
    \arrayrulecolor{black}\midrule
    \multirow{24}*{\texttt{Eval}} & \multirow{12}*{\shortstack[l]{Deep\\ Ensemble}}
        & \text{BC, LQ, K=1} & 11.16 & 20.84 & 12.91 & 64.9 & 65.5 & 65.0 & 88.9 & 85.6 & 88.4 & 52.8 \\
        & & \text{BC, LQ, K=3} & 11.31 & 17.09 & 12.38 & 63.4 & 64.8 & 63.7 & 88.4 & 86.4 & 88.0 & 50.9 \\
        & & \text{BC, LQ, K=5} & 9.77 & 15.95 & 10.88 & 64.3 & 65.4 & 64.5 & 89.5 & 87.1 & 89.1 & 51.4 \\
        & & \text{BC, MA, K=1} & 11.17 & 20.84 & 12.91 & 64.9 & 65.5 & 65.0 & 88.9 & 85.6 & 88.4 & 52.8 \\
        & & \text{BC, MA, K=3} & 9.40 & 16.76 & 10.73 & 64.8 & 65.6 & 65.0 & 90.2 & 87.5 & 89.7 & 51.3 \\
        & & \text{BC, MA, K=5} & 9.20 & 16.85 & 10.57 & 65.0 & 65.6 & 65.1 & 90.2 & 87.5 & 89.7 & 52.1 \\
        & & \text{DIM, LQ, K=1} & 12.78 & 20.78 & 14.28 & 63.5 & 63.7 & 63.6 & 86.9 & 83.9 & 86.3 & 52.0 \\
        & & \text{DIM, LQ, K=3} & 12.66 & 21.40 & 14.32 & 63.6 & 63.9 & 63.7 & 89.1 & 86.0 & 88.5 & 51.4 \\
        & & \text{DIM, LQ, K=5} & 13.26 & 22.59 & 15.05 & 63.5 & 63.8 & 63.6 & 89.5 & 86.5 & 88.9 & 51.2 \\
        & & \text{DIM, MA, K=1} & 12.81 & 20.83 & 14.32 & 63.6 & 63.8 & 63.6 & 86.9 & 83.9 & 86.3 & 51.8 \\
        & & \text{DIM, MA, K=3} & 12.74 & 21.51 & 14.42 & 63.6 & 63.9 & 63.7 & 89.1 & 86.0 & 88.5 & 51.1 \\
        & & \text{DIM, MA, K=5} & 13.37 & 22.68 & 15.16 & 63.5 & 63.7 & 63.5 & 89.5 & 86.5 & 88.9 & 50.9 \\
    \arrayrulecolor{black!25}\cmidrule(lr{1em}){2-13}
    & \multirow{12}*{\shortstack[l]{Dropout\\ Ensemble}}
        & \text{BC, LQ, K=1} & 9.06 & 15.49 & 10.22 & \bf{65.3} & \bf{66.1} & \bf{65.5} & 89.5 & 86.4 & 89.0 & 53.7 \\
        & & \text{BC, LQ, K=3} & \bf{8.22} & \bf{14.83} & \bf{9.39} & 64.9 & 65.6 & 65.1 & \bf{90.5} & 87.5 & \bf{90.0} & 53.9 \\
        & & \text{BC, LQ, K=5} & 8.39 & 15.16 & 9.57 & 65.0 & 65.7 & 65.2 & 90.4 & 87.6 & 89.9 & \bf{54.5} \\
        & & \text{BC, MA, K=1} & 9.07 & 15.50 & 10.22 & \bf{65.3} & \bf{66.1} & \bf{65.5} & 89.5 & 86.4 & 89.0 & 53.7 \\
        & & \text{BC, MA, K=3} & 8.69 & 15.90 & 9.99 & 64.9 & 65.6 & 65.1 & \bf{90.5} & \bf{87.7} & \bf{90.0} & 53.0 \\
        & & \text{BC, MA, K=5} & 9.05 & 16.69 & 10.41 & 65.0 & 65.6 & 65.1 & 90.4 & 87.6 & 89.9 & 53.2 \\
        & & \text{DIM, LQ, K=1} & 12.45 & 20.27 & 13.92 & 63.6 & 63.7 & 63.6 & 87.7 & 84.7 & 87.1 & 51.8 \\
        & & \text{DIM, LQ, K=3} & 12.63 & 21.32 & 14.29 & 63.7 & 63.9 & 63.7 & 89.1 & 86.1 & 88.6 & 51.3 \\
        & & \text{DIM, LQ, K=5} & 13.22 & 22.78 & 15.04 & 63.5 & 63.8 & 63.6 & 89.4 & 86.3 & 88.8 & 51.2 \\
        & & \text{DIM, MA, K=1} & 12.51 & 20.33 & 14.00 & 63.6 & 63.8 & 63.7 & 87.7 & 84.7 & 87.1 & 51.5 \\
        & & \text{DIM, MA, K=3} & 12.73 & 21.43 & 14.40 & 63.6 & 63.9 & 63.7 & 89.1 & 86.0 & 88.5 & 51.1 \\
        & & \text{DIM, MA, K=5} & 13.36 & 22.85 & 15.19 & 63.5 & 63.8 & 63.6 & 89.4 & 86.3 & 88.8 & 50.9 \\
        \bottomrule
    \end{tabular}
}
\end{table}

\begin{figure}[htbp!]
     \centering
     \begin{subfigure}[b]{0.49\textwidth}
         \centering
         \includegraphics[width=\textwidth]{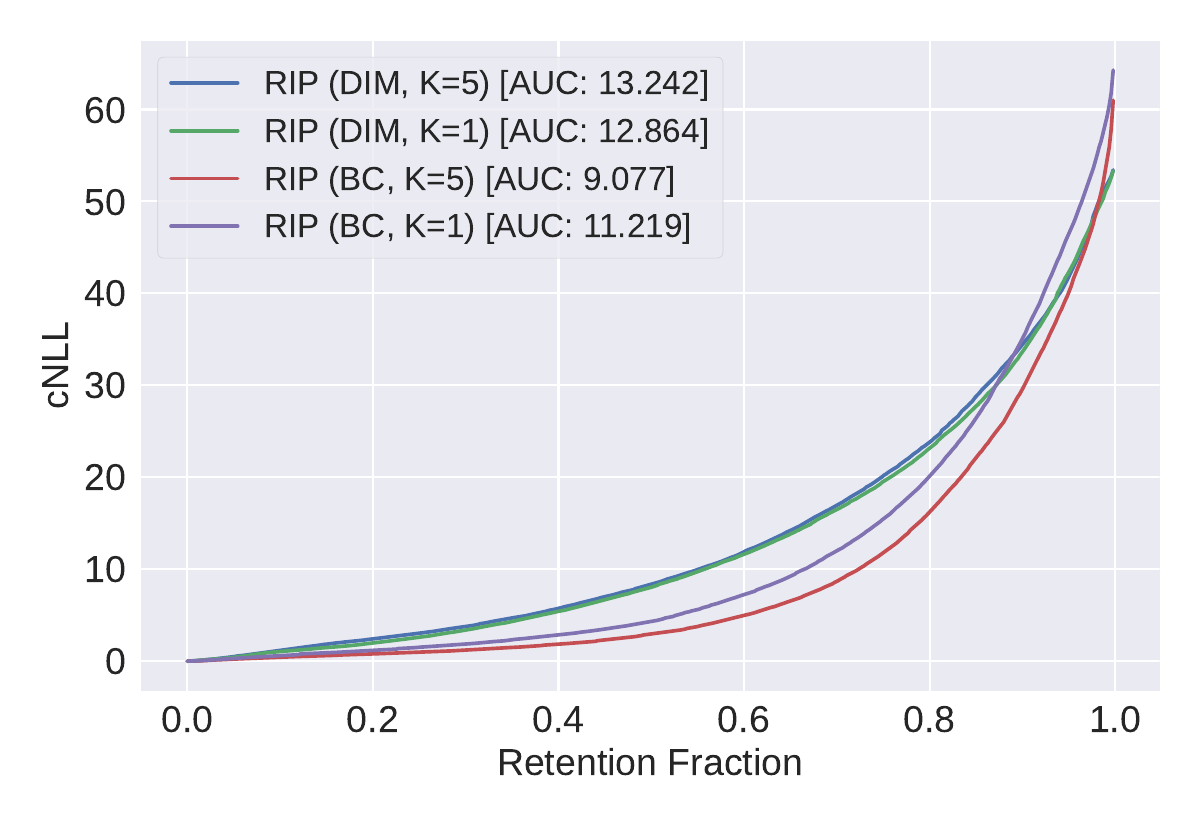}
         \caption{Full \texttt{dev} cNLL retention.}
         \label{fig-apn:sdc_ret_dev}
     \end{subfigure}
     \begin{subfigure}[b]{0.49\textwidth}
         \centering
         \includegraphics[width=\textwidth]{figures/cNLL_error_retention_eval.pdf}
         \caption{Full \texttt{eval} cNLL retention.}
         \label{fig-apn:sdc_ret_eval}
     \end{subfigure}
     \hfill
     \\
     \begin{subfigure}[b]{0.49\textwidth}
         \centering
         \includegraphics[width=\textwidth]{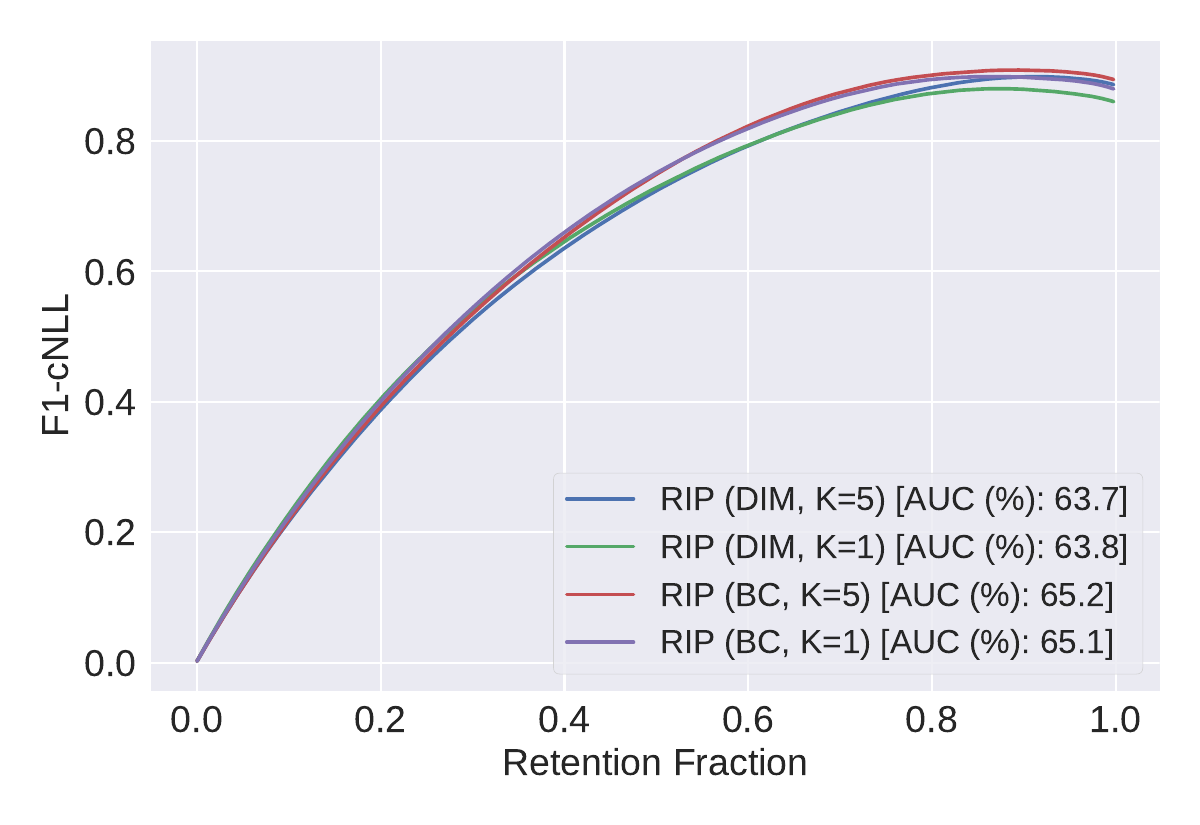}
         \caption{Full \texttt{dev} F1-cNLL retention.}
         \label{fig-apn:sdc_fbeta_dev}
     \end{subfigure}
     \begin{subfigure}[b]{0.49\textwidth}
         \centering
         \includegraphics[width=\textwidth]{figures/cNLL_F1_retention_eval.pdf}
         \caption{Full \texttt{eval} F1-cNLL retention.}
         \label{fig-apn:sdc_fbeta_eval}
     \end{subfigure} 
        \caption{cNLL and F1-cNLL retention curves on the Full (i.e., containing both the in-distribution and distributionally shifted datapoints) \texttt{dev} (left column) and \texttt{eval} (right column) partitions of the Vehicle Motion Prediction dataset.
        Top row: retention on cNLL (lower $\downarrow$ AUC is better). Bottom row: retention on F1-cNLL (higher $\uparrow$ AUC is better).
        We vary the backbone model and number of ensemble members, fix the Model Averaging (MA) aggregation strategy for the per-trajectory aggregation operator $\oplus_{\text{trajectory}}$ and the per--prediction request aggregation operator $\oplus_{\text{pred-req}}$ (based on results from \cref{table:sdc_uncertainty}), and otherwise use the standard RIP settings enumerated in \cref{sec:sdc-setup}.}
        \label{fig-apn:sdc_retention}
\end{figure}

\end{document}